\documentclass[11pt]{article}

\usepackage[preprint]{acl}
\usepackage{times}
\usepackage{latexsym}
\usepackage[T1]{fontenc}
\usepackage[utf8]{inputenc}
\usepackage{microtype}
\usepackage{inconsolata}
\usepackage{graphicx}
\usepackage{amsmath}
\usepackage{amssymb}
\usepackage{pifont}
\usepackage{enumitem}
\usepackage{stfloats}
\usepackage{booktabs}
\usepackage{tabularx}
\usepackage{multicol}
\usepackage{tcolorbox}
\tcbuselibrary{breakable}
\definecolor{appendixBlue}{HTML}{DDECF8}
\definecolor{appendixBlueFrame}{HTML}{7FA8C9}
\definecolor{appendixBlueText}{HTML}{173B57}
\newtcolorbox{appendixquote}[2][]{colback=appendixBlue!35!white,
  colframe=appendixBlueFrame,colbacktitle=appendixBlue,
  coltitle=appendixBlueText,fonttitle=\bfseries\small,title={#2},
  fontupper=\small\raggedright\sloppy,boxrule=.6pt,arc=3pt,left=6pt,right=6pt,
  top=5pt,bottom=5pt,before skip=6pt,after skip=6pt,#1}
\usepackage{tikz}
\usetikzlibrary{arrows.meta}
\makeatletter
\newcommand{\appendixcaption}[2]{\def\@captype{#1}\caption{#2}}
\makeatother

\makeatletter
\AtEndEnvironment{thebibliography}{%
  \if@newlist
    \item[]\textcolor{red}{[References pending: no active citations in the current manuscript.]}%
  \fi
}
\makeatother

\title{FinGREAT: Evaluating LLMs on\\\underline{Fin}ancial \underline{G}raph-guided \underline{Re}trieval \underline{A}nd \underline{T}yping}

\author{
  Ashman Mehra \\
  {\small School of Computer Science} \\
  {\small AI Accounting Lab} \\
  {\small Carnegie Mellon University} \\
  {\small \texttt{ashmanm@andrew.cmu.edu}} \\\And
  Bryan Routledge \\
  {\small Tepper School of Business} \\
  {\small AI Accounting Lab} \\
  {\small Carnegie Mellon University} \\
  {\small \texttt{routledge@andrew.cmu.edu}} \\\And
  Jane JaeYeon Pyo \\
  {\small Tepper School of Business} \\
  {\small AI Accounting Lab} \\
  {\small Carnegie Mellon University} \\
  {\small \texttt{jaeyeonp@andrew.cmu.edu}} \\\And
  Pierre Jinghong Liang \\
  {\small Tepper School of Business} \\
  {\small AI Accounting Lab} \\
  {\small Carnegie Mellon University} \\
  {\small \texttt{liangj@andrew.cmu.edu}} \\
}

\usepackage{pgfplots,multirow,arydshln}
\pgfplotsset{compat=1.18}
\definecolor{teal}{HTML}{008C95}
\definecolor{orange}{HTML}{E69F00}
\definecolor{violet}{HTML}{9B59B6}
\definecolor{blue}{HTML}{0072B2}
\definecolor{plotRed}{HTML}{D55E00}
\definecolor{rankGold}{HTML}{F7E4A3}
\definecolor{rankSilver}{HTML}{E2E6ED}
\definecolor{rankBronze}{HTML}{EAC7AB}
\newif\ifleaderboardhighlights
\leaderboardhighlightstrue
\newcommand{\paperbadge}[3][none]{\begingroup
\ifleaderboardhighlights\def\badgefill{#2}\def\badgeborder{#1}\else\def\badgefill{none}\def\badgeborder{none}\fi
\tikz[baseline=(score.base)]{\node[fill=\badgefill,draw=\badgeborder,line width=.6pt,rounded corners=2pt,
inner xsep=2pt,inner ysep=1pt,outer sep=0pt,text=black,
font=\normalfont\selectfont] (score) {#3};}\endgroup}
\newcommand{\paperheadername}[1]{\begin{tabular}[t]{@{}c@{}}#1\end{tabular}}

\definecolor{familyBaralibaba}{HTML}{ff7018}
\newcommand{\providerLogoalibaba}{\tikz[baseline=-.5ex,scale=0.229167,yscale=-1]{
\path[use as bounding box] (0pt,0pt) rectangle (48.0pt,48.0pt);
\definecolor{familyLogoalibaba0}{HTML}{FF5A00}
\begin{scope}[fill=familyLogoalibaba0]
\pgfpathmoveto{\pgfpoint{31.412700pt}{8.500000pt}}
\pgfpathlineto{\pgfpoint{33.565100pt}{8.500000pt}}
\pgfpathcurveto{\pgfpoint{36.590500pt}{8.671850pt}}{\pgfpoint{39.803200pt}{9.235140pt}}{\pgfpoint{42.212700pt}{11.221000pt}}
\pgfpathcurveto{\pgfpoint{44.136500pt}{12.783500pt}}{\pgfpoint{44.952400pt}{15.453600pt}}{\pgfpoint{44.479400pt}{17.859500pt}}
\pgfpathcurveto{\pgfpoint{44.117500pt}{19.654400pt}}{\pgfpoint{43.292100pt}{21.309300pt}}{\pgfpoint{42.403200pt}{22.890900pt}}
\pgfpathcurveto{\pgfpoint{40.600000pt}{26.022400pt}}{\pgfpoint{38.384100pt}{28.880300pt}}{\pgfpoint{36.276200pt}{31.804900pt}}
\pgfpathcurveto{\pgfpoint{35.898400pt}{32.288600pt}}{\pgfpoint{35.603200pt}{32.880600pt}}{\pgfpoint{35.641300pt}{33.510700pt}}
\pgfpathcurveto{\pgfpoint{35.676200pt}{34.010300pt}}{\pgfpoint{36.057100pt}{34.414500pt}}{\pgfpoint{36.504800pt}{34.592700pt}}
\pgfpathcurveto{\pgfpoint{37.371400pt}{34.946000pt}}{\pgfpoint{38.330200pt}{34.875900pt}}{\pgfpoint{39.241300pt}{34.825000pt}}
\pgfpathcurveto{\pgfpoint{42.120600pt}{34.522700pt}}{\pgfpoint{44.895200pt}{33.647500pt}}{\pgfpoint{47.641300pt}{32.769200pt}}
\pgfpathcurveto{\pgfpoint{47.850800pt}{32.638700pt}}{\pgfpoint{47.920600pt}{32.851900pt}}{\pgfpoint{48.000000pt}{33.004700pt}}
\pgfpathcurveto{\pgfpoint{45.082500pt}{34.955500pt}}{\pgfpoint{41.942900pt}{36.581700pt}}{\pgfpoint{38.641300pt}{37.778300pt}}
\pgfpathcurveto{\pgfpoint{36.815900pt}{38.392500pt}}{\pgfpoint{34.885700pt}{38.974900pt}}{\pgfpoint{32.939700pt}{38.745800pt}}
\pgfpathcurveto{\pgfpoint{31.917500pt}{38.643900pt}}{\pgfpoint{30.892100pt}{38.032900pt}}{\pgfpoint{30.571400pt}{37.011300pt}}
\pgfpathcurveto{\pgfpoint{30.174600pt}{35.681100pt}}{\pgfpoint{30.320600pt}{34.207600pt}}{\pgfpoint{30.895200pt}{32.950600pt}}
\pgfpathcurveto{\pgfpoint{31.768300pt}{30.987000pt}}{\pgfpoint{33.123800pt}{29.300300pt}}{\pgfpoint{34.409500pt}{27.597700pt}}
\pgfpathcurveto{\pgfpoint{36.539700pt}{24.797200pt}}{\pgfpoint{38.739700pt}{22.012600pt}}{\pgfpoint{40.412700pt}{18.900200pt}}
\pgfpathcurveto{\pgfpoint{40.965100pt}{17.795900pt}}{\pgfpoint{41.555600pt}{16.456100pt}}{\pgfpoint{41.006300pt}{15.237200pt}}
\pgfpathcurveto{\pgfpoint{40.488900pt}{14.113800pt}}{\pgfpoint{39.365100pt}{13.451900pt}}{\pgfpoint{38.288900pt}{12.952200pt}}
\pgfpathcurveto{\pgfpoint{36.377800pt}{12.096100pt}}{\pgfpoint{34.339700pt}{11.580600pt}}{\pgfpoint{32.330200pt}{11.007800pt}}
\pgfpathcurveto{\pgfpoint{32.027500pt}{11.207200pt}}{\pgfpoint{31.725900pt}{11.408700pt}}{\pgfpoint{31.425400pt}{11.612400pt}}
\pgfpathcurveto{\pgfpoint{32.123800pt}{12.147100pt}}{\pgfpoint{32.822200pt}{12.681700pt}}{\pgfpoint{33.514300pt}{13.225900pt}}
\pgfpathcurveto{\pgfpoint{30.869800pt}{13.703300pt}}{\pgfpoint{28.234900pt}{14.237900pt}}{\pgfpoint{25.631700pt}{14.903000pt}}
\pgfpathcurveto{\pgfpoint{21.666700pt}{15.889600pt}}{\pgfpoint{17.796800pt}{17.210300pt}}{\pgfpoint{13.961900pt}{18.610600pt}}
\pgfpathcurveto{\pgfpoint{14.336500pt}{19.358400pt}}{\pgfpoint{14.707900pt}{20.109500pt}}{\pgfpoint{15.063500pt}{20.866900pt}}
\pgfpathcurveto{\pgfpoint{14.196800pt}{21.828000pt}}{\pgfpoint{13.336500pt}{22.798600pt}}{\pgfpoint{12.469800pt}{23.759700pt}}
\pgfpathcurveto{\pgfpoint{14.552400pt}{24.383500pt}}{\pgfpoint{16.796800pt}{24.469400pt}}{\pgfpoint{18.920600pt}{23.998400pt}}
\pgfpathcurveto{\pgfpoint{20.736500pt}{23.591100pt}}{\pgfpoint{22.444400pt}{22.719100pt}}{\pgfpoint{23.831700pt}{21.474700pt}}
\pgfpathcurveto{\pgfpoint{23.454000pt}{21.003700pt}}{\pgfpoint{22.977800pt}{20.625000pt}}{\pgfpoint{22.463500pt}{20.313200pt}}
\pgfpathcurveto{\pgfpoint{24.149200pt}{20.348200pt}}{\pgfpoint{25.682500pt}{21.821600pt}}{\pgfpoint{25.685700pt}{23.530600pt}}
\pgfpathcurveto{\pgfpoint{25.355600pt}{23.532700pt}}{\pgfpoint{25.027500pt}{23.534800pt}}{\pgfpoint{24.701600pt}{23.537000pt}}
\pgfpathcurveto{\pgfpoint{24.631700pt}{23.164600pt}}{\pgfpoint{24.533300pt}{22.805000pt}}{\pgfpoint{24.403200pt}{22.451800pt}}
\pgfpathcurveto{\pgfpoint{22.488900pt}{24.186200pt}}{\pgfpoint{20.015900pt}{25.271400pt}}{\pgfpoint{17.454000pt}{25.551400pt}}
\pgfpathcurveto{\pgfpoint{15.196800pt}{25.812400pt}}{\pgfpoint{12.888900pt}{25.500500pt}}{\pgfpoint{10.765100pt}{24.704900pt}}
\pgfpathcurveto{\pgfpoint{10.907900pt}{26.022400pt}}{\pgfpoint{11.050800pt}{27.333600pt}}{\pgfpoint{11.184100pt}{28.651100pt}}
\pgfpathcurveto{\pgfpoint{9.977780pt}{29.122100pt}}{\pgfpoint{8.857140pt}{29.803200pt}}{\pgfpoint{7.853970pt}{30.624200pt}}
\pgfpathcurveto{\pgfpoint{7.057140pt}{31.321200pt}}{\pgfpoint{6.266670pt}{32.123100pt}}{\pgfpoint{5.961900pt}{33.167000pt}}
\pgfpathcurveto{\pgfpoint{5.669840pt}{34.086700pt}}{\pgfpoint{6.146030pt}{35.130500pt}}{\pgfpoint{6.949210pt}{35.623800pt}}
\pgfpathcurveto{\pgfpoint{7.923810pt}{36.228500pt}}{\pgfpoint{9.060320pt}{36.508500pt}}{\pgfpoint{10.184100pt}{36.670800pt}}
\pgfpathcurveto{\pgfpoint{11.666700pt}{36.865000pt}}{\pgfpoint{13.165100pt}{36.849000pt}}{\pgfpoint{14.654000pt}{36.744000pt}}
\pgfpathcurveto{\pgfpoint{17.577800pt}{36.527600pt}}{\pgfpoint{20.466700pt}{35.977100pt}}{\pgfpoint{23.311100pt}{35.286500pt}}
\pgfpathcurveto{\pgfpoint{23.593600pt}{35.171900pt}}{\pgfpoint{23.727000pt}{35.639700pt}}{\pgfpoint{23.438100pt}{35.712900pt}}
\pgfpathcurveto{\pgfpoint{21.168300pt}{36.884000pt}}{\pgfpoint{18.793600pt}{37.870600pt}}{\pgfpoint{16.317500pt}{38.507100pt}}
\pgfpathcurveto{\pgfpoint{13.990500pt}{39.108600pt}}{\pgfpoint{11.593700pt}{39.417300pt}}{\pgfpoint{9.193650pt}{39.500000pt}}
\pgfpathcurveto{\pgfpoint{6.746030pt}{39.423600pt}}{\pgfpoint{4.146030pt}{38.949400pt}}{\pgfpoint{2.206350pt}{37.342300pt}}
\pgfpathcurveto{\pgfpoint{0.765079pt}{36.171200pt}}{\pgfpoint{0.006349pt}{34.312600pt}}{\pgfpoint{0.000000pt}{32.470000pt}}
\pgfpathlineto{\pgfpoint{0.000000pt}{32.371400pt}}
\pgfpathcurveto{\pgfpoint{0.047619pt}{30.423700pt}}{\pgfpoint{0.704762pt}{28.539700pt}}{\pgfpoint{1.650790pt}{26.853000pt}}
\pgfpathcurveto{\pgfpoint{2.895240pt}{24.472600pt}}{\pgfpoint{4.428570pt}{22.238500pt}}{\pgfpoint{6.241270pt}{20.255900pt}}
\pgfpathcurveto{\pgfpoint{9.374600pt}{16.796600pt}}{\pgfpoint{13.304800pt}{14.120200pt}}{\pgfpoint{17.530200pt}{12.182100pt}}
\pgfpathcurveto{\pgfpoint{21.907900pt}{10.193100pt}}{\pgfpoint{26.593600pt}{8.764140pt}}{\pgfpoint{31.412700pt}{8.500000pt}}
\pgfpathclose
\pgfusepath{fill}
\end{scope}
}}
\definecolor{familyBargoogle}{HTML}{34A853}
\newcommand{\providerLogogoogle}{\tikz[baseline=-.5ex,scale=0.687500,yscale=-1]{
\path[use as bounding box] (0pt,0pt) rectangle (16.0pt,16.0pt);
\definecolor{familyLogogoogle0}{HTML}{4285F4}
\begin{scope}[fill=familyLogogoogle0]
\pgfpathmoveto{\pgfpoint{14.900000pt}{8.161000pt}}
\pgfpathcurveto{\pgfpoint{14.900000pt}{7.685000pt}}{\pgfpoint{14.861000pt}{7.207000pt}}{\pgfpoint{14.779000pt}{6.739000pt}}
\pgfpathlineto{\pgfpoint{8.139000pt}{6.739000pt}}
\pgfpathlineto{\pgfpoint{8.139000pt}{9.434000pt}}
\pgfpathlineto{\pgfpoint{11.941000pt}{9.434000pt}}
\pgfpathcurveto{\pgfpoint{11.862560pt}{9.865266pt}}{\pgfpoint{11.697411pt}{10.276162pt}}{\pgfpoint{11.455573pt}{10.641754pt}}
\pgfpathcurveto{\pgfpoint{11.213736pt}{11.007347pt}}{\pgfpoint{10.900203pt}{11.320088pt}}{\pgfpoint{10.534000pt}{11.561000pt}}
\pgfpathlineto{\pgfpoint{10.534000pt}{13.311000pt}}
\pgfpathlineto{\pgfpoint{12.803000pt}{13.311000pt}}
\pgfpathcurveto{\pgfpoint{14.135000pt}{12.091000pt}}{\pgfpoint{14.900000pt}{10.291000pt}}{\pgfpoint{14.900000pt}{8.161000pt}}
\pgfpathlineto{\pgfpoint{14.900000pt}{8.161000pt}}
\pgfpathclose
\pgfusepath{fill}
\end{scope}
\definecolor{familyLogogoogle1}{HTML}{34A853}
\begin{scope}[fill=familyLogogoogle1]
\pgfpathmoveto{\pgfpoint{8.140000pt}{15.000000pt}}
\pgfpathcurveto{\pgfpoint{10.038000pt}{15.000000pt}}{\pgfpoint{11.639000pt}{14.380000pt}}{\pgfpoint{12.805000pt}{13.310000pt}}
\pgfpathlineto{\pgfpoint{10.537000pt}{11.561000pt}}
\pgfpathcurveto{\pgfpoint{9.906000pt}{11.988000pt}}{\pgfpoint{9.091000pt}{12.230000pt}}{\pgfpoint{8.142000pt}{12.230000pt}}
\pgfpathcurveto{\pgfpoint{6.306000pt}{12.230000pt}}{\pgfpoint{4.749000pt}{10.998000pt}}{\pgfpoint{4.190000pt}{9.342000pt}}
\pgfpathlineto{\pgfpoint{1.850000pt}{9.342000pt}}
\pgfpathlineto{\pgfpoint{1.850000pt}{11.145000pt}}
\pgfpathcurveto{\pgfpoint{2.439546pt}{12.306121pt}}{\pgfpoint{3.339685pt}{13.281330pt}}{\pgfpoint{4.449969pt}{13.961798pt}}
\pgfpathcurveto{\pgfpoint{5.560253pt}{14.642266pt}}{\pgfpoint{6.837785pt}{15.001704pt}}{\pgfpoint{8.140000pt}{15.000000pt}}
\pgfpathclose
\pgfusepath{fill}
\end{scope}
\definecolor{familyLogogoogle2}{HTML}{FBBC04}
\begin{scope}[fill=familyLogogoogle2]
\pgfpathmoveto{\pgfpoint{4.187000pt}{9.342000pt}}
\pgfpathcurveto{\pgfpoint{3.892168pt}{8.473161pt}}{\pgfpoint{3.892168pt}{7.530839pt}}{\pgfpoint{4.187000pt}{6.662000pt}}
\pgfpathlineto{\pgfpoint{4.187000pt}{4.859000pt}}
\pgfpathlineto{\pgfpoint{1.849000pt}{4.859000pt}}
\pgfpathcurveto{\pgfpoint{1.356668pt}{5.833503pt}}{\pgfpoint{1.100129pt}{6.910191pt}}{\pgfpoint{1.100129pt}{8.002000pt}}
\pgfpathcurveto{\pgfpoint{1.100129pt}{9.093809pt}}{\pgfpoint{1.356668pt}{10.170497pt}}{\pgfpoint{1.849000pt}{11.145000pt}}
\pgfpathlineto{\pgfpoint{4.187000pt}{9.342000pt}}
\pgfpathlineto{\pgfpoint{4.187000pt}{9.342000pt}}
\pgfpathclose
\pgfusepath{fill}
\end{scope}
\definecolor{familyLogogoogle3}{HTML}{EA4335}
\begin{scope}[fill=familyLogogoogle3]
\pgfpathmoveto{\pgfpoint{8.140000pt}{3.770000pt}}
\pgfpathcurveto{\pgfpoint{9.142264pt}{3.753752pt}}{\pgfpoint{10.112043pt}{4.130888pt}}{\pgfpoint{10.840000pt}{4.820000pt}}
\pgfpathlineto{\pgfpoint{12.850000pt}{2.821000pt}}
\pgfpathcurveto{\pgfpoint{11.572995pt}{1.631466pt}}{\pgfpoint{9.885068pt}{0.979231pt}}{\pgfpoint{8.140000pt}{1.001000pt}}
\pgfpathcurveto{\pgfpoint{6.837425pt}{0.999350pt}}{\pgfpoint{5.559582pt}{1.359142pt}}{\pgfpoint{4.449227pt}{2.040184pt}}
\pgfpathcurveto{\pgfpoint{3.338872pt}{2.721225pt}}{\pgfpoint{2.438951pt}{3.697173pt}}{\pgfpoint{1.850000pt}{4.859000pt}}
\pgfpathlineto{\pgfpoint{4.186000pt}{6.660000pt}}
\pgfpathcurveto{\pgfpoint{4.742000pt}{5.002000pt}}{\pgfpoint{6.302000pt}{3.770000pt}}{\pgfpoint{8.138000pt}{3.770000pt}}
\pgfpathlineto{\pgfpoint{8.140000pt}{3.770000pt}}
\pgfpathclose
\pgfusepath{fill}
\end{scope}
}}
\definecolor{familyBarmistral}{HTML}{f32e02}
\newcommand{\providerLogomistral}{\tikz[baseline=-.5ex,scale=0.229167,yscale=-1]{
\path[use as bounding box] (0pt,0pt) rectangle (48.0pt,48.0pt);
\definecolor{familyLogomistral0}{HTML}{FFAF01}
\begin{scope}[fill=familyLogomistral0]
\pgfpathmoveto{\pgfpoint{13.710100pt}{7.000000pt}}
\pgfpathlineto{\pgfpoint{6.852620pt}{7.000000pt}}
\pgfpathlineto{\pgfpoint{6.852620pt}{13.800100pt}}
\pgfpathlineto{\pgfpoint{13.710100pt}{13.800100pt}}
\pgfpathlineto{\pgfpoint{13.710100pt}{7.000000pt}}
\pgfpathclose
\pgfusepath{fill}
\end{scope}
\definecolor{familyLogomistral1}{HTML}{FF8204}
\begin{scope}[fill=familyLogomistral1]
\pgfpathmoveto{\pgfpoint{20.567600pt}{13.799500pt}}
\pgfpathlineto{\pgfpoint{6.852620pt}{13.800100pt}}
\pgfpathlineto{\pgfpoint{6.852620pt}{20.599500pt}}
\pgfpathlineto{\pgfpoint{20.567600pt}{20.599500pt}}
\pgfpathlineto{\pgfpoint{20.567600pt}{13.799500pt}}
\pgfpathclose
\pgfusepath{fill}
\end{scope}
\definecolor{familyLogomistral2}{HTML}{FA500F}
\begin{scope}[fill=familyLogomistral2]
\pgfpathmoveto{\pgfpoint{41.140100pt}{20.600000pt}}
\pgfpathlineto{\pgfpoint{6.852620pt}{20.599500pt}}
\pgfpathlineto{\pgfpoint{6.852620pt}{27.400000pt}}
\pgfpathlineto{\pgfpoint{41.140100pt}{27.400000pt}}
\pgfpathlineto{\pgfpoint{41.140100pt}{20.600000pt}}
\pgfpathclose
\pgfusepath{fill}
\end{scope}
\definecolor{familyLogomistral3}{HTML}{E51300}
\begin{scope}[fill=familyLogomistral3]
\pgfpathmoveto{\pgfpoint{41.137600pt}{27.399400pt}}
\pgfpathlineto{\pgfpoint{34.280200pt}{27.399400pt}}
\pgfpathlineto{\pgfpoint{34.280200pt}{34.199500pt}}
\pgfpathlineto{\pgfpoint{41.137600pt}{34.199500pt}}
\pgfpathlineto{\pgfpoint{41.137600pt}{27.399400pt}}
\pgfpathclose
\pgfusepath{fill}
\end{scope}
\definecolor{familyLogomistral4}{HTML}{E51300}
\begin{scope}[fill=familyLogomistral4]
\pgfpathmoveto{\pgfpoint{27.423900pt}{27.399400pt}}
\pgfpathlineto{\pgfpoint{20.566400pt}{27.399400pt}}
\pgfpathlineto{\pgfpoint{20.566400pt}{34.199500pt}}
\pgfpathlineto{\pgfpoint{27.423900pt}{34.199500pt}}
\pgfpathlineto{\pgfpoint{27.423900pt}{27.399400pt}}
\pgfpathclose
\pgfusepath{fill}
\end{scope}
\definecolor{familyLogomistral5}{HTML}{E51300}
\begin{scope}[fill=familyLogomistral5]
\pgfpathmoveto{\pgfpoint{13.710100pt}{27.399400pt}}
\pgfpathlineto{\pgfpoint{6.852620pt}{27.400000pt}}
\pgfpathlineto{\pgfpoint{6.852620pt}{34.199500pt}}
\pgfpathlineto{\pgfpoint{13.710100pt}{34.199500pt}}
\pgfpathlineto{\pgfpoint{13.710100pt}{27.399400pt}}
\pgfpathclose
\pgfusepath{fill}
\end{scope}
\definecolor{familyLogomistral6}{HTML}{C4001D}
\begin{scope}[fill=familyLogomistral6]
\pgfpathmoveto{\pgfpoint{48.000000pt}{34.199900pt}}
\pgfpathlineto{\pgfpoint{27.427500pt}{34.199900pt}}
\pgfpathlineto{\pgfpoint{27.427500pt}{41.000000pt}}
\pgfpathlineto{\pgfpoint{48.000000pt}{41.000000pt}}
\pgfpathlineto{\pgfpoint{48.000000pt}{34.199900pt}}
\pgfpathclose
\pgfusepath{fill}
\end{scope}
\definecolor{familyLogomistral7}{HTML}{C4001D}
\begin{scope}[fill=familyLogomistral7]
\pgfpathmoveto{\pgfpoint{20.572500pt}{34.199900pt}}
\pgfpathlineto{\pgfpoint{0.000000pt}{34.199900pt}}
\pgfpathlineto{\pgfpoint{0.000000pt}{41.000000pt}}
\pgfpathlineto{\pgfpoint{20.572500pt}{41.000000pt}}
\pgfpathlineto{\pgfpoint{20.572500pt}{34.199900pt}}
\pgfpathclose
\pgfusepath{fill}
\end{scope}
\definecolor{familyLogomistral8}{HTML}{FF8204}
\begin{scope}[fill=familyLogomistral8]
\pgfpathmoveto{\pgfpoint{41.142500pt}{13.799500pt}}
\pgfpathlineto{\pgfpoint{27.427500pt}{13.799500pt}}
\pgfpathlineto{\pgfpoint{27.427500pt}{20.599500pt}}
\pgfpathlineto{\pgfpoint{41.142500pt}{20.599500pt}}
\pgfpathlineto{\pgfpoint{41.142500pt}{13.799500pt}}
\pgfpathclose
\pgfusepath{fill}
\end{scope}
\definecolor{familyLogomistral9}{HTML}{FFAF01}
\begin{scope}[fill=familyLogomistral9]
\pgfpathmoveto{\pgfpoint{41.137600pt}{7.000000pt}}
\pgfpathlineto{\pgfpoint{34.280200pt}{7.000000pt}}
\pgfpathlineto{\pgfpoint{34.280200pt}{13.800100pt}}
\pgfpathlineto{\pgfpoint{41.137600pt}{13.800100pt}}
\pgfpathlineto{\pgfpoint{41.137600pt}{7.000000pt}}
\pgfpathclose
\pgfusepath{fill}
\end{scope}
}}
\definecolor{familyBarazure}{HTML}{0078D5}
\newcommand{\providerLogoazure}{\tikz[baseline=-.5ex,scale=0.229167,yscale=-1]{
\path[use as bounding box] (0pt,0pt) rectangle (48.0pt,48.0pt);
\definecolor{familyLogoazure0}{HTML}{F25022}
\begin{scope}[fill=familyLogoazure0]
\pgfpathmoveto{\pgfpoint{22.999800pt}{4.000000pt}}
\pgfpathlineto{\pgfpoint{4.000000pt}{4.000000pt}}
\pgfpathlineto{\pgfpoint{4.000000pt}{23.000000pt}}
\pgfpathlineto{\pgfpoint{22.999800pt}{23.000000pt}}
\pgfpathlineto{\pgfpoint{22.999800pt}{4.000000pt}}
\pgfpathclose
\pgfusepath{fill}
\end{scope}
\definecolor{familyLogoazure1}{HTML}{7FBA00}
\begin{scope}[fill=familyLogoazure1]
\pgfpathmoveto{\pgfpoint{44.000000pt}{4.000000pt}}
\pgfpathlineto{\pgfpoint{25.000200pt}{4.000000pt}}
\pgfpathlineto{\pgfpoint{25.000200pt}{23.000000pt}}
\pgfpathlineto{\pgfpoint{44.000000pt}{23.000000pt}}
\pgfpathlineto{\pgfpoint{44.000000pt}{4.000000pt}}
\pgfpathclose
\pgfusepath{fill}
\end{scope}
\definecolor{familyLogoazure2}{HTML}{00A4EF}
\begin{scope}[fill=familyLogoazure2]
\pgfpathmoveto{\pgfpoint{22.999800pt}{25.000000pt}}
\pgfpathlineto{\pgfpoint{4.000000pt}{25.000000pt}}
\pgfpathlineto{\pgfpoint{4.000000pt}{44.000000pt}}
\pgfpathlineto{\pgfpoint{22.999800pt}{44.000000pt}}
\pgfpathlineto{\pgfpoint{22.999800pt}{25.000000pt}}
\pgfpathclose
\pgfusepath{fill}
\end{scope}
\definecolor{familyLogoazure3}{HTML}{FFB900}
\begin{scope}[fill=familyLogoazure3]
\pgfpathmoveto{\pgfpoint{44.000000pt}{25.000000pt}}
\pgfpathlineto{\pgfpoint{25.000200pt}{25.000000pt}}
\pgfpathlineto{\pgfpoint{25.000200pt}{44.000000pt}}
\pgfpathlineto{\pgfpoint{44.000000pt}{44.000000pt}}
\pgfpathlineto{\pgfpoint{44.000000pt}{25.000000pt}}
\pgfpathclose
\pgfusepath{fill}
\end{scope}
}}
\definecolor{familyBarmeta}{HTML}{0089f4}
\newcommand{\providerLogometa}{\tikz[baseline=-.5ex,scale=0.229167,yscale=-1]{
\path[use as bounding box] (0pt,0pt) rectangle (48.0pt,48.0pt);
\definecolor{familyLogometa0}{HTML}{0866FF}
\begin{scope}[fill=familyLogometa0]
\pgfpathmoveto{\pgfpoint{13.790200pt}{8.000000pt}}
\pgfpathcurveto{\pgfpoint{13.774700pt}{8.000000pt}}{\pgfpoint{13.759400pt}{8.000050pt}}{\pgfpoint{13.744000pt}{8.000130pt}}
\pgfpathlineto{\pgfpoint{13.680900pt}{13.230600pt}}
\pgfpathcurveto{\pgfpoint{13.695300pt}{13.230400pt}}{\pgfpoint{13.709700pt}{13.230300pt}}{\pgfpoint{13.724200pt}{13.230300pt}}
\pgfpathcurveto{\pgfpoint{17.155700pt}{13.230300pt}}{\pgfpoint{19.817400pt}{15.944800pt}}{\pgfpoint{25.602500pt}{25.721200pt}}
\pgfpathlineto{\pgfpoint{25.978200pt}{26.355400pt}}
\pgfpathlineto{\pgfpoint{29.216700pt}{21.480400pt}}
\pgfpathlineto{\pgfpoint{29.194200pt}{21.442800pt}}
\pgfpathcurveto{\pgfpoint{28.432100pt}{20.199200pt}}{\pgfpoint{27.699900pt}{19.055000pt}}{\pgfpoint{26.997700pt}{18.010200pt}}
\pgfpathcurveto{\pgfpoint{26.183400pt}{16.799600pt}}{\pgfpoint{25.404500pt}{15.718200pt}}{\pgfpoint{24.648800pt}{14.752700pt}}
\pgfpathcurveto{\pgfpoint{20.821700pt}{9.863860pt}}{\pgfpoint{17.620300pt}{8.000000pt}}{\pgfpoint{13.790200pt}{8.000000pt}}
\pgfpathclose
\pgfusepath{fill}
\end{scope}
\definecolor{familyLogometa1}{HTML}{0866FF}
\begin{scope}[fill=familyLogometa1]
\pgfpathmoveto{\pgfpoint{13.744000pt}{8.000130pt}}
\pgfpathcurveto{\pgfpoint{9.895240pt}{8.020000pt}}{\pgfpoint{6.493550pt}{10.516900pt}}{\pgfpoint{4.036610pt}{14.337700pt}}
\pgfpathcurveto{\pgfpoint{4.029410pt}{14.348900pt}}{\pgfpoint{4.022210pt}{14.360100pt}}{\pgfpoint{4.015040pt}{14.371300pt}}
\pgfpathlineto{\pgfpoint{8.525010pt}{16.834100pt}}
\pgfpathcurveto{\pgfpoint{8.532340pt}{16.822900pt}}{\pgfpoint{8.539670pt}{16.811800pt}}{\pgfpoint{8.547030pt}{16.800700pt}}
\pgfpathcurveto{\pgfpoint{9.981740pt}{14.633500pt}}{\pgfpoint{11.766300pt}{13.251200pt}}{\pgfpoint{13.680900pt}{13.230600pt}}
\pgfpathcurveto{\pgfpoint{13.695300pt}{13.230400pt}}{\pgfpoint{13.709700pt}{13.230300pt}}{\pgfpoint{13.724200pt}{13.230300pt}}
\pgfpathlineto{\pgfpoint{13.790200pt}{8.000000pt}}
\pgfpathcurveto{\pgfpoint{13.774700pt}{8.000000pt}}{\pgfpoint{13.759400pt}{8.000050pt}}{\pgfpoint{13.744000pt}{8.000130pt}}
\pgfpathclose
\pgfusepath{fill}
\end{scope}
\definecolor{familyLogometa2}{HTML}{0866FF}
\begin{scope}[fill=familyLogometa2]
\pgfpathmoveto{\pgfpoint{4.031300pt}{14.340700pt}}
\pgfpathcurveto{\pgfpoint{4.024130pt}{14.351900pt}}{\pgfpoint{4.016930pt}{14.363100pt}}{\pgfpoint{4.009730pt}{14.374300pt}}
\pgfpathcurveto{\pgfpoint{2.395260pt}{16.895700pt}}{\pgfpoint{1.191710pt}{19.989900pt}}{\pgfpoint{0.541283pt}{23.327800pt}}
\pgfpathcurveto{\pgfpoint{0.538465pt}{23.342300pt}}{\pgfpoint{0.535658pt}{23.356700pt}}{\pgfpoint{0.532860pt}{23.371200pt}}
\pgfpathlineto{\pgfpoint{5.599630pt}{24.570600pt}}
\pgfpathcurveto{\pgfpoint{5.602290pt}{24.556200pt}}{\pgfpoint{5.604960pt}{24.541600pt}}{\pgfpoint{5.607620pt}{24.527200pt}}
\pgfpathcurveto{\pgfpoint{6.149210pt}{21.593600pt}}{\pgfpoint{7.180300pt}{18.872900pt}}{\pgfpoint{8.519700pt}{16.837100pt}}
\pgfpathcurveto{\pgfpoint{8.527000pt}{16.825900pt}}{\pgfpoint{8.534360pt}{16.814800pt}}{\pgfpoint{8.541720pt}{16.803700pt}}
\pgfpathlineto{\pgfpoint{4.031300pt}{14.340700pt}}
\pgfpathclose
\pgfusepath{fill}
\end{scope}
\definecolor{familyLogometa3}{HTML}{0866FF}
\begin{scope}[fill=familyLogometa3]
\pgfpathmoveto{\pgfpoint{5.613720pt}{24.528900pt}}
\pgfpathlineto{\pgfpoint{0.547355pt}{23.329500pt}}
\pgfpathcurveto{\pgfpoint{0.544538pt}{23.344000pt}}{\pgfpoint{0.541730pt}{23.358500pt}}{\pgfpoint{0.538932pt}{23.372900pt}}
\pgfpathcurveto{\pgfpoint{0.184261pt}{25.208700pt}}{\pgfpoint{0.003823pt}{27.074000pt}}{\pgfpoint{0.000037pt}{28.943900pt}}
\pgfpathcurveto{\pgfpoint{0.000007pt}{28.959000pt}}{\pgfpoint{-0.000005pt}{28.974200pt}}{\pgfpoint{0.000002pt}{28.989300pt}}
\pgfpathlineto{\pgfpoint{5.196200pt}{29.455700pt}}
\pgfpathcurveto{\pgfpoint{5.195790pt}{29.440500pt}}{\pgfpoint{5.195340pt}{29.425500pt}}{\pgfpoint{5.195000pt}{29.410300pt}}
\pgfpathcurveto{\pgfpoint{5.192430pt}{29.301800pt}}{\pgfpoint{5.191130pt}{29.191900pt}}{\pgfpoint{5.191090pt}{29.080500pt}}
\pgfpathcurveto{\pgfpoint{5.193690pt}{27.568300pt}}{\pgfpoint{5.332450pt}{26.059500pt}}{\pgfpoint{5.605720pt}{24.572400pt}}
\pgfpathcurveto{\pgfpoint{5.608360pt}{24.557900pt}}{\pgfpoint{5.611020pt}{24.543400pt}}{\pgfpoint{5.613720pt}{24.528900pt}}
\pgfpathclose
\pgfusepath{fill}
\end{scope}
\definecolor{familyLogometa4}{HTML}{0866FF}
\begin{scope}[fill=familyLogometa4]
\pgfpathmoveto{\pgfpoint{5.354220pt}{31.074900pt}}
\pgfpathcurveto{\pgfpoint{5.261010pt}{30.537500pt}}{\pgfpoint{5.208190pt}{29.993800pt}}{\pgfpoint{5.196200pt}{29.448400pt}}
\pgfpathcurveto{\pgfpoint{5.195790pt}{29.433300pt}}{\pgfpoint{5.195340pt}{29.418300pt}}{\pgfpoint{5.195000pt}{29.403000pt}}
\pgfpathlineto{\pgfpoint{0.000037pt}{28.936700pt}}
\pgfpathcurveto{\pgfpoint{0.000007pt}{28.951800pt}}{\pgfpoint{-0.000005pt}{28.966900pt}}{\pgfpoint{0.000002pt}{28.982100pt}}
\pgfpathcurveto{\pgfpoint{0.000002pt}{30.145800pt}}{\pgfpoint{0.096734pt}{31.238400pt}}{\pgfpoint{0.284733pt}{32.245000pt}}
\pgfpathcurveto{\pgfpoint{0.287452pt}{32.259600pt}}{\pgfpoint{0.290126pt}{32.273800pt}}{\pgfpoint{0.292883pt}{32.288400pt}}
\pgfpathlineto{\pgfpoint{5.362080pt}{31.118500pt}}
\pgfpathcurveto{\pgfpoint{5.359420pt}{31.104000pt}}{\pgfpoint{5.356820pt}{31.089500pt}}{\pgfpoint{5.354220pt}{31.074900pt}}
\pgfpathclose
\pgfusepath{fill}
\end{scope}
\definecolor{familyLogometa5}{HTML}{0866FF}
\begin{scope}[fill=familyLogometa5]
\pgfpathmoveto{\pgfpoint{6.538340pt}{33.779600pt}}
\pgfpathcurveto{\pgfpoint{5.972930pt}{33.160400pt}}{\pgfpoint{5.572770pt}{32.267200pt}}{\pgfpoint{5.362380pt}{31.124900pt}}
\pgfpathcurveto{\pgfpoint{5.359710pt}{31.110400pt}}{\pgfpoint{5.357140pt}{31.095900pt}}{\pgfpoint{5.354540pt}{31.081300pt}}
\pgfpathlineto{\pgfpoint{0.285018pt}{32.251400pt}}
\pgfpathcurveto{\pgfpoint{0.287734pt}{32.266000pt}}{\pgfpoint{0.290408pt}{32.280300pt}}{\pgfpoint{0.293168pt}{32.294800pt}}
\pgfpathcurveto{\pgfpoint{0.676340pt}{34.313200pt}}{\pgfpoint{1.427810pt}{35.995300pt}}{\pgfpoint{2.503970pt}{37.267900pt}}
\pgfpathcurveto{\pgfpoint{2.513540pt}{37.279200pt}}{\pgfpoint{2.523140pt}{37.290500pt}}{\pgfpoint{2.532780pt}{37.301800pt}}
\pgfpathlineto{\pgfpoint{6.568580pt}{33.812200pt}}
\pgfpathcurveto{\pgfpoint{6.558490pt}{33.801400pt}}{\pgfpoint{6.548340pt}{33.790500pt}}{\pgfpoint{6.538340pt}{33.779600pt}}
\pgfpathclose
\pgfusepath{fill}
\end{scope}
\definecolor{familyLogometa6}{HTML}{0082FB}
\begin{scope}[fill=familyLogometa6]
\pgfpathmoveto{\pgfpoint{21.556400pt}{19.307600pt}}
\pgfpathcurveto{\pgfpoint{18.501500pt}{24.008600pt}}{\pgfpoint{16.650900pt}{26.957200pt}}{\pgfpoint{16.650900pt}{26.957200pt}}
\pgfpathcurveto{\pgfpoint{12.581500pt}{33.357200pt}}{\pgfpoint{11.173700pt}{34.791600pt}}{\pgfpoint{8.908060pt}{34.791600pt}}
\pgfpathcurveto{\pgfpoint{7.962660pt}{34.791600pt}}{\pgfpoint{7.172870pt}{34.453900pt}}{\pgfpoint{6.567800pt}{33.808300pt}}
\pgfpathcurveto{\pgfpoint{6.557710pt}{33.797600pt}}{\pgfpoint{6.547590pt}{33.786600pt}}{\pgfpoint{6.537600pt}{33.775700pt}}
\pgfpathlineto{\pgfpoint{2.503200pt}{37.264100pt}}
\pgfpathcurveto{\pgfpoint{2.512770pt}{37.275400pt}}{\pgfpoint{2.522370pt}{37.286700pt}}{\pgfpoint{2.532010pt}{37.298000pt}}
\pgfpathcurveto{\pgfpoint{4.017940pt}{39.035700pt}}{\pgfpoint{6.115360pt}{40.000000pt}}{\pgfpoint{8.710110pt}{40.000000pt}}
\pgfpathcurveto{\pgfpoint{12.635800pt}{40.000000pt}}{\pgfpoint{15.459100pt}{38.143100pt}}{\pgfpoint{20.478300pt}{29.340700pt}}
\pgfpathcurveto{\pgfpoint{20.478300pt}{29.340700pt}}{\pgfpoint{22.570600pt}{25.633700pt}}{\pgfpoint{24.010000pt}{23.080200pt}}
\pgfpathcurveto{\pgfpoint{23.118400pt}{21.636000pt}}{\pgfpoint{22.307000pt}{20.385600pt}}{\pgfpoint{21.556400pt}{19.307600pt}}
\pgfpathclose
\pgfusepath{fill}
\end{scope}
\definecolor{familyLogometa7}{HTML}{0866FF}
\begin{scope}[fill=familyLogometa7]
\pgfpathmoveto{\pgfpoint{27.000100pt}{11.891400pt}}
\pgfpathcurveto{\pgfpoint{26.989800pt}{11.902400pt}}{\pgfpoint{26.979300pt}{11.913600pt}}{\pgfpoint{26.969000pt}{11.924600pt}}
\pgfpathcurveto{\pgfpoint{26.166800pt}{12.782100pt}}{\pgfpoint{25.396900pt}{13.739800pt}}{\pgfpoint{24.647900pt}{14.755800pt}}
\pgfpathcurveto{\pgfpoint{25.403700pt}{15.721400pt}}{\pgfpoint{26.183900pt}{16.804900pt}}{\pgfpoint{26.998300pt}{18.015500pt}}
\pgfpathcurveto{\pgfpoint{27.958100pt}{16.529200pt}}{\pgfpoint{28.854100pt}{15.325400pt}}{\pgfpoint{29.732200pt}{14.402700pt}}
\pgfpathcurveto{\pgfpoint{29.742600pt}{14.391800pt}}{\pgfpoint{29.752800pt}{14.381200pt}}{\pgfpoint{29.763200pt}{14.370300pt}}
\pgfpathlineto{\pgfpoint{27.000100pt}{11.891400pt}}
\pgfpathclose
\pgfusepath{fill}
\end{scope}
\definecolor{familyLogometa8}{HTML}{0082FB}
\begin{scope}[fill=familyLogometa8]
\pgfpathmoveto{\pgfpoint{41.827400pt}{11.426500pt}}
\pgfpathcurveto{\pgfpoint{39.696900pt}{9.267050pt}}{\pgfpoint{37.156700pt}{8.000000pt}}{\pgfpoint{34.441200pt}{8.000000pt}}
\pgfpathcurveto{\pgfpoint{31.577500pt}{8.000000pt}}{\pgfpoint{29.168800pt}{9.574360pt}}{\pgfpoint{26.996300pt}{11.887300pt}}
\pgfpathcurveto{\pgfpoint{26.986000pt}{11.898200pt}}{\pgfpoint{26.975500pt}{11.909400pt}}{\pgfpoint{26.965200pt}{11.920500pt}}
\pgfpathlineto{\pgfpoint{29.728400pt}{14.398600pt}}
\pgfpathcurveto{\pgfpoint{29.738800pt}{14.387600pt}}{\pgfpoint{29.749000pt}{14.377000pt}}{\pgfpoint{29.759400pt}{14.366200pt}}
\pgfpathcurveto{\pgfpoint{31.190600pt}{12.872600pt}}{\pgfpoint{32.575800pt}{12.126900pt}}{\pgfpoint{34.111300pt}{12.126900pt}}
\pgfpathcurveto{\pgfpoint{35.764100pt}{12.126900pt}}{\pgfpoint{37.310900pt}{12.907400pt}}{\pgfpoint{38.650600pt}{14.275700pt}}
\pgfpathcurveto{\pgfpoint{38.661100pt}{14.286400pt}}{\pgfpoint{38.671500pt}{14.297000pt}}{\pgfpoint{38.682000pt}{14.307800pt}}
\pgfpathlineto{\pgfpoint{41.859100pt}{11.458600pt}}
\pgfpathcurveto{\pgfpoint{41.848700pt}{11.447800pt}}{\pgfpoint{41.837900pt}{11.437200pt}}{\pgfpoint{41.827400pt}{11.426500pt}}
\pgfpathclose
\pgfusepath{fill}
\end{scope}
\definecolor{familyLogometa9}{HTML}{0866FF}
\begin{scope}[fill=familyLogometa9]
\pgfpathmoveto{\pgfpoint{47.986600pt}{28.249600pt}}
\pgfpathcurveto{\pgfpoint{47.867300pt}{21.316200pt}}{\pgfpoint{45.448700pt}{15.119000pt}}{\pgfpoint{41.859100pt}{11.458600pt}}
\pgfpathcurveto{\pgfpoint{41.848700pt}{11.447800pt}}{\pgfpoint{41.837900pt}{11.437200pt}}{\pgfpoint{41.827400pt}{11.426500pt}}
\pgfpathlineto{\pgfpoint{38.650600pt}{14.275700pt}}
\pgfpathcurveto{\pgfpoint{38.661100pt}{14.286400pt}}{\pgfpoint{38.671500pt}{14.297000pt}}{\pgfpoint{38.682000pt}{14.307800pt}}
\pgfpathcurveto{\pgfpoint{41.382300pt}{17.090800pt}}{\pgfpoint{43.235000pt}{22.266000pt}}{\pgfpoint{43.403500pt}{28.248100pt}}
\pgfpathcurveto{\pgfpoint{43.403800pt}{28.263200pt}}{\pgfpoint{43.404100pt}{28.278200pt}}{\pgfpoint{43.404700pt}{28.293300pt}}
\pgfpathlineto{\pgfpoint{47.987200pt}{28.294700pt}}
\pgfpathcurveto{\pgfpoint{47.987200pt}{28.279700pt}}{\pgfpoint{47.986900pt}{28.264700pt}}{\pgfpoint{47.986600pt}{28.249600pt}}
\pgfpathclose
\pgfusepath{fill}
\end{scope}
\definecolor{familyLogometa10}{HTML}{0866FF}
\begin{scope}[fill=familyLogometa10]
\pgfpathmoveto{\pgfpoint{47.996200pt}{28.299600pt}}
\pgfpathcurveto{\pgfpoint{47.996200pt}{28.284500pt}}{\pgfpoint{47.995900pt}{28.269500pt}}{\pgfpoint{47.995600pt}{28.254400pt}}
\pgfpathlineto{\pgfpoint{43.412400pt}{28.252900pt}}
\pgfpathcurveto{\pgfpoint{43.412700pt}{28.268000pt}}{\pgfpoint{43.413100pt}{28.283000pt}}{\pgfpoint{43.413700pt}{28.298100pt}}
\pgfpathcurveto{\pgfpoint{43.421000pt}{28.578900pt}}{\pgfpoint{43.424800pt}{28.861300pt}}{\pgfpoint{43.424800pt}{29.145500pt}}
\pgfpathcurveto{\pgfpoint{43.424800pt}{30.776500pt}}{\pgfpoint{43.182100pt}{32.094900pt}}{\pgfpoint{42.688200pt}{33.046800pt}}
\pgfpathcurveto{\pgfpoint{42.680900pt}{33.060900pt}}{\pgfpoint{42.673300pt}{33.075100pt}}{\pgfpoint{42.666000pt}{33.089000pt}}
\pgfpathlineto{\pgfpoint{46.082900pt}{36.654600pt}}
\pgfpathcurveto{\pgfpoint{46.091500pt}{36.641700pt}}{\pgfpoint{46.099700pt}{36.628800pt}}{\pgfpoint{46.108300pt}{36.615800pt}}
\pgfpathcurveto{\pgfpoint{47.348700pt}{34.695400pt}}{\pgfpoint{48.000000pt}{32.027500pt}}{\pgfpoint{48.000000pt}{28.792400pt}}
\pgfpathcurveto{\pgfpoint{48.000000pt}{28.627700pt}}{\pgfpoint{47.998700pt}{28.463400pt}}{\pgfpoint{47.996200pt}{28.299600pt}}
\pgfpathclose
\pgfusepath{fill}
\end{scope}
\definecolor{familyLogometa11}{HTML}{0866FF}
\begin{scope}[fill=familyLogometa11]
\pgfpathmoveto{\pgfpoint{42.688700pt}{33.039500pt}}
\pgfpathcurveto{\pgfpoint{42.681400pt}{33.053600pt}}{\pgfpoint{42.673800pt}{33.067800pt}}{\pgfpoint{42.666500pt}{33.081700pt}}
\pgfpathcurveto{\pgfpoint{42.239200pt}{33.884100pt}}{\pgfpoint{41.629200pt}{34.419200pt}}{\pgfpoint{40.831900pt}{34.653200pt}}
\pgfpathlineto{\pgfpoint{42.389600pt}{39.578700pt}}
\pgfpathcurveto{\pgfpoint{42.595800pt}{39.508400pt}}{\pgfpoint{42.796600pt}{39.429500pt}}{\pgfpoint{42.992000pt}{39.342000pt}}
\pgfpathcurveto{\pgfpoint{43.050400pt}{39.315900pt}}{\pgfpoint{43.108400pt}{39.289100pt}}{\pgfpoint{43.165900pt}{39.261500pt}}
\pgfpathcurveto{\pgfpoint{43.198800pt}{39.245600pt}}{\pgfpoint{43.231800pt}{39.229500pt}}{\pgfpoint{43.264500pt}{39.213100pt}}
\pgfpathcurveto{\pgfpoint{44.294600pt}{38.698000pt}}{\pgfpoint{45.165100pt}{37.938400pt}}{\pgfpoint{45.869000pt}{36.959600pt}}
\pgfpathcurveto{\pgfpoint{45.912200pt}{36.899800pt}}{\pgfpoint{45.954700pt}{36.839500pt}}{\pgfpoint{45.996200pt}{36.778200pt}}
\pgfpathcurveto{\pgfpoint{46.025800pt}{36.734900pt}}{\pgfpoint{46.054900pt}{36.691300pt}}{\pgfpoint{46.083500pt}{36.647300pt}}
\pgfpathcurveto{\pgfpoint{46.092100pt}{36.634400pt}}{\pgfpoint{46.100300pt}{36.621500pt}}{\pgfpoint{46.108900pt}{36.608500pt}}
\pgfpathlineto{\pgfpoint{42.688700pt}{33.039500pt}}
\pgfpathclose
\pgfusepath{fill}
\end{scope}
\definecolor{familyLogometa12}{HTML}{0866FF}
\begin{scope}[fill=familyLogometa12]
\pgfpathmoveto{\pgfpoint{39.833800pt}{34.786800pt}}
\pgfpathcurveto{\pgfpoint{39.311600pt}{34.786800pt}}{\pgfpoint{38.850700pt}{34.708800pt}}{\pgfpoint{38.399600pt}{34.506600pt}}
\pgfpathlineto{\pgfpoint{36.804500pt}{39.549300pt}}
\pgfpathcurveto{\pgfpoint{37.701000pt}{39.856500pt}}{\pgfpoint{38.657500pt}{39.994900pt}}{\pgfpoint{39.724000pt}{39.994900pt}}
\pgfpathcurveto{\pgfpoint{40.706500pt}{39.994900pt}}{\pgfpoint{41.608400pt}{39.847500pt}}{\pgfpoint{42.426300pt}{39.563100pt}}
\pgfpathlineto{\pgfpoint{40.869000pt}{34.639300pt}}
\pgfpathcurveto{\pgfpoint{40.533300pt}{34.740200pt}}{\pgfpoint{40.184400pt}{34.789900pt}}{\pgfpoint{39.833800pt}{34.786800pt}}
\pgfpathclose
\pgfusepath{fill}
\end{scope}
\definecolor{familyLogometa13}{HTML}{0866FF}
\begin{scope}[fill=familyLogometa13]
\pgfpathmoveto{\pgfpoint{36.638100pt}{33.067600pt}}
\pgfpathcurveto{\pgfpoint{36.628300pt}{33.056300pt}}{\pgfpoint{36.618500pt}{33.044800pt}}{\pgfpoint{36.608300pt}{33.033300pt}}
\pgfpathlineto{\pgfpoint{32.938200pt}{36.862800pt}}
\pgfpathcurveto{\pgfpoint{32.948600pt}{36.873800pt}}{\pgfpoint{32.958800pt}{36.884800pt}}{\pgfpoint{32.968900pt}{36.895800pt}}
\pgfpathcurveto{\pgfpoint{34.244200pt}{38.259300pt}}{\pgfpoint{35.462100pt}{39.105100pt}}{\pgfpoint{36.843100pt}{39.569300pt}}
\pgfpathlineto{\pgfpoint{38.436900pt}{34.530300pt}}
\pgfpathcurveto{\pgfpoint{37.855000pt}{34.279500pt}}{\pgfpoint{37.291900pt}{33.824900pt}}{\pgfpoint{36.638100pt}{33.067600pt}}
\pgfpathclose
\pgfusepath{fill}
\end{scope}
\definecolor{familyLogometa14}{HTML}{0866FF}
\begin{scope}[fill=familyLogometa14]
\pgfpathmoveto{\pgfpoint{36.619300pt}{33.029500pt}}
\pgfpathcurveto{\pgfpoint{35.518500pt}{31.745100pt}}{\pgfpoint{34.156600pt}{29.606600pt}}{\pgfpoint{32.014600pt}{26.148500pt}}
\pgfpathlineto{\pgfpoint{29.222900pt}{21.477400pt}}
\pgfpathlineto{\pgfpoint{29.200500pt}{21.439800pt}}
\pgfpathlineto{\pgfpoint{25.961400pt}{26.313400pt}}
\pgfpathlineto{\pgfpoint{27.962500pt}{29.690700pt}}
\pgfpathcurveto{\pgfpoint{29.879900pt}{32.909800pt}}{\pgfpoint{31.441900pt}{35.238500pt}}{\pgfpoint{32.949200pt}{36.859000pt}}
\pgfpathcurveto{\pgfpoint{32.959700pt}{36.870100pt}}{\pgfpoint{32.969800pt}{36.881000pt}}{\pgfpoint{32.980000pt}{36.892000pt}}
\pgfpathlineto{\pgfpoint{36.648800pt}{33.063800pt}}
\pgfpathcurveto{\pgfpoint{36.639000pt}{33.052500pt}}{\pgfpoint{36.629100pt}{33.041100pt}}{\pgfpoint{36.619300pt}{33.029500pt}}
\pgfpathclose
\pgfusepath{fill}
\end{scope}
}}
\definecolor{familyBaropenai}{HTML}{1f1f1f}
\newcommand{\providerLogoopenai}{\tikz[baseline=-.5ex,scale=0.229167,yscale=-1]{
\path[use as bounding box] (0pt,0pt) rectangle (48.0pt,48.0pt);
\begin{scope}[fill=white]
\fill (0pt,0pt) rectangle (48pt,48pt);
\end{scope}
\begin{scope}[fill=black]
\pgfpathmoveto{\pgfpoint{19.341800pt}{18.559900pt}}
\pgfpathlineto{\pgfpoint{19.341800pt}{14.759900pt}}
\pgfpathcurveto{\pgfpoint{19.341800pt}{14.439900pt}}{\pgfpoint{19.460800pt}{14.199800pt}}{\pgfpoint{19.738200pt}{14.040000pt}}
\pgfpathlineto{\pgfpoint{27.310200pt}{9.639970pt}}
\pgfpathcurveto{\pgfpoint{28.340900pt}{9.039990pt}}{\pgfpoint{29.569900pt}{8.760140pt}}{\pgfpoint{30.838300pt}{8.760140pt}}
\pgfpathcurveto{\pgfpoint{35.595400pt}{8.760140pt}}{\pgfpoint{38.608500pt}{12.480200pt}}{\pgfpoint{38.608500pt}{16.440100pt}}
\pgfpathcurveto{\pgfpoint{38.608500pt}{16.720000pt}}{\pgfpoint{38.608500pt}{17.040000pt}}{\pgfpoint{38.568700pt}{17.360100pt}}
\pgfpathlineto{\pgfpoint{30.719400pt}{12.720000pt}}
\pgfpathcurveto{\pgfpoint{30.243700pt}{12.440100pt}}{\pgfpoint{29.767800pt}{12.440100pt}}{\pgfpoint{29.292200pt}{12.720000pt}}
\pgfpathlineto{\pgfpoint{19.341800pt}{18.559900pt}}
\pgfpathclose
\pgfpathmoveto{\pgfpoint{37.022600pt}{33.360000pt}}
\pgfpathlineto{\pgfpoint{37.022600pt}{24.279900pt}}
\pgfpathcurveto{\pgfpoint{37.022600pt}{23.719700pt}}{\pgfpoint{36.784600pt}{23.319700pt}}{\pgfpoint{36.309000pt}{23.039800pt}}
\pgfpathlineto{\pgfpoint{26.358600pt}{17.199800pt}}
\pgfpathlineto{\pgfpoint{29.609300pt}{15.319700pt}}
\pgfpathcurveto{\pgfpoint{29.886800pt}{15.159900pt}}{\pgfpoint{30.124700pt}{15.159900pt}}{\pgfpoint{30.402200pt}{15.319700pt}}
\pgfpathlineto{\pgfpoint{37.974100pt}{19.719700pt}}
\pgfpathcurveto{\pgfpoint{40.154700pt}{20.999900pt}}{\pgfpoint{41.621300pt}{23.719700pt}}{\pgfpoint{41.621300pt}{26.359600pt}}
\pgfpathcurveto{\pgfpoint{41.621300pt}{29.399500pt}}{\pgfpoint{39.837500pt}{32.199900pt}}{\pgfpoint{37.022600pt}{33.360000pt}}
\pgfpathclose
\pgfpathmoveto{\pgfpoint{17.002900pt}{25.360100pt}}
\pgfpathlineto{\pgfpoint{13.752200pt}{23.440200pt}}
\pgfpathcurveto{\pgfpoint{13.474800pt}{23.280400pt}}{\pgfpoint{13.355700pt}{23.040200pt}}{\pgfpoint{13.355700pt}{22.720200pt}}
\pgfpathlineto{\pgfpoint{13.355700pt}{13.920300pt}}
\pgfpathcurveto{\pgfpoint{13.355700pt}{9.640390pt}}{\pgfpoint{16.606500pt}{6.400160pt}}{\pgfpoint{21.006900pt}{6.400160pt}}
\pgfpathcurveto{\pgfpoint{22.672200pt}{6.400160pt}}{\pgfpoint{24.217900pt}{6.960290pt}}{\pgfpoint{25.526500pt}{7.960250pt}}
\pgfpathlineto{\pgfpoint{17.716800pt}{12.520400pt}}
\pgfpathcurveto{\pgfpoint{17.241200pt}{12.800300pt}}{\pgfpoint{17.003300pt}{13.200200pt}}{\pgfpoint{17.003300pt}{13.760500pt}}
\pgfpathlineto{\pgfpoint{17.002900pt}{25.360100pt}}
\pgfpathclose
\pgfpathmoveto{\pgfpoint{24.000000pt}{29.440000pt}}
\pgfpathlineto{\pgfpoint{19.341800pt}{26.800100pt}}
\pgfpathlineto{\pgfpoint{19.341800pt}{21.200300pt}}
\pgfpathlineto{\pgfpoint{24.000000pt}{18.560400pt}}
\pgfpathlineto{\pgfpoint{28.657800pt}{21.200300pt}}
\pgfpathlineto{\pgfpoint{28.657800pt}{26.800100pt}}
\pgfpathlineto{\pgfpoint{24.000000pt}{29.440000pt}}
\pgfpathclose
\pgfpathmoveto{\pgfpoint{26.993000pt}{41.600200pt}}
\pgfpathcurveto{\pgfpoint{25.327800pt}{41.600200pt}}{\pgfpoint{23.782100pt}{41.040000pt}}{\pgfpoint{22.473500pt}{40.040200pt}}
\pgfpathlineto{\pgfpoint{30.283100pt}{35.479900pt}}
\pgfpathcurveto{\pgfpoint{30.758800pt}{35.200100pt}}{\pgfpoint{30.996700pt}{34.800100pt}}{\pgfpoint{30.996700pt}{34.239900pt}}
\pgfpathlineto{\pgfpoint{30.996700pt}{22.639900pt}}
\pgfpathlineto{\pgfpoint{34.287300pt}{24.559800pt}}
\pgfpathcurveto{\pgfpoint{34.564600pt}{24.719600pt}}{\pgfpoint{34.683700pt}{24.959700pt}}{\pgfpoint{34.683700pt}{25.279800pt}}
\pgfpathlineto{\pgfpoint{34.683700pt}{34.079700pt}}
\pgfpathcurveto{\pgfpoint{34.683700pt}{38.359600pt}}{\pgfpoint{31.393100pt}{41.600200pt}}{\pgfpoint{26.993000pt}{41.600200pt}}
\pgfpathclose
\pgfpathmoveto{\pgfpoint{17.597500pt}{32.680200pt}}
\pgfpathlineto{\pgfpoint{10.025500pt}{28.280300pt}}
\pgfpathcurveto{\pgfpoint{7.844930pt}{27.000100pt}}{\pgfpoint{6.378330pt}{24.280300pt}}{\pgfpoint{6.378330pt}{21.640400pt}}
\pgfpathcurveto{\pgfpoint{6.378330pt}{18.560400pt}}{\pgfpoint{8.201930pt}{15.800400pt}}{\pgfpoint{11.016400pt}{14.640300pt}}
\pgfpathlineto{\pgfpoint{11.016400pt}{23.760200pt}}
\pgfpathcurveto{\pgfpoint{11.016400pt}{24.320400pt}}{\pgfpoint{11.254400pt}{24.720400pt}}{\pgfpoint{11.730000pt}{25.000300pt}}
\pgfpathlineto{\pgfpoint{21.641000pt}{30.800100pt}}
\pgfpathlineto{\pgfpoint{18.390200pt}{32.680200pt}}
\pgfpathcurveto{\pgfpoint{18.112900pt}{32.840000pt}}{\pgfpoint{17.874900pt}{32.840000pt}}{\pgfpoint{17.597500pt}{32.680200pt}}
\pgfpathclose
\pgfpathmoveto{\pgfpoint{17.161700pt}{39.240200pt}}
\pgfpathcurveto{\pgfpoint{12.682000pt}{39.240200pt}}{\pgfpoint{9.391510pt}{35.840200pt}}{\pgfpoint{9.391510pt}{31.640100pt}}
\pgfpathcurveto{\pgfpoint{9.391510pt}{31.320100pt}}{\pgfpoint{9.431250pt}{31.000100pt}}{\pgfpoint{9.470660pt}{30.680000pt}}
\pgfpathlineto{\pgfpoint{17.280300pt}{35.240200pt}}
\pgfpathcurveto{\pgfpoint{17.755900pt}{35.520100pt}}{\pgfpoint{18.231900pt}{35.520100pt}}{\pgfpoint{18.707400pt}{35.240200pt}}
\pgfpathlineto{\pgfpoint{28.657800pt}{29.440400pt}}
\pgfpathlineto{\pgfpoint{28.657800pt}{33.240400pt}}
\pgfpathcurveto{\pgfpoint{28.657800pt}{33.560500pt}}{\pgfpoint{28.538800pt}{33.800500pt}}{\pgfpoint{28.261400pt}{33.960400pt}}
\pgfpathlineto{\pgfpoint{20.689400pt}{38.360400pt}}
\pgfpathcurveto{\pgfpoint{19.658600pt}{38.960300pt}}{\pgfpoint{18.430100pt}{39.240200pt}}{\pgfpoint{17.161700pt}{39.240200pt}}
\pgfpathclose
\pgfpathmoveto{\pgfpoint{26.993000pt}{44.000000pt}}
\pgfpathcurveto{\pgfpoint{31.789900pt}{44.000000pt}}{\pgfpoint{35.793600pt}{40.560100pt}}{\pgfpoint{36.705700pt}{36.000000pt}}
\pgfpathcurveto{\pgfpoint{41.145700pt}{34.839900pt}}{\pgfpoint{44.000000pt}{30.639900pt}}{\pgfpoint{44.000000pt}{26.360000pt}}
\pgfpathcurveto{\pgfpoint{44.000000pt}{23.559800pt}}{\pgfpoint{42.810800pt}{20.840100pt}}{\pgfpoint{40.670100pt}{18.880000pt}}
\pgfpathcurveto{\pgfpoint{40.868300pt}{18.039900pt}}{\pgfpoint{40.987200pt}{17.199800pt}}{\pgfpoint{40.987200pt}{16.360200pt}}
\pgfpathcurveto{\pgfpoint{40.987200pt}{10.640300pt}}{\pgfpoint{36.388500pt}{6.359980pt}}{\pgfpoint{31.076300pt}{6.359980pt}}
\pgfpathcurveto{\pgfpoint{30.006200pt}{6.359980pt}}{\pgfpoint{28.975400pt}{6.519790pt}}{\pgfpoint{27.944600pt}{6.880010pt}}
\pgfpathcurveto{\pgfpoint{26.160400pt}{5.119920pt}}{\pgfpoint{23.702500pt}{4.000000pt}}{\pgfpoint{21.006900pt}{4.000000pt}}
\pgfpathcurveto{\pgfpoint{16.210100pt}{4.000000pt}}{\pgfpoint{12.206400pt}{7.439800pt}}{\pgfpoint{11.294300pt}{12.000000pt}}
\pgfpathcurveto{\pgfpoint{6.854300pt}{13.160100pt}}{\pgfpoint{4.000000pt}{17.360100pt}}{\pgfpoint{4.000000pt}{21.639900pt}}
\pgfpathcurveto{\pgfpoint{4.000000pt}{24.440100pt}}{\pgfpoint{5.189160pt}{27.159900pt}}{\pgfpoint{7.329950pt}{29.119900pt}}
\pgfpathcurveto{\pgfpoint{7.131740pt}{29.960000pt}}{\pgfpoint{7.012770pt}{30.800100pt}}{\pgfpoint{7.012770pt}{31.639800pt}}
\pgfpathcurveto{\pgfpoint{7.012770pt}{37.359700pt}}{\pgfpoint{11.611400pt}{41.639900pt}}{\pgfpoint{16.923600pt}{41.639900pt}}
\pgfpathcurveto{\pgfpoint{17.993800pt}{41.639900pt}}{\pgfpoint{19.024600pt}{41.480100pt}}{\pgfpoint{20.055400pt}{41.119900pt}}
\pgfpathcurveto{\pgfpoint{21.839200pt}{42.880000pt}}{\pgfpoint{24.297100pt}{44.000000pt}}{\pgfpoint{26.993000pt}{44.000000pt}}
\pgfpathclose
\pgfusepath{fill}
\end{scope}
}}
\definecolor{familyBaranthropic}{HTML}{cc785c}
\newcommand{\providerLogoanthropic}{\tikz[baseline=-.5ex,scale=0.229167,yscale=-1]{
\path[use as bounding box] (0pt,0pt) rectangle (48.0pt,48.0pt);
\definecolor{familyLogoanthropic0}{HTML}{F0EFEA}
\begin{scope}[fill=familyLogoanthropic0]
\fill (0pt,0pt) rectangle (48pt,48pt);
\end{scope}
\begin{scope}[fill=black]
\pgfpathmoveto{\pgfpoint{32.840000pt}{10.000000pt}}
\pgfpathlineto{\pgfpoint{26.720000pt}{10.000000pt}}
\pgfpathlineto{\pgfpoint{37.880000pt}{38.000000pt}}
\pgfpathlineto{\pgfpoint{44.000000pt}{38.000000pt}}
\pgfpathlineto{\pgfpoint{32.840000pt}{10.000000pt}}
\pgfpathclose
\pgfusepath{fill}
\end{scope}
\begin{scope}[fill=black]
\pgfpathmoveto{\pgfpoint{15.160000pt}{10.000000pt}}
\pgfpathlineto{\pgfpoint{4.000000pt}{38.000000pt}}
\pgfpathlineto{\pgfpoint{10.240000pt}{38.000000pt}}
\pgfpathlineto{\pgfpoint{12.522400pt}{32.120000pt}}
\pgfpathlineto{\pgfpoint{24.197600pt}{32.120000pt}}
\pgfpathlineto{\pgfpoint{26.480000pt}{38.000000pt}}
\pgfpathlineto{\pgfpoint{32.720000pt}{38.000000pt}}
\pgfpathlineto{\pgfpoint{21.560000pt}{10.000000pt}}
\pgfpathlineto{\pgfpoint{15.160000pt}{10.000000pt}}
\pgfpathclose
\pgfpathmoveto{\pgfpoint{14.540800pt}{26.920000pt}}
\pgfpathlineto{\pgfpoint{18.360000pt}{17.080000pt}}
\pgfpathlineto{\pgfpoint{22.179300pt}{26.920000pt}}
\pgfpathlineto{\pgfpoint{14.540800pt}{26.920000pt}}
\pgfpathclose
\pgfusepath{fill}
\end{scope}
}}
\definecolor{familyBardeepseek}{HTML}{4D6BFE}
\newcommand{\providerLogodeepseek}{\tikz[baseline=-.5ex,scale=0.229167,yscale=-1]{
\path[use as bounding box] (0pt,0pt) rectangle (48.0pt,48.0pt);
\definecolor{familyLogodeepseek0}{HTML}{4D6BFE}
\begin{scope}[fill=familyLogodeepseek0]
\pgfpathmoveto{\pgfpoint{47.496400pt}{9.437570pt}}
\pgfpathcurveto{\pgfpoint{46.988000pt}{9.190720pt}}{\pgfpoint{46.769000pt}{9.661150pt}}{\pgfpoint{46.471700pt}{9.900130pt}}
\pgfpathcurveto{\pgfpoint{46.370000pt}{9.977220pt}}{\pgfpoint{46.284000pt}{10.077400pt}}{\pgfpoint{46.197900pt}{10.169900pt}}
\pgfpathcurveto{\pgfpoint{45.454800pt}{10.956200pt}}{\pgfpoint{44.586600pt}{11.472900pt}}{\pgfpoint{43.452300pt}{11.411200pt}}
\pgfpathcurveto{\pgfpoint{41.794000pt}{11.318600pt}}{\pgfpoint{40.378100pt}{11.835300pt}}{\pgfpoint{39.126500pt}{13.092000pt}}
\pgfpathcurveto{\pgfpoint{38.860500pt}{11.542300pt}}{\pgfpoint{37.976700pt}{10.617000pt}}{\pgfpoint{36.631200pt}{10.023400pt}}
\pgfpathcurveto{\pgfpoint{35.927300pt}{9.714970pt}}{\pgfpoint{35.215400pt}{9.406510pt}}{\pgfpoint{34.722600pt}{8.735700pt}}
\pgfpathcurveto{\pgfpoint{34.378400pt}{8.257830pt}}{\pgfpoint{34.284500pt}{7.725800pt}}{\pgfpoint{34.112400pt}{7.201470pt}}
\pgfpathcurveto{\pgfpoint{34.002900pt}{6.885400pt}}{\pgfpoint{33.893400pt}{6.561460pt}}{\pgfpoint{33.525700pt}{6.507470pt}}
\pgfpathcurveto{\pgfpoint{33.126800pt}{6.445870pt}}{\pgfpoint{32.970300pt}{6.777250pt}}{\pgfpoint{32.813800pt}{7.054990pt}}
\pgfpathcurveto{\pgfpoint{32.188100pt}{8.188440pt}}{\pgfpoint{31.945600pt}{9.437490pt}}{\pgfpoint{31.969000pt}{10.701900pt}}
\pgfpathcurveto{\pgfpoint{32.023700pt}{13.547100pt}}{\pgfpoint{33.236300pt}{15.813700pt}}{\pgfpoint{35.645500pt}{17.425200pt}}
\pgfpathcurveto{\pgfpoint{35.919300pt}{17.610300pt}}{\pgfpoint{35.989600pt}{17.795200pt}}{\pgfpoint{35.903600pt}{18.065200pt}}
\pgfpathcurveto{\pgfpoint{35.739300pt}{18.620400pt}}{\pgfpoint{35.543700pt}{19.160000pt}}{\pgfpoint{35.371700pt}{19.715100pt}}
\pgfpathcurveto{\pgfpoint{35.262200pt}{20.069900pt}}{\pgfpoint{35.097800pt}{20.147000pt}}{\pgfpoint{34.714600pt}{19.992600pt}}
\pgfpathcurveto{\pgfpoint{33.392600pt}{19.445300pt}}{\pgfpoint{32.250500pt}{18.635700pt}}{\pgfpoint{31.241400pt}{17.656600pt}}
\pgfpathcurveto{\pgfpoint{29.528400pt}{16.014300pt}}{\pgfpoint{27.979500pt}{14.202400pt}}{\pgfpoint{26.047500pt}{12.783700pt}}
\pgfpathcurveto{\pgfpoint{25.593800pt}{12.452100pt}}{\pgfpoint{25.140100pt}{12.143800pt}}{\pgfpoint{24.670700pt}{11.850700pt}}
\pgfpathcurveto{\pgfpoint{22.699400pt}{9.954120pt}}{\pgfpoint{24.928900pt}{8.396530pt}}{\pgfpoint{25.445100pt}{8.211630pt}}
\pgfpathcurveto{\pgfpoint{25.984800pt}{8.018850pt}}{\pgfpoint{25.632900pt}{7.355660pt}}{\pgfpoint{23.888400pt}{7.363530pt}}
\pgfpathcurveto{\pgfpoint{22.144000pt}{7.371140pt}}{\pgfpoint{20.548300pt}{7.949460pt}}{\pgfpoint{18.514600pt}{8.720470pt}}
\pgfpathcurveto{\pgfpoint{18.217300pt}{8.836150pt}}{\pgfpoint{17.904300pt}{8.920860pt}}{\pgfpoint{17.583600pt}{8.990250pt}}
\pgfpathcurveto{\pgfpoint{15.737600pt}{8.643380pt}}{\pgfpoint{13.821100pt}{8.566120pt}}{\pgfpoint{11.818600pt}{8.789860pt}}
\pgfpathcurveto{\pgfpoint{8.048190pt}{9.206130pt}}{\pgfpoint{5.036710pt}{10.971800pt}}{\pgfpoint{2.823000pt}{13.986500pt}}
\pgfpathcurveto{\pgfpoint{0.163470pt}{17.610300pt}}{\pgfpoint{-0.462297pt}{21.727400pt}}{\pgfpoint{0.304303pt}{26.022100pt}}
\pgfpathcurveto{\pgfpoint{1.110020pt}{30.547800pt}}{\pgfpoint{3.440990pt}{34.295000pt}}{\pgfpoint{7.023670pt}{37.225000pt}}
\pgfpathcurveto{\pgfpoint{10.739300pt}{40.262600pt}}{\pgfpoint{15.018100pt}{41.750700pt}}{\pgfpoint{19.899200pt}{41.465500pt}}
\pgfpathcurveto{\pgfpoint{22.863900pt}{41.295900pt}}{\pgfpoint{26.164900pt}{40.902700pt}}{\pgfpoint{29.888400pt}{37.780100pt}}
\pgfpathcurveto{\pgfpoint{30.827000pt}{38.242700pt}}{\pgfpoint{31.812600pt}{38.427600pt}}{\pgfpoint{33.447500pt}{38.566400pt}}
\pgfpathcurveto{\pgfpoint{34.706900pt}{38.682100pt}}{\pgfpoint{35.919400pt}{38.504800pt}}{\pgfpoint{36.858000pt}{38.312100pt}}
\pgfpathcurveto{\pgfpoint{38.328600pt}{38.003700pt}}{\pgfpoint{38.227000pt}{36.654400pt}}{\pgfpoint{37.695100pt}{36.407700pt}}
\pgfpathcurveto{\pgfpoint{33.385000pt}{34.418500pt}}{\pgfpoint{34.331400pt}{35.228000pt}}{\pgfpoint{33.471000pt}{34.572600pt}}
\pgfpathcurveto{\pgfpoint{35.661300pt}{32.005200pt}}{\pgfpoint{38.962400pt}{29.337400pt}}{\pgfpoint{40.253000pt}{20.694200pt}}
\pgfpathcurveto{\pgfpoint{40.354600pt}{20.008100pt}}{\pgfpoint{40.268600pt}{19.576300pt}}{\pgfpoint{40.253000pt}{19.021100pt}}
\pgfpathcurveto{\pgfpoint{40.245100pt}{18.682000pt}}{\pgfpoint{40.323400pt}{18.551000pt}}{\pgfpoint{40.714500pt}{18.512300pt}}
\pgfpathcurveto{\pgfpoint{41.794000pt}{18.388900pt}}{\pgfpoint{42.842200pt}{18.096000pt}}{\pgfpoint{43.804300pt}{17.571700pt}}
\pgfpathcurveto{\pgfpoint{46.596900pt}{16.060600pt}}{\pgfpoint{47.723300pt}{13.577900pt}}{\pgfpoint{47.989200pt}{10.601700pt}}
\pgfpathcurveto{\pgfpoint{48.028400pt}{10.146800pt}}{\pgfpoint{47.981400pt}{9.676380pt}}{\pgfpoint{47.496400pt}{9.437570pt}}
\pgfpathclose
\pgfpathmoveto{\pgfpoint{23.161100pt}{36.222500pt}}
\pgfpathcurveto{\pgfpoint{18.984000pt}{32.969000pt}}{\pgfpoint{16.958100pt}{31.897100pt}}{\pgfpoint{16.121100pt}{31.943400pt}}
\pgfpathcurveto{\pgfpoint{15.338800pt}{31.989700pt}}{\pgfpoint{15.479700pt}{32.876400pt}}{\pgfpoint{15.651700pt}{33.454700pt}}
\pgfpathcurveto{\pgfpoint{15.831600pt}{34.025200pt}}{\pgfpoint{16.066300pt}{34.418500pt}}{\pgfpoint{16.394800pt}{34.919600pt}}
\pgfpathcurveto{\pgfpoint{16.621600pt}{35.251100pt}}{\pgfpoint{16.778200pt}{35.744500pt}}{\pgfpoint{16.168000pt}{36.114700pt}}
\pgfpathcurveto{\pgfpoint{14.822600pt}{36.939700pt}}{\pgfpoint{12.483600pt}{35.837100pt}}{\pgfpoint{12.374100pt}{35.783200pt}}
\pgfpathcurveto{\pgfpoint{9.652020pt}{34.194800pt}}{\pgfpoint{7.375710pt}{32.097500pt}}{\pgfpoint{5.772050pt}{29.229500pt}}
\pgfpathcurveto{\pgfpoint{4.223220pt}{26.469100pt}}{\pgfpoint{3.323730pt}{23.508600pt}}{\pgfpoint{3.175040pt}{20.347400pt}}
\pgfpathcurveto{\pgfpoint{3.135920pt}{19.584100pt}}{\pgfpoint{3.362760pt}{19.314100pt}}{\pgfpoint{4.129360pt}{19.175500pt}}
\pgfpathcurveto{\pgfpoint{5.138430pt}{18.990400pt}}{\pgfpoint{6.178840pt}{18.951700pt}}{\pgfpoint{7.187900pt}{19.098200pt}}
\pgfpathcurveto{\pgfpoint{11.451200pt}{19.715100pt}}{\pgfpoint{15.080700pt}{21.604100pt}}{\pgfpoint{18.123500pt}{24.595600pt}}
\pgfpathcurveto{\pgfpoint{19.860000pt}{26.299500pt}}{\pgfpoint{21.174200pt}{28.335200pt}}{\pgfpoint{22.527500pt}{30.324300pt}}
\pgfpathcurveto{\pgfpoint{23.966800pt}{32.436800pt}}{\pgfpoint{25.515600pt}{34.449300pt}}{\pgfpoint{27.486800pt}{36.099200pt}}
\pgfpathcurveto{\pgfpoint{28.183000pt}{36.677500pt}}{\pgfpoint{28.738400pt}{37.116900pt}}{\pgfpoint{29.270300pt}{37.440800pt}}
\pgfpathcurveto{\pgfpoint{27.666800pt}{37.618100pt}}{\pgfpoint{24.991500pt}{37.656800pt}}{\pgfpoint{23.161100pt}{36.222500pt}}
\pgfpathclose
\pgfpathmoveto{\pgfpoint{25.163600pt}{23.462300pt}}
\pgfpathcurveto{\pgfpoint{25.163600pt}{23.123000pt}}{\pgfpoint{25.437400pt}{22.853200pt}}{\pgfpoint{25.781700pt}{22.853200pt}}
\pgfpathcurveto{\pgfpoint{25.859900pt}{22.853200pt}}{\pgfpoint{25.930300pt}{22.868500pt}}{\pgfpoint{25.992900pt}{22.891600pt}}
\pgfpathcurveto{\pgfpoint{26.079000pt}{22.922500pt}}{\pgfpoint{26.157200pt}{22.968800pt}}{\pgfpoint{26.219800pt}{23.038000pt}}
\pgfpathcurveto{\pgfpoint{26.329300pt}{23.146100pt}}{\pgfpoint{26.391900pt}{23.300200pt}}{\pgfpoint{26.391900pt}{23.462200pt}}
\pgfpathcurveto{\pgfpoint{26.391900pt}{23.801400pt}}{\pgfpoint{26.118100pt}{24.071200pt}}{\pgfpoint{25.774000pt}{24.071200pt}}
\pgfpathcurveto{\pgfpoint{25.429900pt}{24.071200pt}}{\pgfpoint{25.163600pt}{23.801500pt}}{\pgfpoint{25.163600pt}{23.462300pt}}
\pgfpathclose
\pgfpathmoveto{\pgfpoint{31.382400pt}{26.623500pt}}
\pgfpathcurveto{\pgfpoint{30.983500pt}{26.785400pt}}{\pgfpoint{30.584600pt}{26.924100pt}}{\pgfpoint{30.201300pt}{26.939500pt}}
\pgfpathcurveto{\pgfpoint{29.606800pt}{26.970300pt}}{\pgfpoint{28.957500pt}{26.731300pt}}{\pgfpoint{28.605600pt}{26.438300pt}}
\pgfpathcurveto{\pgfpoint{28.058000pt}{25.983400pt}}{\pgfpoint{27.666900pt}{25.729100pt}}{\pgfpoint{27.502600pt}{24.934900pt}}
\pgfpathcurveto{\pgfpoint{27.432200pt}{24.595600pt}}{\pgfpoint{27.471300pt}{24.071300pt}}{\pgfpoint{27.534000pt}{23.770700pt}}
\pgfpathcurveto{\pgfpoint{27.674700pt}{23.123000pt}}{\pgfpoint{27.518300pt}{22.706700pt}}{\pgfpoint{27.056800pt}{22.328800pt}}
\pgfpathcurveto{\pgfpoint{26.681300pt}{22.020400pt}}{\pgfpoint{26.204100pt}{21.935700pt}}{\pgfpoint{25.680000pt}{21.935700pt}}
\pgfpathcurveto{\pgfpoint{25.484400pt}{21.935700pt}}{\pgfpoint{25.304600pt}{21.850800pt}}{\pgfpoint{25.171600pt}{21.781400pt}}
\pgfpathcurveto{\pgfpoint{24.952500pt}{21.673500pt}}{\pgfpoint{24.772700pt}{21.403600pt}}{\pgfpoint{24.944700pt}{21.072100pt}}
\pgfpathcurveto{\pgfpoint{24.999500pt}{20.964300pt}}{\pgfpoint{25.265500pt}{20.702000pt}}{\pgfpoint{25.328100pt}{20.655800pt}}
\pgfpathcurveto{\pgfpoint{26.040000pt}{20.254800pt}}{\pgfpoint{26.861300pt}{20.385900pt}}{\pgfpoint{27.620100pt}{20.686600pt}}
\pgfpathcurveto{\pgfpoint{28.324200pt}{20.971900pt}}{\pgfpoint{28.856100pt}{21.496200pt}}{\pgfpoint{29.622600pt}{22.236300pt}}
\pgfpathcurveto{\pgfpoint{30.404800pt}{23.130600pt}}{\pgfpoint{30.545600pt}{23.377600pt}}{\pgfpoint{30.991500pt}{24.048200pt}}
\pgfpathcurveto{\pgfpoint{31.343600pt}{24.572500pt}}{\pgfpoint{31.664300pt}{25.112200pt}}{\pgfpoint{31.883300pt}{25.729100pt}}
\pgfpathcurveto{\pgfpoint{32.016200pt}{26.114600pt}}{\pgfpoint{31.844000pt}{26.430700pt}}{\pgfpoint{31.382400pt}{26.623500pt}}
\pgfpathclose
\pgfusepath{fill}
\end{scope}
}}
\definecolor{familyBestOpen}{HTML}{0072B2}
\definecolor{familyBestPro}{HTML}{D55E00}

\definecolor{effColor1}{HTML}{ff7018}
\pgfdeclareplotmark{effM1}{\pgfpathcircle{\pgfpointorigin}{\pgfplotmarksize}\pgfusepath{fill,stroke}}
\definecolor{effColor2}{HTML}{ff7018}
\pgfdeclareplotmark{effM2}{\pgfpathmoveto{\pgfpoint{-0.80000\pgfplotmarksize}{-0.80000\pgfplotmarksize}}\pgfpathlineto{\pgfpoint{0.80000\pgfplotmarksize}{-0.80000\pgfplotmarksize}}\pgfpathlineto{\pgfpoint{0.80000\pgfplotmarksize}{0.80000\pgfplotmarksize}}\pgfpathlineto{\pgfpoint{-0.80000\pgfplotmarksize}{0.80000\pgfplotmarksize}}\pgfpathclose\pgfusepath{fill,stroke}}
\definecolor{effColor3}{HTML}{ff7018}
\pgfdeclareplotmark{effM3}{\pgfpathmoveto{\pgfpoint{0.00000\pgfplotmarksize}{1.00000\pgfplotmarksize}}\pgfpathlineto{\pgfpoint{-0.86600\pgfplotmarksize}{-0.50000\pgfplotmarksize}}\pgfpathlineto{\pgfpoint{0.86600\pgfplotmarksize}{-0.50000\pgfplotmarksize}}\pgfpathclose\pgfusepath{fill,stroke}}
\definecolor{effColor4}{HTML}{ff7018}
\pgfdeclareplotmark{effM4}{\pgfpathmoveto{\pgfpoint{0.00000\pgfplotmarksize}{1.00000\pgfplotmarksize}}\pgfpathlineto{\pgfpoint{-1.00000\pgfplotmarksize}{0.00000\pgfplotmarksize}}\pgfpathlineto{\pgfpoint{0.00000\pgfplotmarksize}{-1.00000\pgfplotmarksize}}\pgfpathlineto{\pgfpoint{1.00000\pgfplotmarksize}{0.00000\pgfplotmarksize}}\pgfpathclose\pgfusepath{fill,stroke}}
\definecolor{effColor5}{HTML}{ff7018}
\pgfdeclareplotmark{effM5}{\pgfpathmoveto{\pgfpoint{0.00000\pgfplotmarksize}{1.00000\pgfplotmarksize}}\pgfpathlineto{\pgfpoint{-0.95106\pgfplotmarksize}{0.30902\pgfplotmarksize}}\pgfpathlineto{\pgfpoint{-0.58779\pgfplotmarksize}{-0.80902\pgfplotmarksize}}\pgfpathlineto{\pgfpoint{0.58779\pgfplotmarksize}{-0.80902\pgfplotmarksize}}\pgfpathlineto{\pgfpoint{0.95106\pgfplotmarksize}{0.30902\pgfplotmarksize}}\pgfpathclose\pgfusepath{fill,stroke}}
\definecolor{effColor6}{HTML}{ff7018}
\pgfdeclareplotmark{effM6}{\pgfpathmoveto{\pgfpoint{-1.00000\pgfplotmarksize}{-0.50000\pgfplotmarksize}}\pgfpathlineto{\pgfpoint{1.00000\pgfplotmarksize}{-0.50000\pgfplotmarksize}}\pgfpathlineto{\pgfpoint{1.00000\pgfplotmarksize}{0.50000\pgfplotmarksize}}\pgfpathlineto{\pgfpoint{-1.00000\pgfplotmarksize}{0.50000\pgfplotmarksize}}\pgfpathclose\pgfusepath{fill,stroke}}
\definecolor{effColor9}{HTML}{34A853}
\pgfdeclareplotmark{effM9}{\pgfpathcircle{\pgfpointorigin}{\pgfplotmarksize}\pgfusepath{fill,stroke}}
\definecolor{effColor10}{HTML}{34A853}
\pgfdeclareplotmark{effM10}{\pgfpathmoveto{\pgfpoint{-0.80000\pgfplotmarksize}{-0.80000\pgfplotmarksize}}\pgfpathlineto{\pgfpoint{0.80000\pgfplotmarksize}{-0.80000\pgfplotmarksize}}\pgfpathlineto{\pgfpoint{0.80000\pgfplotmarksize}{0.80000\pgfplotmarksize}}\pgfpathlineto{\pgfpoint{-0.80000\pgfplotmarksize}{0.80000\pgfplotmarksize}}\pgfpathclose\pgfusepath{fill,stroke}}
\definecolor{effColor11}{HTML}{34A853}
\pgfdeclareplotmark{effM11}{\pgfpathmoveto{\pgfpoint{0.00000\pgfplotmarksize}{1.00000\pgfplotmarksize}}\pgfpathlineto{\pgfpoint{-0.86600\pgfplotmarksize}{-0.50000\pgfplotmarksize}}\pgfpathlineto{\pgfpoint{0.86600\pgfplotmarksize}{-0.50000\pgfplotmarksize}}\pgfpathclose\pgfusepath{fill,stroke}}
\definecolor{effColor12}{HTML}{34A853}
\pgfdeclareplotmark{effM12}{\pgfpathmoveto{\pgfpoint{0.00000\pgfplotmarksize}{1.00000\pgfplotmarksize}}\pgfpathlineto{\pgfpoint{-1.00000\pgfplotmarksize}{0.00000\pgfplotmarksize}}\pgfpathlineto{\pgfpoint{0.00000\pgfplotmarksize}{-1.00000\pgfplotmarksize}}\pgfpathlineto{\pgfpoint{1.00000\pgfplotmarksize}{0.00000\pgfplotmarksize}}\pgfpathclose\pgfusepath{fill,stroke}}
\definecolor{effColor14}{HTML}{34A853}
\pgfdeclareplotmark{effM14}{\pgfpathmoveto{\pgfpoint{-1.00000\pgfplotmarksize}{-0.50000\pgfplotmarksize}}\pgfpathlineto{\pgfpoint{1.00000\pgfplotmarksize}{-0.50000\pgfplotmarksize}}\pgfpathlineto{\pgfpoint{1.00000\pgfplotmarksize}{0.50000\pgfplotmarksize}}\pgfpathlineto{\pgfpoint{-1.00000\pgfplotmarksize}{0.50000\pgfplotmarksize}}\pgfpathclose\pgfusepath{fill,stroke}}
\definecolor{effColor15}{HTML}{34A853}
\pgfdeclareplotmark{effM15}{\pgfpathmoveto{\pgfpoint{-0.50000\pgfplotmarksize}{-1.00000\pgfplotmarksize}}\pgfpathlineto{\pgfpoint{0.50000\pgfplotmarksize}{-1.00000\pgfplotmarksize}}\pgfpathlineto{\pgfpoint{0.50000\pgfplotmarksize}{1.00000\pgfplotmarksize}}\pgfpathlineto{\pgfpoint{-0.50000\pgfplotmarksize}{1.00000\pgfplotmarksize}}\pgfpathclose\pgfusepath{fill,stroke}}
\definecolor{effColor16}{HTML}{34A853}
\pgfdeclareplotmark{effM16}{\pgfpathmoveto{\pgfpoint{0.00000\pgfplotmarksize}{1.00000\pgfplotmarksize}}\pgfpathlineto{\pgfpoint{-0.86603\pgfplotmarksize}{0.50000\pgfplotmarksize}}\pgfpathlineto{\pgfpoint{-0.86603\pgfplotmarksize}{-0.50000\pgfplotmarksize}}\pgfpathlineto{\pgfpoint{-0.00000\pgfplotmarksize}{-1.00000\pgfplotmarksize}}\pgfpathlineto{\pgfpoint{0.86603\pgfplotmarksize}{-0.50000\pgfplotmarksize}}\pgfpathlineto{\pgfpoint{0.86603\pgfplotmarksize}{0.50000\pgfplotmarksize}}\pgfpathclose\pgfusepath{fill,stroke}}
\definecolor{effColor17}{HTML}{34A853}
\pgfdeclareplotmark{effM17}{\pgfpathmoveto{\pgfpoint{0.00000\pgfplotmarksize}{1.00000\pgfplotmarksize}}\pgfpathlineto{\pgfpoint{-0.70711\pgfplotmarksize}{0.70711\pgfplotmarksize}}\pgfpathlineto{\pgfpoint{-1.00000\pgfplotmarksize}{0.00000\pgfplotmarksize}}\pgfpathlineto{\pgfpoint{-0.70711\pgfplotmarksize}{-0.70711\pgfplotmarksize}}\pgfpathlineto{\pgfpoint{-0.00000\pgfplotmarksize}{-1.00000\pgfplotmarksize}}\pgfpathlineto{\pgfpoint{0.70711\pgfplotmarksize}{-0.70711\pgfplotmarksize}}\pgfpathlineto{\pgfpoint{1.00000\pgfplotmarksize}{-0.00000\pgfplotmarksize}}\pgfpathlineto{\pgfpoint{0.70711\pgfplotmarksize}{0.70711\pgfplotmarksize}}\pgfpathclose\pgfusepath{fill,stroke}}
\definecolor{effColor19}{HTML}{f32e02}
\pgfdeclareplotmark{effM19}{\pgfpathcircle{\pgfpointorigin}{\pgfplotmarksize}\pgfusepath{fill,stroke}}
\definecolor{effColor20}{HTML}{f32e02}
\pgfdeclareplotmark{effM20}{\pgfpathmoveto{\pgfpoint{-0.80000\pgfplotmarksize}{-0.80000\pgfplotmarksize}}\pgfpathlineto{\pgfpoint{0.80000\pgfplotmarksize}{-0.80000\pgfplotmarksize}}\pgfpathlineto{\pgfpoint{0.80000\pgfplotmarksize}{0.80000\pgfplotmarksize}}\pgfpathlineto{\pgfpoint{-0.80000\pgfplotmarksize}{0.80000\pgfplotmarksize}}\pgfpathclose\pgfusepath{fill,stroke}}
\definecolor{effColor21}{HTML}{f32e02}
\pgfdeclareplotmark{effM21}{\pgfpathmoveto{\pgfpoint{0.00000\pgfplotmarksize}{1.00000\pgfplotmarksize}}\pgfpathlineto{\pgfpoint{-0.86600\pgfplotmarksize}{-0.50000\pgfplotmarksize}}\pgfpathlineto{\pgfpoint{0.86600\pgfplotmarksize}{-0.50000\pgfplotmarksize}}\pgfpathclose\pgfusepath{fill,stroke}}
\definecolor{effColor22}{HTML}{0078D5}
\pgfdeclareplotmark{effM22}{\pgfpathcircle{\pgfpointorigin}{\pgfplotmarksize}\pgfusepath{fill,stroke}}
\definecolor{effColor23}{HTML}{0078D5}
\pgfdeclareplotmark{effM23}{\pgfpathmoveto{\pgfpoint{-0.80000\pgfplotmarksize}{-0.80000\pgfplotmarksize}}\pgfpathlineto{\pgfpoint{0.80000\pgfplotmarksize}{-0.80000\pgfplotmarksize}}\pgfpathlineto{\pgfpoint{0.80000\pgfplotmarksize}{0.80000\pgfplotmarksize}}\pgfpathlineto{\pgfpoint{-0.80000\pgfplotmarksize}{0.80000\pgfplotmarksize}}\pgfpathclose\pgfusepath{fill,stroke}}
\definecolor{effColor30}{HTML}{0089f4}
\pgfdeclareplotmark{effM30}{\pgfpathmoveto{\pgfpoint{0.00000\pgfplotmarksize}{1.00000\pgfplotmarksize}}\pgfpathlineto{\pgfpoint{-0.86600\pgfplotmarksize}{-0.50000\pgfplotmarksize}}\pgfpathlineto{\pgfpoint{0.86600\pgfplotmarksize}{-0.50000\pgfplotmarksize}}\pgfpathclose\pgfusepath{fill,stroke}}
\definecolor{effColor31}{HTML}{0089f4}
\pgfdeclareplotmark{effM31}{\pgfpathmoveto{\pgfpoint{0.00000\pgfplotmarksize}{1.00000\pgfplotmarksize}}\pgfpathlineto{\pgfpoint{-1.00000\pgfplotmarksize}{0.00000\pgfplotmarksize}}\pgfpathlineto{\pgfpoint{0.00000\pgfplotmarksize}{-1.00000\pgfplotmarksize}}\pgfpathlineto{\pgfpoint{1.00000\pgfplotmarksize}{0.00000\pgfplotmarksize}}\pgfpathclose\pgfusepath{fill,stroke}}
\definecolor{effColor35}{HTML}{1f1f1f}
\pgfdeclareplotmark{effM35}{\pgfpathcircle{\pgfpointorigin}{\pgfplotmarksize}\pgfusepath{fill,stroke}}
\definecolor{effColor36}{HTML}{1f1f1f}
\pgfdeclareplotmark{effM36}{\pgfpathmoveto{\pgfpoint{-0.80000\pgfplotmarksize}{-0.80000\pgfplotmarksize}}\pgfpathlineto{\pgfpoint{0.80000\pgfplotmarksize}{-0.80000\pgfplotmarksize}}\pgfpathlineto{\pgfpoint{0.80000\pgfplotmarksize}{0.80000\pgfplotmarksize}}\pgfpathlineto{\pgfpoint{-0.80000\pgfplotmarksize}{0.80000\pgfplotmarksize}}\pgfpathclose\pgfusepath{fill,stroke}}
\definecolor{effColor37}{HTML}{1f1f1f}
\pgfdeclareplotmark{effM37}{\pgfpathmoveto{\pgfpoint{0.00000\pgfplotmarksize}{1.00000\pgfplotmarksize}}\pgfpathlineto{\pgfpoint{-0.86600\pgfplotmarksize}{-0.50000\pgfplotmarksize}}\pgfpathlineto{\pgfpoint{0.86600\pgfplotmarksize}{-0.50000\pgfplotmarksize}}\pgfpathclose\pgfusepath{fill,stroke}}
\definecolor{effColor38}{HTML}{cc785c}
\pgfdeclareplotmark{effM38}{\pgfpathcircle{\pgfpointorigin}{\pgfplotmarksize}\pgfusepath{fill,stroke}}
\definecolor{effColor39}{HTML}{cc785c}
\pgfdeclareplotmark{effM39}{\pgfpathmoveto{\pgfpoint{-0.80000\pgfplotmarksize}{-0.80000\pgfplotmarksize}}\pgfpathlineto{\pgfpoint{0.80000\pgfplotmarksize}{-0.80000\pgfplotmarksize}}\pgfpathlineto{\pgfpoint{0.80000\pgfplotmarksize}{0.80000\pgfplotmarksize}}\pgfpathlineto{\pgfpoint{-0.80000\pgfplotmarksize}{0.80000\pgfplotmarksize}}\pgfpathclose\pgfusepath{fill,stroke}}
\definecolor{effColor40}{HTML}{cc785c}
\pgfdeclareplotmark{effM40}{\pgfpathmoveto{\pgfpoint{0.00000\pgfplotmarksize}{1.00000\pgfplotmarksize}}\pgfpathlineto{\pgfpoint{-0.86600\pgfplotmarksize}{-0.50000\pgfplotmarksize}}\pgfpathlineto{\pgfpoint{0.86600\pgfplotmarksize}{-0.50000\pgfplotmarksize}}\pgfpathclose\pgfusepath{fill,stroke}}
\definecolor{effColor41}{HTML}{cc785c}
\pgfdeclareplotmark{effM41}{\pgfpathmoveto{\pgfpoint{0.00000\pgfplotmarksize}{1.00000\pgfplotmarksize}}\pgfpathlineto{\pgfpoint{-1.00000\pgfplotmarksize}{0.00000\pgfplotmarksize}}\pgfpathlineto{\pgfpoint{0.00000\pgfplotmarksize}{-1.00000\pgfplotmarksize}}\pgfpathlineto{\pgfpoint{1.00000\pgfplotmarksize}{0.00000\pgfplotmarksize}}\pgfpathclose\pgfusepath{fill,stroke}}
\definecolor{effColor42}{HTML}{34A853}
\pgfdeclareplotmark{effM42}{\pgfpathmoveto{\pgfpoint{-0.80000\pgfplotmarksize}{-0.80000\pgfplotmarksize}}\pgfpathlineto{\pgfpoint{0.80000\pgfplotmarksize}{-0.80000\pgfplotmarksize}}\pgfpathlineto{\pgfpoint{0.80000\pgfplotmarksize}{0.80000\pgfplotmarksize}}\pgfpathlineto{\pgfpoint{-0.80000\pgfplotmarksize}{0.80000\pgfplotmarksize}}\pgfpathclose\pgfusepath{fill,stroke}\pgfpathmoveto{\pgfpoint{-0.45000\pgfplotmarksize}{0.00000\pgfplotmarksize}}\pgfpathlineto{\pgfpoint{0.45000\pgfplotmarksize}{0.00000\pgfplotmarksize}}\pgfpathmoveto{\pgfpoint{0.00000\pgfplotmarksize}{-0.45000\pgfplotmarksize}}\pgfpathlineto{\pgfpoint{0.00000\pgfplotmarksize}{0.45000\pgfplotmarksize}}\pgfusepath{stroke}}
\definecolor{effColor43}{HTML}{34A853}
\pgfdeclareplotmark{effM43}{\pgfpathcircle{\pgfpointorigin}{\pgfplotmarksize}\pgfusepath{fill,stroke}\pgfpathmoveto{\pgfpoint{-0.45000\pgfplotmarksize}{0.00000\pgfplotmarksize}}\pgfpathlineto{\pgfpoint{0.45000\pgfplotmarksize}{0.00000\pgfplotmarksize}}\pgfpathmoveto{\pgfpoint{0.00000\pgfplotmarksize}{-0.45000\pgfplotmarksize}}\pgfpathlineto{\pgfpoint{0.00000\pgfplotmarksize}{0.45000\pgfplotmarksize}}\pgfusepath{stroke}}
\definecolor{effColor44}{HTML}{34A853}
\pgfdeclareplotmark{effM44}{\pgfpathmoveto{\pgfpoint{-0.80000\pgfplotmarksize}{-0.80000\pgfplotmarksize}}\pgfpathlineto{\pgfpoint{0.80000\pgfplotmarksize}{-0.80000\pgfplotmarksize}}\pgfpathlineto{\pgfpoint{0.80000\pgfplotmarksize}{0.80000\pgfplotmarksize}}\pgfpathlineto{\pgfpoint{-0.80000\pgfplotmarksize}{0.80000\pgfplotmarksize}}\pgfpathclose\pgfusepath{fill,stroke}\pgfpathcircle{\pgfpointorigin}{.28\pgfplotmarksize}\pgfusepath{stroke}}

\begin{document}
\maketitle
\begin{abstract}

Financial disclosures express the underlying relationships of double-entry accounting through unstructured text. While financial LLM benchmarks predominantly emphasize question answering and classification, schema-guided evidence discovery remains relatively under-explored. We investigate whether open-weight LLMs can extract and classify evidence corresponding to predefined double-entry accounting relationships underlying the text. Using a financial accounting graph comprising 11 nodes (accounts) and 11 edges (transaction types), we evaluate mid-sized models from 5 open-weight families across 1,300 SEC 10-K Item 7 filings spanning 26 years, comparing their extraction performance against three proprietary models. Results reveal substantial differences in extraction volume, edge distributions, and cross-model agreement. Building on this study, we introduce a human-annotated relation classification benchmark comprising 483 passages across three financial accounting categories, evaluating 34 open-weight and 10 proprietary models. The strongest open-weight and proprietary models achieve 84.1\% and 92.8\% accuracy, respectively, with considerable variation across accounting categories. Our findings highlight the importance of evaluating evidence discovery and relation understanding separately, while investigating locally deployable models for scalable, privacy-sensitive financial analysis.

\end{abstract}

\section{Introduction}

\suppressfloats[t]
\begin{figure}[!t]
\centering
\resizebox{\columnwidth}{!}{%
\definecolor{pipelineBlue}{HTML}{2496ED}%
\definecolor{pipelineOrange}{HTML}{FF9F1C}%
\definecolor{pipelineViolet}{HTML}{9B5DE5}%
\definecolor{pipelineGreen}{HTML}{57CC99}%
\definecolor{pipelinePink}{HTML}{F26B9B}%
\definecolor{pipelineTeal}{HTML}{00B4D8}%
\begin{tikzpicture}[
  stage/.style={rounded corners=3pt,line width=0.95pt,
    text width=2.1cm,minimum height=1.85cm,align=center,
    font=\fontsize{8}{9.2}\selectfont,inner sep=3pt},
  stageLabel/.style={align=center,text width=2.12cm,anchor=north,
    font=\fontsize{7.6}{8.5}\selectfont,inner sep=0pt},
  link/.style={-{Stealth[length=1.6mm]},draw=black!80,line width=0.75pt}]
\draw[draw=pipelineBlue!60!black,dashed,line width=1pt,
  rounded corners=5pt,fill=yellow!12!white] (-1.42,-3.55) rectangle (6.72,1.20);
\node[stage,draw=pipelineBlue!65!black,fill=pipelineBlue!10!white] (docs) at (0,0) {};
\node[stage,draw=pipelineOrange!65!black,fill=pipelineOrange!12!white] (extract) at (2.65,0) {};
\node[stage,draw=pipelineViolet!65!black,fill=pipelineViolet!10!white] (compare) at (5.3,0) {};
\node[stage,draw=pipelineGreen!60!black,fill=pipelineGreen!12!white] (pool) at (0,-2.35) {};
\node[stage,draw=pipelinePink!65!black,fill=pipelinePink!10!white] (gold) at (2.65,-2.35) {};
\node[stage,draw=pipelineTeal!65!black,fill=pipelineTeal!10!white] (classify) at (5.3,-2.35) {};
\foreach \x/\y/\accent in {0/0/pipelineBlue,0/-2.35/pipelineGreen}{
  \begin{scope}[shift={(\x,\y)},draw=black,line width=0.8pt,line join=round]
    \draw[fill=\accent!40!white] (-0.19,0.28) -- (0.19,0.28) -- (0.19,0.61)
      -- (0.07,0.73) -- (-0.19,0.73) -- cycle;
    \draw[fill=\accent] (0.07,0.73) -- (0.07,0.61) -- (0.19,0.61) -- cycle;
    \draw (-0.11,0.54) -- (0.10,0.54);
    \draw (-0.11,0.46) -- (0.10,0.46);
    \draw (-0.11,0.38) -- (0.04,0.38);
  \end{scope}
}
\begin{scope}[shift={(2.65,0)},draw=black,line width=0.8pt]
  \draw[fill=pipelineOrange] (-0.05,0.55) circle (0.19);
  \draw[fill=white,line width=0.55pt] (-0.05,0.55) circle (0.14);
  \draw[fill=pipelineOrange,rounded corners=0.4pt]
    (0.06,0.40) -- (0.10,0.44) -- (0.25,0.29) -- (0.21,0.25) -- cycle;
  \draw[line width=0.5pt] (-0.13,0.54) -- (-0.02,0.62) -- (0.05,0.50);
  \foreach \p in {(-0.13,0.54),(-0.02,0.62),(0.05,0.50)}
    \fill \p circle (0.025);
\end{scope}
\foreach \x/\y/\accent in {5.3/0/pipelineViolet,5.3/-2.35/pipelineTeal}{
  \begin{scope}[shift={(\x,\y)},draw=black,line join=round]
    \draw[fill=\accent,line width=0.8pt]
      (0.2450,0.5100) -- (0.2403,0.5578) -- (0.1663,0.5789) -- (0.1497,0.6100) -- (0.1732,0.6832) -- (0.1361,0.7137) -- (0.0689,0.6763) -- (0.0351,0.6865) -- (0.0000,0.7550) -- (-0.0478,0.7503) -- (-0.0689,0.6763) -- (-0.1000,0.6597) -- (-0.1732,0.6832) -- (-0.2037,0.6461) -- (-0.1663,0.5789) -- (-0.1765,0.5451) -- (-0.2450,0.5100) -- (-0.2403,0.4622) -- (-0.1663,0.4411) -- (-0.1497,0.4100) -- (-0.1732,0.3368) -- (-0.1361,0.3063) -- (-0.0689,0.3437) -- (-0.0351,0.3335) -- (-0.0000,0.2650) -- (0.0478,0.2697) -- (0.0689,0.3437) -- (0.1000,0.3603) -- (0.1732,0.3368) -- (0.2037,0.3739) -- (0.1663,0.4411) -- (0.1765,0.4749) -- cycle;
    \draw[fill=white,line width=0.8pt] (0,0.51) circle (0.075);
  \end{scope}
}
\begin{scope}[shift={(2.65,-2.35)},draw=black,line width=0.8pt]
  \draw[fill=pipelinePink,rounded corners=1pt]
    (-0.23,0.28) .. controls (-0.23,0.47) and (-0.12,0.48) .. (0,0.48)
    .. controls (0.12,0.48) and (0.23,0.47) .. (0.23,0.28) -- cycle;
  \draw[fill=orange!25!white] (0,0.65) circle (0.11);
\end{scope}
\node[stageLabel] at (0,0.09) {SEC 10-K filings\\Item 7 Corpus};
\node[stageLabel] at (2.65,0.09) {Task 1:\\Graph Condition\\Retrieval};
\node[stageLabel] at (5.3,0.09) {Evaluation\\Cross-Model\\Alignment Metrics};
\node[stageLabel] at (0,-2.26) {Selected\\Retrieved Passages};
\node[stageLabel] at (2.65,-2.26) {Human Annotation\\for Classification};
\node[stageLabel,font=\fontsize{7}{7.4}\selectfont] at (5.3,-2.19)
  {Task 2:\\Classification and\\Evaluation against\\Human Labels};
\draw[link] (docs) -- (extract);
\draw[link] (extract) -- (compare);
\draw[link] (pool) -- (gold);
\draw[link] (gold) -- (classify);
\draw[link,dashed] (extract.south) -- ++(0,-0.25) -| (pool.north);
\end{tikzpicture}%
}
\caption{FinGREAT study design. Task I retrieves and assigns evidence to graph edges. Task II evaluates edge classification against human-reviewed answer sets.}
\label{fig:study_pipeline}
\end{figure}
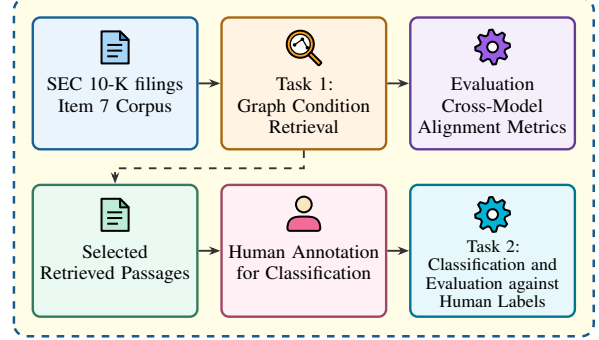

Financial analysis requires more than producing plausible summaries: analysts must locate relevant evidence and interpret the accounting relationships it describes. High-volume disclosure review demands both scale and precision. Financial question-answering benchmarks such as FinQA and TAT-QA demonstrate the challenge of reasoning over dispersed textual and numerical information \citep{chen-etal-2021-finqa,zhu-etal-2021-tat}. Beyond answering specific questions, analysts must systematically discover evidence across double-entry accounting relationships throughout a filing. This motivates schema-guided retrieval, in which financial relationships, predefined by regulatory accounting standards, guide both evidence discovery and its organization for analysis.

These demands also make model deployment consequential. Local open-weight and hosted proprietary models represent different deployment settings: the former permits control over confidential inputs, model adaptation, and inference resources, while the latter provides a managed-service alternative. Small and mid-sized open-weight models offer a practical route to locally hosted analysis, complementing efforts to make financial language models more accessible \citep{yang2023fingpt}. However, their suitability depends on whether they preserve the accounting relation-level accuracy needed by analysts. We therefore compare capability across these settings, asking which financial relationships locally deployable models handle reliably and where proprietary models provide additional value.

We focus on \emph{graph-conditioned financial evidence extraction}. We study this task in Item 7 of annual SEC Form 10-K filings, which contains Management's Discussion and Analysis (MD\&A): management's narrative explanation of financial results, business developments, material trends, uncertainties, and outlook.\ SEC guidance emphasizes explanations of financial performance rather than simply reporting financial figures \citep{sec2003mda}. This narrative provides a natural setting for retrieving explanations of the accounting relationships represented in financial statements. In contrast to open information extraction without predefined target relations \citep{banko2007open}, or retrieval directed by an individual financial question, the accounting schema specifies which relationships to discover across a document. A model must both retrieve relevant passages and map heterogeneous financial expressions to the prescribed edges. Following the bookkeeping-graph framework \citep{liang2023bookkeeping}, our fixed schema contains 11 accounting relationships grouped into Cash Flow edges, Net Income edges, and Equity edges. This formulation connects narrative evidence to a structured representation of financial statements rather than leaving analysts to reconcile unconstrained outputs.

We introduce FinGREAT, a study that separates this joint retrieval-and-assignment task from classification of supplied evidence under the same graph schema. It addresses three research questions:
\begin{enumerate}[leftmargin=0pt]
\setlength{\itemsep}{1pt}
\setlength{\parskip}{0pt}
\item \textbf{RQ1:} How competitive are open-weight models with proprietary models at retrieving evidence under a predefined accounting graph?
\item \textbf{RQ2:} How reliably can models map financial language to accounting relationships, and which categories expose capability gaps?
\item \textbf{RQ3:} How do open-weight and proprietary models compare in edge-classification accuracy?
\end{enumerate}

Our contributions are threefold:
\begin{itemize}[leftmargin=0pt]
\setlength{\itemsep}{1pt}
\setlength{\parskip}{0pt}
\item \textbf{Graph-conditioned retrieval study.} We characterize five open-weight extractors across 1,300 SEC Item 7 filings spanning 2000--2025, with three proprietary references on a shared 300-filing subset.
\item \textbf{Expert-reviewed classification benchmark.} We evaluate 34 open-weight classifiers below 36B parameters and ten proprietary models on 483 passages reviewed by three accounting experts.
\item \textbf{Multi-dimensional model analysis.} We examine annual evidence allocation, cross-model retrieval and classification agreement, tag--edge organization, vocabulary overlap, and category-specific accuracy, complemented by comparisons with externally reported API pricing and first-token latency.
\end{itemize}

\section{Related Works}

\paragraph{Financial language-model evaluation.}
FinQA \citep{chen-etal-2021-finqa} and TAT-QA \citep{zhu-etal-2021-tat} establish benchmarks for numerical reasoning over financial text and tables. FinBen extends evaluation across financial extraction, analysis, reasoning, and decision-making tasks \citep{NEURIPS2024_adb1d9fa}, while FinGPT advances accessible financial model development \citep{yang2023fingpt}. We assess narrative--accounting relation fidelity rather than general financial knowledge or answer generation.

\paragraph{Schema-guided financial extraction.}
Schema-free OpenIE \citep{banko2007open} and schema-conditioned frameworks such as UIE \citep{lu-etal-2022-unified} represent distinct extraction settings. In finance, FIRE benchmarks entity-pair relations \citep{hamad-etal-2024-fire}; FinKario uses templates to construct event-enhanced knowledge graphs \citep{li-etal-2026-finkario}; and TASER extracts schema-conforming information from complex tables \citep{cho-etal-2026-taser}. Accounting structure also guides FinReporting's canonical mapping \citep{zhang-etal-2026-finreporting}, FinTagging's numerical concept linking \citep{wang2025fintagging}, and FinAuditing's taxonomy-aligned consistency tasks \citep{wang2025finauditing}. Thus, schema guidance itself is established prior art. FinGREAT instead examines passage retrieval and classification under a compact graph of accounting flows, grounded in long-standing double-entry accounting principles \citep{liang2023bookkeeping}.

\paragraph{Financial retrieval and extraction assessment.}
FinGEAR combines financial lexicons and document hierarchy for answer retrieval \citep{li-etal-2025-fingear}, and FinMTEB evaluates finance-specific embeddings across retrieval and other tasks \citep{tang-yang-2025-finmteb}. OmniEval separates retrieval from generation in financial RAG \citep{wang-etal-2025-omnieval}, while FinMRAGBench studies multimodal, multi-document evidence \citep{yang-etal-2026-finmragbench}. LEDGER extends retrieval and extraction evaluation to long annual reports and predefined financial indicators \citep{moslonka2026ledger}. FinReflectKG--EvalBench evaluates financial knowledge-graph extraction through multi-dimensional, bias-controlled judging \citep{dimino2025finreflectkg}. These studies motivate distinguishing evidence discovery from the quality of its interpretation.

Our study complements this literature by comparing graph-conditioned retrieval behavior with expert-reviewed passage classification under the same accounting schema. Cross-model evidence agreement characterizes retrieval behavior; human labels assess relation correctness. Category, family, API cost and latency comparisons inform model selection for scalable financial workflows, not a new retrieval architecture.

\begin{table}[t]
\centering
\begingroup
\footnotesize
\setlength{\tabcolsep}{3pt}
\renewcommand{\arraystretch}{1.08}
\DeclareRobustCommand{\rwyes}{\textcolor{green!45!black}{\ding{51}}}
\DeclareRobustCommand{\rwno}{\textcolor{red!80!black}{\ding{55}}}
\DeclareRobustCommand{\rwpartial}{\textcolor{black}{$\ast$}}
\resizebox{\columnwidth}{!}{%
\begin{tabular}{@{}lccccc@{}}
\toprule
Study & SG & ED & RM & SE & HR \\
\midrule
FIRE \citeyearpar{hamad-etal-2024-fire}
 & \rwyes & \rwno & \rwyes & \rwno & \rwyes \\
FinGEAR \citeyearpar{li-etal-2025-fingear}
 & \rwpartial & \rwyes & \rwno & \rwpartial & \rwno \\
FinMRAGBench \citeyearpar{yang-etal-2026-finmragbench}
 & \rwno & \rwyes & \rwno & \rwpartial & \rwno \\
FinKario \citeyearpar{li-etal-2026-finkario}
 & \rwyes & \rwyes & \rwyes & \rwpartial & \rwno \\
FinTagging \citeyearpar{wang2025fintagging}
 & \rwyes & \rwpartial & \rwno & \rwpartial & \rwno \\
FinReflectKG--EvalBench \citeyearpar{dimino2025finreflectkg}
 & \rwyes & \rwyes & \rwyes & \rwpartial & \rwpartial \\
\midrule
\textbf{FinGREAT (ours)}
 & \rwyes & \rwyes & \rwyes & \rwyes & \rwyes \\
\bottomrule
\end{tabular}
}
\caption{Financial benchmark comparison.
SG: schema/structure guidance; ED: document-level evidence discovery;
RM: relation mapping; SE: separate extraction/relation evaluation;
HR: human-labeled relations.
\rwyes: explicit; \rwpartial: partial; \rwno: not evaluated.}
\label{tab:related_benchmark_comparison}
\endgroup
\end{table}

\section{Problem Formulation}
\label{sec:problem_formulation}

We introduce the accounting graph and formulate graph-conditioned evidence extraction, financial-relation classification, and their evaluation.

\subsection{Underlying Accounting Graph}
\label{sec:accounting_graph}

Annual financial statements follow US Generally Accepted Accounting Principles (GAAP) and double-entry bookkeeping, organized around the accounting identity
\begin{equation}
\mathrm{Equity}=\mathrm{Assets}-\mathrm{Liabilities}.
\label{eq:nodes}
\end{equation}

Financial records can be represented as a directed, weighted graph: accounts are nodes, and journal entries are edges whose directions and weights encode bookkeeping flows. This perspective traces to \citet{charnes1963breakeven,mattessich1964accounting} and is developed by \citet{ijiri1975theory,arya2000inferring}.

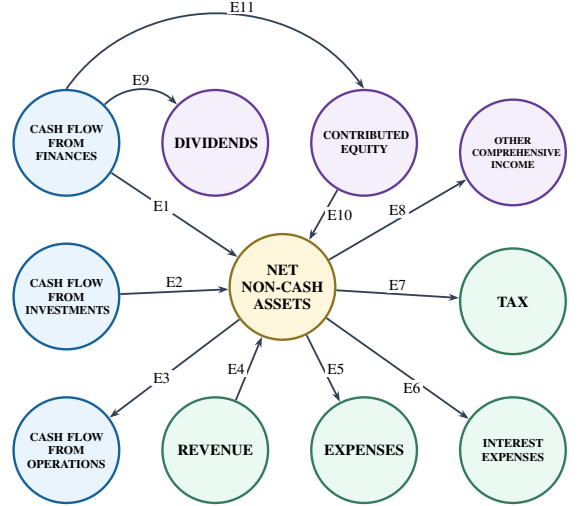
\begin{figure}[t]
\centering
\resizebox{\columnwidth}{!}{%
\definecolor{graphCash}{HTML}{2496ED}
\definecolor{graphIncome}{HTML}{57CC99}
\definecolor{graphDividend}{HTML}{9B5DE5}
\definecolor{graphCapital}{HTML}{F26B9B}
\definecolor{graphAssets}{HTML}{FFD23F}
\definecolor{graphInk}{HTML}{344054}
\begin{tikzpicture}[
    financial node/.style={circle,draw=graphInk,line width=1.4pt,
        minimum size=2.2cm,text width=1.7cm,inner sep=0pt,
        align=center,font=\footnotesize\bfseries,text=black},
    cash/.style={financial node,draw=graphCash!65!black,fill=graphCash!10!white},
    income/.style={financial node,draw=graphIncome!60!black,fill=graphIncome!12!white},
    equity/.style={financial node,draw=graphDividend!65!black,fill=graphDividend!10!white},
    flow/.style={-{Stealth[length=2.2mm,width=1.5mm]},
        draw=graphInk,line width=1.1pt}]
\node[financial node,draw=graphAssets!55!black,fill=graphAssets!18!white,font=\footnotesize\bfseries] (nca) at (0,0) {\MakeUppercase{Net\\Non-cash\\Assets}};
\node[cash,font=\scriptsize\bfseries] (cff) at (-4.5,3) {\MakeUppercase{Cash Flow\\from\\Finances}};
\node[cash,font=\scriptsize\bfseries] (cfi) at (-4.5,-0.2) {\MakeUppercase{Cash Flow\\from\\Investments}};
\node[cash,font=\scriptsize\bfseries] (cfo) at (-4.5,-3.4) {\MakeUppercase{Cash Flow\\from\\Operations}};
\node[equity] (div) at (-1.4,3) {\MakeUppercase{Dividends}};
\node[equity,font=\scriptsize\bfseries] (contributed) at (1.7,3) {\MakeUppercase{Contributed\\Equity}};
\node[income] (revenue) at (-1.4,-3.4) {\MakeUppercase{Revenue}};
\node[income] (expense) at (1.7,-3.4) {\MakeUppercase{Expenses}};
\node[income,font=\scriptsize\bfseries] (interest) at (4.8,-3.4) {\MakeUppercase{Interest\\Expenses}};
\node[income,font=\small\bfseries] (tax) at (4.8,-0.3) {\MakeUppercase{Tax}};
\node[equity,font=\tiny\bfseries] (oci) at (4.8,2.8) {\MakeUppercase{Other\\Comprehensive\\Income}};
\tikzset{edge label/.style={font=\footnotesize,text=black,fill=white,
    inner sep=1.5pt,outer sep=1pt}}
\draw[flow] (cff) -- node[edge label,midway,above left] {E1} (nca);
\draw[flow] (cfi) -- node[edge label,midway,above] {E2} (nca);
\draw[flow] (nca) -- node[edge label,midway,below left] {E3} (cfo);
\draw[flow] (cff.north east) to[out=45,in=135] node[edge label,midway,above] {E9} (div.north west);
\draw[flow] (cff.north) to[out=55,in=125,looseness=1.05] node[edge label,pos=0.58,above] {E11} (contributed.north);
\draw[flow] (contributed) -- node[edge label,midway,right] {E10} (nca);
\draw[flow] (revenue) -- node[edge label,midway,left] {E4} (nca);
\draw[flow] (nca) -- node[edge label,midway,right] {E5} (expense);
\draw[flow] (nca) -- node[edge label,pos=0.6,below] {E6} (interest);
\draw[flow] (nca) -- node[edge label,midway,above] {E7} (tax);
\draw[flow] (nca) -- node[edge label,midway,above] {E8} (oci);
\end{tikzpicture}%
}
\captionsetup{font=footnotesize}
\caption{Underlying accounting graph used in the task formulation.
Cash Flow edges connect the blue nodes to net non-cash assets (yellow), Net Income edges connect the green nodes, and Equity edges involve the purple nodes. Directed edges represent bookkeeping flows.
The flow identity is $\Delta\mathrm{Cash}+\Delta\mathrm{NCA}-\Delta\mathrm{CE}-\mathrm{NI}+\mathrm{Div}=0$, where NCA denotes net non-cash assets, CE contributed equity, NI net income, and Div dividends.}
\label{fig:accounting-graph}
\end{figure}

Following \citet{liang2026accounting}, we use the fixed 11-node, 11-edge schema in Fig.~\ref{fig:accounting-graph}. The schema constrains allowable relationships. For example, income-statement accounts connect to net non-cash assets rather than directly to dividends. Although account balances and flow values or directions vary across firm-years, the underlying relationships remain fixed. At the level of these aggregated accounting concepts, the graph comprehensively represents transactions affecting the accounting identity. Each transaction's bookkeeping inflows and outflows balance to zero, preserving Eq.~\ref{eq:nodes}. Consequently, any Item 7 discussion of transactions affecting this identity must relate to one or more of the 11 edges in the schema. This domain-specific completeness motivates the graph-conditioned extraction and annotation tasks studied here.

\begin{table}[t]
\centering
\caption{Mathematical notation used in the task formulation
and evaluation.}
\footnotesize
\setlength{\tabcolsep}{4pt}
\begin{tabularx}{\columnwidth}{@{}lX@{}}
\toprule
\textbf{Symbol} & \textbf{Definition} \\
\midrule
$G=(V,E)$
    & Accounting graph with 11 nodes, 11 edges \\
$e \in E$
    & An accounting relationship \\
$m \in \mathcal{M}$
    & An evaluated language model \\
$\mathcal{M}_O,\mathcal{M}_P$
    & Open-weight and proprietary model sets \\
$\mathcal{Y}$
    & Set of years (2000--2025) \\
$\mathcal{C}_y$
    & Corpus of SEC 10-K Item 7 texts from year $y$ \\
$c \in \mathcal{C}_y$
    & An individual Item 7 text \\
$\mathcal{R}_{m,c}$
    & Retrieved extractions by model $m$ from $c$ \\
$x=(s,\hat{e},T)$
    & Extraction containing a passage, edge, and semantic tags \\
$n_{m,y,e}$
    & Extractions assigned to edge $e$ by model $m$ in year $y$ \\
$\mathcal{O},\mathcal{P}$
    & Open-weight and proprietary extraction sets on common texts \\
$K,A$
    & Matched passages and matches with agreeing edge labels between two models\\
$\mathcal{H}$
    & Human-annotated classification benchmark \\
$Y_i \subseteq E$
    & Human-accepted edges for example $i$ \\
$\hat{e}_{m,i}$
    & Edge predicted by model $m$ for example $i$ \\
\bottomrule
\end{tabularx}
\label{tab:notation}
\end{table}


\subsection{Task I: Graph-Conditioned Evidence Extraction}
\label{sec:extraction_task}

Using the accounting graph $G$ introduced in
Section~\ref{sec:accounting_graph}, we formulate financial
evidence extraction as identifying textual passages
corresponding to predefined accounting relationships.
The notation is summarized in Table~\ref{tab:notation}.
Unlike conventional query-conditioned retrieval, our
task requires models to systematically discover evidence
associated with the graph's relationships throughout
each document. For a model $m$ and an Item 7 text $c$, the extraction task is defined as:

\begin{equation}
f_m(c,G) \rightarrow \mathcal{R}_{m,c}
\label{eq:extraction}
\end{equation}

Each extraction $x=(s,\hat{e},T)$ contains a textual
passage $s$, its assigned graph edge $\hat{e}\in E$,
and a set of model-generated descriptive tags $T$.
Multiple passages may correspond to the same edge, while
some edges may have no supporting evidence.

Our objective is to evaluate models' ability to discover
relevant financial evidence and consistently
associate it with accounting relationships (graph edges). Owing to the high volume of text processed in this task, we compare the performance of open-weight models (\(\mathcal{M}_O\)) against proprietary models (\(\mathcal{M}_P\)).


\subsection{Task II: Financial Relation Classification}
\label{sec:classification_task}

Our second task isolates financial accounting relation understanding
from document-level evidence discovery. Task I focuses on retrieval and classification simultaneously. To isolate financial relation understanding
from document-level evidence discovery, we introduce Task II - given a financial passage and the accounting graph, a model must identify the financial accounting relationship expressed by the passage. The relationships are derived from the same graph schema described in Figure \ref{fig:accounting-graph}. We define a human-annotated benchmark:

\begin{equation}
\mathcal{H} = \{(s_i,Y_i)\}_{i=1}^{N}
\label{eq:gold_dataset}
\end{equation}

where $Y_i\subseteq E$ represents the nonempty set of
human-labelled edges, accommodating cases with multiple
valid relationships. A particular passage may discuss interactions among multiple accounting concepts. Some passages therefore have multiple accepted edges i.e. \(|Y_i| > 1\). However, we simplify the objective for the model - classify the passage to one among the accepted edges. The classification task is:

\begin{equation}
h_m(s_i,G) \rightarrow \hat{e}_{m,i}\in E
\label{eq:classification}
\end{equation}

A prediction is correct if $\hat{e}_{m,i}\in Y_i$. By directly providing the evidence passage, this task
enables human-grounded comparisons of financial relation
classification across models, independently of their
evidence discovery capabilities.


\subsection{Evaluation Objectives and Metrics}
\label{sec:evaluation_metrics}

We evaluate models along four complementary dimensions:
(i) extraction behavior, (ii) cross-model agreement,
(iii) semantic consistency, and
(iv) human-grounded classification. The first three dimensions are evaluated using Task I and the fourth using Task II.

\paragraph{Evidence Distribution and Allocation.}

We characterize extraction volume using annual counts
$n_{m,y,e}$ and the proportion of extractions assigned
to each graph edge:

\begin{equation}
C_{m,e} =
\frac{\sum_{y\in\mathcal{Y}}n_{m,y,e}}
{\sum_{y\in\mathcal{Y}}\sum_{e'\in E}n_{m,y,e'}}
\label{eq:edge_coverage}
\end{equation}

The edge distribution sums to unity for each model
with valid predictions. We additionally examine
longitudinal extraction trends across the corpus.

\paragraph{Cross-Model Agreement.}

We compare open-weight models $m  \in \mathcal{M_{O}}$ against proprietary reference models $m' \in \mathcal{M_{P}}$ along two dimensions. First, we measure \textit{retrieval recall}, assessing whether
open-weight models recover the same evidence passages
as proprietary models, irrespective of their assigned
accounting relationships. Because coverage alone does not penalize additional unmatched extractions, we also study the Jaccard Similarity between the retrieved passages of the two models. Second, we measure \textit{conditional classification precision} on these recalled passages, evaluating whether both models assign the same accounting relationship. These metrics are defined below.

\begin{align}
\text{Retrieval Recall}
&= \frac{|K|}{|\mathcal{P}|},
\label{eq:retrieval_recall}\\
\text{Jaccard Similarity}
&= \frac{|K|}{|\mathcal{O}|+|\mathcal{P}|-|K|},
\label{eq:evidence_jaccard}\\
\text{Conditional Precision}
&= \frac{|A|}{|K|}.
\label{eq:conditional_precision}
\end{align}

\paragraph{Semantic Consistency.}
There are two sets of vocabulary that the models are dealing with - one that is presented to the model in the form of a graph (Figure \ref{fig:accounting-graph}) and one that they inherently possess from their training data. The models are asked to produce tags \(T\) along with the relationship \(\hat{e}\) for each retrieved passage. We aim to study the semantic consistency between model-generated tags and graph edges. This informs whether models are able to consistently translate the given schema to their internally learned vocabulary. Using the tag--edge distribution $p(t,e)$, we calculate purity, entropy, and homogeneity:

\begin{align}
\operatorname{Purity}
&= \sum_t p(t)\max_e p(e\mid t),
\label{eq:purity}\\
H(E\mid T)
&= -\sum_{t,e}p(t,e)\log_2 p(e\mid t),
\label{eq:conditional_entropy}\\
\operatorname{Homogeneity}
&= 1-\frac{H(E\mid T)}{H(E)}.
\label{eq:homogeneity}
\end{align}

These metrics are meant to evaluate internal tag-to-edge consistency rather than the factual correctness of
predicted relationships.

\paragraph{Human-Grounded Classification.}
This evaluation is done using task II. We measure the overall and per-edge accuracy of the models against expert finance professionals on the same task. Let $\mathcal{H}_e$ denote retrieved passages whose human-labeled edge is $e$. Per-edge accuracy is:

\begin{equation}
\operatorname{Acc}_{m,e} =
\frac{1}{|\mathcal{H}_e|}
\sum_{i\in\mathcal{H}_e}
\mathbf{1}[\hat{e}_{m,i}\in Y_i]
\label{eq:edge_accuracy}
\end{equation}

We additionally report category-level performance to
identify systematic differences in models\' understanding
of accounting relationships.

\section{Experimental Methodology}
\label{sec:methodology}

Our study examines two complementary tasks.
Task I retrieves passages from financial disclosures and assigns each to an accounting relationship (graph edge) while Task II takes supplied passages and evaluates edge classification alone against human-accepted labels. Figure~\ref{fig:study_pipeline} summarizes the two stages.

\subsection{Financial Corpus and Dataset}
\label{sec:disclosure_corpus}

We use annual SEC Form 10-K MD\&A (Item 7) texts from 2000--2025. All five open-weight extractors process 1300 firm-year filings (1237 unique firms), with 50 filings in each of the 26 years. The corpus covers SEC registrants across 9 industry groups and spans a broad range of annual assets. Processed Item 7 texts have a median length of 57.4k characters (IQR: 36.6k--78.3k), illustrating substantial variation in disclosure length. This annual coverage supports longitudinal analysis of how models allocate narrative evidence across accounting relationships.  Comparisons with the three proprietary references use a common subset of 300 firm-year filings: the same 50 selected firm-year filings in each of the years \(\{2000, 2005, 2010, 2015, 2020, 2025\}\). Cross-model comparisons therefore hold the input documents fixed. Corpus selection and associated statistics are documented in Appendix~\ref{app:industry_composition}--\ref{app:document_length}.


\begin{table*}[!t]\centering\scriptsize\setlength{\tabcolsep}{4pt}
\resizebox{\textwidth}{!}{%
\begin{tabular}{llrrrrccc}\toprule
Reference model & Open-weight model & $P$ & $O$ & $K$ & $A$ & \shortstack{Retrieval Recall\\{[$K/P$]}} & \shortstack{Jaccard Similarity\\{[$K/(O+P-K)$]}} & \shortstack{Conditional Classification\\Precision [$A/K$]} \\
\midrule
\multirow{5}{*}{\shortstack[l]{Gemini 3.1\\Flash Lite}} & Gemma 4 31B & 4,061 & 7,534 & 2,782 & 2,217 & 68.5 & \textbf{31.6} & \textbf{79.7} \\
 & Nemotron 3 Super & 4,061 & 8,742 & 2,110 & 1,248 & 52.0 & 19.7 & 59.1 \\
 & Qwen 3.6 35B A3B & 4,061 & 8,023 & 2,668 & 1,915 & 65.7 & 28.3 & 71.8 \\
 & Ministral 3 14B & 4,061 & 19,983 & 3,170 & 1,674 & \textbf{78.1} & 15.2 & 52.8 \\
 & Phi 4 & 4,061 & 12,547 & 1,797 & 768 & 44.3 & 12.1 & 42.7 \\
\midrule
\multirow{5}{*}{\shortstack[l]{Claude\\Sonnet 5}} & Gemma 4 31B & 14,802 & 7,534 & 5,602 & 4,577 & 37.8 & 33.5 & \textbf{81.7} \\
 & Nemotron 3 Super & 14,802 & 8,742 & 5,165 & 3,152 & 34.9 & 28.1 & 61.0 \\
 & Qwen 3.6 35B A3B & 14,802 & 8,023 & 5,679 & 4,163 & 38.4 & 33.1 & 73.3 \\
 & Ministral 3 14B & 14,802 & 19,983 & 8,842 & 4,788 & \textbf{59.7} & \textbf{34.1} & 54.2 \\
 & Phi 4 & 14,802 & 12,547 & 4,790 & 2,220 & 32.4 & 21.2 & 46.3 \\
\midrule
\multirow{5}{*}{\shortstack[l]{GPT-6\\Sol}} & Gemma 4 31B & 13,571 & 7,534 & 5,160 & 4,050 & 38.0 & \textbf{32.4} & \textbf{78.5} \\
 & Nemotron 3 Super & 13,571 & 8,742 & 4,130 & 2,336 & 30.4 & 22.7 & 56.6 \\
 & Qwen 3.6 35B A3B & 13,571 & 8,023 & 5,115 & 3,568 & 37.7 & 31.0 & 69.8 \\
 & Ministral 3 14B & 13,571 & 19,983 & 7,721 & 3,742 & \textbf{56.9} & 29.9 & 48.5 \\
 & Phi 4 & 13,571 & 12,547 & 4,150 & 1,678 & 30.6 & 18.9 & 40.4 \\
\bottomrule\end{tabular}}
\caption{Cross-model agreement on 300 common filings (\%). $P,O$: reference/open extraction counts. $K$: evidence matches. $A$: matches agreeing on the edge. Bold marks the best score per reference.}
\label{tab:results_reference_agreement}
\end{table*}

\begin{table}[b]\centering\scriptsize\setlength{\tabcolsep}{1.5pt}
\resizebox{\columnwidth}{!}{%
\begin{tabular}{lrrrrr}\toprule
\paperheadername{Model} & \paperheadername{Purity\\(\%, $\uparrow$)} & \paperheadername{$H(E\mid T)$\\(bits, $\downarrow$)} & \paperheadername{Homogeneity\\(\%, $\uparrow$)} & \paperheadername{Distinct\\tags} & \paperheadername{Used once\\(\%)} \\
\midrule
Gemma 4 31B & 81.5 & 0.790 & \textbf{74.2} & 165 & 55.2 \\
Nemotron 3 Super & 79.5 & 0.962 & 71.7 & 588 & 60.7 \\
Qwen 3.6 35B A3B & \textbf{85.7} & \textbf{0.775} & \textbf{74.2} & 92 & 41.3 \\
Ministral 3 14B & 63.5 & 1.665 & 50.3 & 2,213 & 46.6 \\
Phi 4 & 65.8 & 1.376 & 59.8 & 3,897 & 54.3 \\
\hdashline[1pt/2pt]
Gemini 3.1 Flash Lite & 74.6 & 0.990 & 68.5 & 28 & 10.7 \\
Claude Sonnet 5 & 70.8 & 1.296 & 59.7 & 118 & 44.9 \\
GPT-6 Sol & 64.2 & 1.477 & 53.4 & 107 & 36.4 \\
\bottomrule\end{tabular}}
\caption{Tag--edge associations. Arrows indicate preferred directions. Bold marks the best open-weight scores at displayed precision. Above/below the dotted line: 1,300/300 filings.}
\label{tab:results_tag_organization}
\end{table}

\begin{table*}[!t]\centering\scriptsize
\setlength{\tabcolsep}{1.5pt}\renewcommand{\arraystretch}{1.05}
\setlength{\aboverulesep}{0pt}\setlength{\belowrulesep}{0pt}
\resizebox{\textwidth}{!}{%
\begin{tabular}{l!{\vrule width .45pt\hspace{2.55pt}}rrrr!{\vrule width .45pt\hspace{2.55pt}}rrrrr!{\vrule width .45pt\hspace{2.55pt}}rrrrr!{\vrule width .45pt\hspace{2.55pt}}r}\hline
\multicolumn{1}{l!{\hspace{3pt}}}{\rule{0pt}{2.8ex}} & \multicolumn{4}{c!{\hspace{3pt}}}{CASH FLOW EDGES} & \multicolumn{5}{c!{\hspace{3pt}}}{NET INCOME EDGES} & \multicolumn{5}{c!{\hspace{3pt}}}{EQUITY EDGES} & \multicolumn{1}{c}{ALL EDGES} \\[5pt]
\multicolumn{1}{l!{\vrule width .45pt\hspace{2.55pt}}}{\paperheadername{Model}} & \paperheadername{Finance} & \paperheadername{Investment} & \paperheadername{Operations} & \multicolumn{1}{r!{\vrule width .45pt\hspace{2.55pt}}}{\paperheadername{Overall}} & \paperheadername{Revenue} & \paperheadername{Expenses} & \paperheadername{Interest\\Expenses} & \paperheadername{Tax} & \multicolumn{1}{r!{\vrule width .45pt\hspace{2.55pt}}}{\paperheadername{Overall}} & \paperheadername{OCI} & \paperheadername{Dividends} & \paperheadername{Non-cash\\Equity} & \paperheadername{Cash\\Equity} & \multicolumn{1}{r!{\vrule width .45pt\hspace{2.55pt}}}{\paperheadername{Overall}} & \paperheadername{Overall} \\
 & (E1) & (E2) & (E3) & (E1--E3) & (E4) & (E5) & (E6) & (E7) & (E4--E7) & (E8) & (E9) & (E10) & (E11) & (E8--E11) & (E1--E11) \\
 & $n=82$ & $n=70$ & $n=66$ & $n=195$ & $n=11$ & $n=72$ & $n=37$ & $n=60$ & $n=174$ & $n=17$ & $n=44$ & $n=29$ & $n=103$ & $n=172$ & $n=483$ \\
\midrule
Qwen3.6-35B-A3B & 54.9 & 57.1 & 80.3 & 61.5 & 27.3 & 87.5 & 89.2 & 95.0 & 88.5 & 94.1 & 97.7 & 10.3 & 98.1 & 83.1 & 77.0 \\
Qwen3.6-27B & 58.5 & 80.0 & 89.4 & \paperbadge[black]{rankGold}{75.4} & 54.5 & 86.1 & 94.6 & 96.7 & \paperbadge[black]{rankSilver}{91.4} & 100.0 & 90.9 & 62.1 & 97.1 & \paperbadge[black]{rankSilver}{89.5} & \paperbadge[black]{rankGold}{84.1} \\
Qwen3.5-35B-A3B & 52.4 & 55.7 & 80.3 & 61.0 & 45.5 & 88.9 & 97.3 & 96.7 & \paperbadge[black]{rankGold}{92.0} & 100.0 & 97.7 & 31.0 & 99.0 & 87.2 & 78.5 \\
Qwen3.5-27B & 51.2 & 80.0 & 86.4 & \paperbadge[black]{rankSilver}{72.8} & 54.5 & 86.1 & 94.6 & 96.7 & \paperbadge[black]{rankBronze}{90.8} & 100.0 & 95.5 & 58.6 & 98.1 & \paperbadge[black]{rankGold}{91.3} & \paperbadge[black]{rankGold}{84.1} \\
Qwen3.5-9B & 68.3 & 45.7 & 87.9 & 65.6 & 18.2 & 83.3 & 94.6 & 96.7 & 87.9 & 76.5 & 84.1 & 62.1 & 86.4 & 83.7 & 78.7 \\
Qwen3.5-4B & 59.8 & 28.6 & 83.3 & 54.9 & 45.5 & 76.4 & 94.6 & 96.7 & 86.2 & 100.0 & 90.9 & 62.1 & 55.3 & 66.3 & 69.2 \\
Qwen3.5-2B & 89.0 & 14.3 & 68.2 & 56.4 & 45.5 & 11.1 & 59.5 & 76.7 & 46.0 & 47.1 & 27.3 & 3.4 & 33.0 & 28.5 & 41.8 \\
Qwen3.5-0.8B & 86.6 & 20.0 & 43.9 & 49.7 & 9.1 & 11.1 & 24.3 & 85.0 & 39.7 & 17.6 & 27.3 & 31.0 & 29.1 & 27.9 & 36.4 \\
\hdashline[1pt/2pt]
Gemma-4-31B-IT & 53.7 & 88.6 & 92.4 & \paperbadge[black]{rankGold}{75.4} & 45.5 & 81.9 & 89.2 & 95.0 & 87.4 & 94.1 & 95.5 & 51.7 & 96.1 & 87.8 & \paperbadge[black]{rankBronze}{83.0} \\
Gemma-4-26B-A4B-IT & 51.2 & 72.9 & 87.9 & 68.2 & 54.5 & 86.1 & 89.2 & 96.7 & 90.2 & 100.0 & 95.5 & 27.6 & 96.1 & 84.3 & 79.9 \\
Gemma-4-12B-IT & 48.8 & 87.1 & 87.9 & \paperbadge[black]{rankBronze}{72.3} & 45.5 & 87.5 & 97.3 & 98.3 & \paperbadge[black]{rankSilver}{91.4} & 100.0 & 95.5 & 75.9 & 87.4 & \paperbadge[black]{rankBronze}{88.4} & \paperbadge[black]{rankSilver}{83.2} \\
Gemma-4-E4B-IT & 42.7 & 68.6 & 72.7 & 60.5 & 27.3 & 75.0 & 91.9 & 93.3 & 84.5 & 100.0 & 95.5 & 34.5 & 82.5 & 78.5 & 74.5 \\
Gemma-4-E2B-IT & 26.8 & 52.9 & 53.0 & 44.6 & 45.5 & 94.4 & 100.0 & 91.7 & \paperbadge[black]{rankGold}{92.0} & 70.6 & 79.5 & 75.9 & 40.8 & 57.6 & 65.2 \\
Gemma-3-27B-IT & 47.6 & 51.4 & 75.8 & 55.4 & 18.2 & 91.7 & 91.9 & 98.3 & 89.7 & 94.1 & 93.2 & 65.5 & 87.4 & 84.9 & 74.9 \\
Gemma-3-12B-IT & 52.4 & 55.7 & 81.8 & 61.5 & 45.5 & 72.2 & 94.6 & 98.3 & 85.1 & 70.6 & 93.2 & 51.7 & 68.0 & 70.3 & 71.2 \\
Gemma-3-4B-IT & 29.3 & 34.3 & 28.8 & 32.3 & 36.4 & 63.9 & 94.6 & 90.0 & 78.2 & 76.5 & 88.6 & 62.1 & 43.7 & 55.8 & 54.2 \\
Gemma-3-1B-IT & 59.8 & 10.0 & 19.7 & 30.3 & 9.1 & 1.4 & 37.8 & 36.7 & 21.8 & 0.0 & 50.0 & 0.0 & 23.3 & 22.7 & 22.4 \\
Gemma-3-270M-IT & 20.7 & 2.9 & 3.0 & 9.2 & 36.4 & 2.8 & 0.0 & 0.0 & 2.9 & 35.3 & 61.4 & 0.0 & 13.6 & 19.2 & 8.9 \\
\hdashline[1pt/2pt]
Ministral-3-14B-Instruct & 53.7 & 80.0 & 80.3 & 69.7 & 36.4 & 80.6 & 91.9 & 96.7 & 86.8 & 88.2 & 88.6 & 65.5 & 89.3 & 87.2 & 80.1 \\
Ministral-3-8B-Instruct & 31.7 & 74.3 & 63.6 & 53.8 & 45.5 & 73.6 & 89.2 & 93.3 & 82.8 & 88.2 & 81.8 & 86.2 & 35.0 & 56.4 & 64.8 \\
Ministral-3-3B-Instruct & 39.0 & 80.0 & 53.0 & 55.9 & 36.4 & 62.5 & 94.6 & 96.7 & 80.5 & 76.5 & 84.1 & 69.0 & 56.3 & 66.9 & 66.9 \\
\hdashline[1pt/2pt]
Phi-4 & 57.3 & 62.9 & 83.3 & 65.6 & 18.2 & 68.1 & 97.3 & 96.7 & 80.5 & 100.0 & 95.5 & 44.8 & 48.5 & 59.9 & 70.2 \\
Phi-4-Mini-Instruct & 35.4 & 68.6 & 16.7 & 41.5 & 9.1 & 18.1 & 81.1 & 56.7 & 43.7 & 70.6 & 84.1 & 55.2 & 46.6 & 56.4 & 45.1 \\
Phi-3.5-Mini-Instruct & 39.0 & 48.6 & 69.7 & 54.4 & 9.1 & 31.9 & 91.9 & 53.3 & 51.1 & 70.6 & 72.7 & 62.1 & 59.2 & 64.0 & 54.7 \\
Phi-3-Medium-128K-Instruct & 26.8 & 35.7 & 13.6 & 24.6 & 27.3 & 4.2 & 18.9 & 3.3 & 5.7 & 29.4 & 38.6 & 17.2 & 44.7 & 34.9 & 19.5 \\
Phi-3-Small-128K-Instruct & 40.2 & 45.7 & 43.9 & 45.1 & 0.0 & 56.9 & 43.2 & 76.7 & 59.2 & 82.4 & 86.4 & 17.2 & 95.1 & 77.9 & 58.8 \\
Phi-3-Mini-128K-Instruct & 58.5 & 38.6 & 65.2 & 56.4 & 9.1 & 19.4 & 83.8 & 53.3 & 44.3 & 82.4 & 61.4 & 41.4 & 43.7 & 50.6 & 49.5 \\
\hdashline[1pt/2pt]
Llama-3.2-3B-Instruct & 61.0 & 15.7 & 84.8 & 51.3 & 9.1 & 33.3 & 83.8 & 91.7 & 63.8 & 23.5 & 34.1 & 51.7 & 26.2 & 33.1 & 48.0 \\
Llama-3.2-1B-Instruct & 70.7 & 12.9 & 13.6 & 31.3 & 9.1 & 2.8 & 54.1 & 26.7 & 22.4 & 11.8 & 22.7 & 20.7 & 32.0 & 25.6 & 23.0 \\
Llama-3.1-8B-Instruct & 45.1 & 34.3 & 31.8 & 37.9 & 0.0 & 68.1 & 86.5 & 75.0 & 70.7 & 47.1 & 75.0 & 69.0 & 48.5 & 55.8 & 52.8 \\
Meta-Llama-3-8B-Instruct & 85.4 & 17.1 & 62.1 & 53.8 & 18.2 & 37.5 & 81.1 & 60.0 & 53.4 & 0.0 & 56.8 & 3.4 & 43.7 & 37.8 & 46.6 \\
\hdashline[1pt/2pt]
DeepSeek-V2-Lite-Chat & 64.6 & 27.1 & 31.8 & 41.0 & 0.0 & 31.9 & 75.7 & 41.7 & 43.7 & 11.8 & 68.2 & 10.3 & 51.5 & 44.8 & 42.0 \\
DeepSeek-MoE-16B-Chat & 20.7 & 4.3 & 4.5 & 10.3 & 0.0 & 1.4 & 21.6 & 11.7 & 9.2 & 23.5 & 95.5 & 0.0 & 29.1 & 33.7 & 15.9 \\
DeepSeek-LLM-7B-Chat & 22.0 & 31.4 & 18.2 & 24.1 & 0.0 & 12.5 & 18.9 & 26.7 & 18.4 & 29.4 & 61.4 & 3.4 & 14.6 & 22.7 & 20.9 \\
\midrule
GPT-6.1-Sol & 84.1 & 92.9 & 93.9 & \paperbadge{rankSilver}{89.2} & 72.7 & 94.4 & 91.9 & 95.0 & \paperbadge{rankBronze}{93.7} & 94.1 & 93.2 & 96.6 & 97.1 & \paperbadge{rankBronze}{95.3} & \paperbadge{rankSilver}{92.5} \\
GPT-6-Astra & 84.1 & 91.4 & 93.9 & \paperbadge{rankBronze}{88.7} & 72.7 & 93.1 & 89.2 & 93.3 & 93.1 & 94.1 & 97.7 & 96.6 & 99.0 & \paperbadge{rankGold}{97.7} & \paperbadge{rankGold}{92.8} \\
GPT-6-Luna & 56.1 & 68.6 & 78.8 & 66.7 & 72.7 & 81.9 & 91.9 & 90.0 & 87.4 & 94.1 & 90.9 & 34.5 & 94.2 & 84.9 & 79.3 \\
\hdashline[1pt/2pt]
Claude-Opus-5-5 & 84.1 & 88.6 & 93.9 & 88.2 & 81.8 & 94.4 & 86.5 & 96.7 & \paperbadge{rankSilver}{94.3} & 88.2 & 90.9 & 96.6 & 97.1 & \paperbadge{rankBronze}{95.3} & \paperbadge{rankBronze}{92.1} \\
Claude-Sonnet-5-5 & 80.5 & 91.4 & 90.9 & 87.7 & 54.5 & 91.7 & 97.3 & 93.3 & 92.0 & 100.0 & 95.5 & 79.3 & 95.1 & 93.6 & 91.3 \\
Claude-Fable-5-1 & 84.1 & 92.9 & 97.0 & \paperbadge{rankGold}{89.7} & 81.8 & 93.1 & 91.9 & 95.0 & \paperbadge{rankGold}{94.8} & 94.1 & 95.5 & 86.2 & 96.1 & 94.2 & \paperbadge{rankGold}{92.8} \\
Claude-Haiku-4-5 & 72.0 & 77.1 & 89.4 & 78.5 & 54.5 & 84.7 & 89.2 & 90.0 & 87.4 & 100.0 & 97.7 & 72.4 & 88.3 & 88.4 & 84.5 \\
\hdashline[1pt/2pt]
Gemini-3.8-Flash & 79.3 & 91.4 & 93.9 & 86.7 & 54.5 & 90.3 & 89.2 & 93.3 & 90.8 & 100.0 & 93.2 & 100.0 & 95.1 & \paperbadge{rankSilver}{95.9} & 91.3 \\
Gemini-3.5-Flash-Lite & 53.7 & 78.6 & 83.3 & 71.8 & 45.5 & 86.1 & 89.2 & 96.7 & 89.7 & 100.0 & 95.5 & 89.7 & 86.4 & 89.0 & 82.4 \\
Gemini-3.1-Pro-Preview & 78.0 & 84.3 & 86.4 & 82.1 & 72.7 & 95.8 & 94.6 & 96.7 & \paperbadge{rankSilver}{94.3} & 94.1 & 90.9 & 86.2 & 93.2 & 91.9 & 89.6 \\
\midrule
Most frequent baseline & 39.0 & 2.9 & 9.1 & 19.0 & 0.0 & 1.4 & 0.0 & 0.0 & 0.6 & 17.6 & 36.4 & 6.9 & 100.0 & 59.9 & 21.3 \\
Uniform random baseline & 15.6 & 12.7 & 13.6 & 12.9 & 14.9 & 10.6 & 10.6 & 10.0 & 10.2 & 13.4 & 15.7 & 13.2 & 14.6 & 13.2 & 11.1 \\
\bottomrule\end{tabular}}
\caption{Human-reviewed edge accuracy (\%) on 483 passages. $n$ counts accepted-set membership. Category Overall uses the union of its edge cohorts. OCI: other comprehensive income. Medal colors rank category/overall scores separately within open-weight and proprietary groups. Black borders identify open-weight highlights.}
\label{tab:results_human_leaderboard}
\end{table*}

\subsection{Graph-Conditioned Extraction Study}
\label{sec:extraction_study}

Each model receives Item 7 text and the graph's relationship
definitions, and is instructed to identify supporting passages and assign
an edge to each extraction. The requested structured output contains the
quoted passage, its edge endpoints, and optional descriptive tags. These
correspond to $x=(s,\hat e,T)$ in Section~\ref{sec:extraction_task}. To account for varying model context sizes and long-context information retrieval capacities, the item 7 text is chunked into batches of \(\approx 4096\) tokens (accounting for line, paragraph, and section edge cases). Full prompts, model parameter settings, chunking strategy, and output examples are provided in Appendix~\ref{app:retrieval_protocol}.

\begin{figure}[t]\centering
{\scriptsize\setlength{\tabcolsep}{4pt}
\makebox[\columnwidth][c]{\begin{tabular}{@{}ccc@{}}
\textcolor{teal}{\rule{9pt}{2pt}} Gemma 4 31B & \textcolor{blue}{\rule{9pt}{2pt}} Ministral 3 14B & \textcolor{plotRed}{\rule{9pt}{2pt}} Phi 4
\end{tabular}}\par
\makebox[\columnwidth][c]{\begin{tabular}{@{}cc@{}}
\textcolor{orange}{\rule{9pt}{2pt}} Nemotron 3 Super & \textcolor{violet}{\rule{9pt}{2pt}} Qwen 3.6 35B A3B
\end{tabular}}}\par\vspace{2pt}
\begin{tikzpicture}\begin{axis}[width=0.99\columnwidth,height=3.4cm,xmin=2000,xmax=2025,ymin=0,ymax=60,xtick={2000,2001,2002,2003,2004,2005,2006,2007,2008,2009,2010,2011,2012,2013,2014,2015,2016,2017,2018,2019,2020,2021,2022,2023,2024,2025},xticklabels={2000,,,,,2005,,,,,2010,,,,,2015,,,,,2020,,,,,2025},scaled x ticks=false,ytick={0,20,40,60},ylabel={Extractions (\%)},tick label style={font=\scriptsize},label style={font=\scriptsize},title style={font=\small\bfseries},axis lines=left,ymajorgrids,grid style={gray!15},clip=false]
\addplot[thick,teal] coordinates {(2000,26.23139) (2001,23.17814) (2002,23.87153) (2003,20.17756) (2004,23.89181) (2005,25.60423) (2006,24.06563) (2007,23.92638) (2008,23.41505) (2009,22.08801) (2010,26.42643) (2011,23.39527) (2012,24.97713) (2013,23.88759) (2014,20.25424) (2015,21.62554) (2016,18.81343) (2017,21.00840) (2018,20.43478) (2019,21.97205) (2020,23.04498) (2021,22.36760) (2022,22.28491) (2023,22.01714) (2024,23.25315) (2025,21.91419)};
\addplot[thick,orange] coordinates {(2000,33.46693) (2001,34.11978) (2002,34.60938) (2003,33.40263) (2004,34.81053) (2005,33.62012) (2006,32.72311) (2007,32.56558) (2008,32.66476) (2009,33.65173) (2010,35.27875) (2011,32.75720) (2012,32.94849) (2013,32.76501) (2014,32.40392) (2015,30.66833) (2016,31.36747) (2017,32.29959) (2018,33.27616) (2019,33.37646) (2020,32.27384) (2021,33.11966) (2022,32.10470) (2023,32.52119) (2024,31.40455) (2025,31.11472)};
\addplot[thick,violet] coordinates {(2000,33.22851) (2001,31.56450) (2002,30.08314) (2003,29.96441) (2004,32.78255) (2005,30.11583) (2006,28.54852) (2007,28.42040) (2008,31.26761) (2009,27.91347) (2010,30.46127) (2011,30.09378) (2012,32.12389) (2013,29.51945) (2014,26.73841) (2015,26.64736) (2016,26.47273) (2017,25.66735) (2018,28.77264) (2019,29.57746) (2020,29.88268) (2021,32.35646) (2022,27.57229) (2023,29.43445) (2024,28.20366) (2025,30.53385)};
\addplot[thick,blue] coordinates {(2000,36.91164) (2001,37.99176) (2002,41.37515) (2003,38.01453) (2004,38.15409) (2005,36.51714) (2006,37.28470) (2007,36.30363) (2008,37.66791) (2009,40.76563) (2010,38.31259) (2011,39.96528) (2012,37.77452) (2013,37.26610) (2014,38.05190) (2015,40.19636) (2016,36.17619) (2017,40.25559) (2018,40.21922) (2019,37.57881) (2020,37.43516) (2021,39.66883) (2022,38.16332) (2023,41.25346) (2024,40.19009) (2025,39.06582)};
\addplot[thick,plotRed] coordinates {(2000,29.66688) (2001,29.34911) (2002,29.85522) (2003,29.79749) (2004,30.09627) (2005,29.67299) (2006,30.04695) (2007,30.47890) (2008,30.06993) (2009,30.14136) (2010,30.46499) (2011,30.20022) (2012,30.00000) (2013,29.77755) (2014,29.55771) (2015,30.76923) (2016,30.05332) (2017,29.72739) (2018,29.97743) (2019,30.26450) (2020,30.05515) (2021,29.77372) (2022,30.03029) (2023,29.57576) (2024,29.35305) (2025,30.27211)};
\node[anchor=north,font=\scriptsize\bfseries,fill=white,fill opacity=.85,text opacity=1,inner sep=1pt] at (rel axis cs:.5,.98) {Cash Flow Edges};
\end{axis}\end{tikzpicture}
\par\vspace{1pt}
\begin{tikzpicture}\begin{axis}[width=0.99\columnwidth,height=3.4cm,xmin=2000,xmax=2025,ymin=0,ymax=80,xtick={2000,2001,2002,2003,2004,2005,2006,2007,2008,2009,2010,2011,2012,2013,2014,2015,2016,2017,2018,2019,2020,2021,2022,2023,2024,2025},xticklabels={2000,,,,,2005,,,,,2010,,,,,2015,,,,,2020,,,,,2025},scaled x ticks=false,ytick={0,20,40,60,80},ylabel={Extractions (\%)},tick label style={font=\scriptsize},label style={font=\scriptsize},title style={font=\small\bfseries},axis lines=left,ymajorgrids,grid style={gray!15},clip=false]
\addplot[thick,teal] coordinates {(2000,55.44101) (2001,55.76923) (2002,58.24653) (2003,63.92252) (2004,61.08189) (2005,61.70695) (2006,59.43482) (2007,60.73620) (2008,59.67878) (2009,62.20880) (2010,56.75676) (2011,59.37500) (2012,56.72461) (2013,60.88993) (2014,60.67797) (2015,60.59507) (2016,64.79313) (2017,61.55462) (2018,62.10145) (2019,58.30745) (2020,58.61592) (2021,58.69159) (2022,60.43724) (2023,60.31641) (2024,58.36197) (2025,60.52805)};
\addplot[thick,orange] coordinates {(2000,38.67735) (2001,36.02541) (2002,37.57812) (2003,38.66944) (2004,39.56326) (2005,41.03243) (2006,39.74066) (2007,40.24312) (2008,41.04585) (2009,43.42395) (2010,37.80488) (2011,38.93004) (2012,39.25400) (2013,42.10526) (2014,36.17182) (2015,35.66521) (2016,36.87622) (2017,34.93271) (2018,36.02058) (2019,34.79948) (2020,34.22983) (2021,33.81410) (2022,36.27137) (2023,34.58686) (2024,33.53116) (2025,33.60390)};
\addplot[thick,violet] coordinates {(2000,48.74214) (2001,48.76487) (2002,52.98564) (2003,57.15302) (2004,54.59352) (2005,57.78636) (2006,55.73376) (2007,58.27114) (2008,54.50704) (2009,58.12980) (2010,55.61358) (2011,51.57715) (2012,50.26549) (2013,57.28452) (2014,54.96689) (2015,54.96017) (2016,57.60000) (2017,56.39973) (2018,53.85647) (2019,54.11416) (2020,52.51898) (2021,47.84689) (2022,56.35508) (2023,52.50643) (2024,52.28833) (2025,50.45573)};
\addplot[thick,blue] coordinates {(2000,39.30636) (2001,36.08093) (2002,35.55489) (2003,38.71065) (2004,38.10150) (2005,39.49093) (2006,39.14218) (2007,39.02006) (2008,38.95399) (2009,36.49912) (2010,38.38174) (2011,35.48611) (2012,37.51830) (2013,39.99366) (2014,38.01820) (2015,36.99105) (2016,39.05030) (2017,36.15548) (2018,36.73412) (2019,37.70492) (2020,37.52161) (2021,34.36525) (2022,37.47594) (2023,34.88548) (2024,34.51007) (2025,36.22611)};
\addplot[thick,plotRed] coordinates {(2000,34.63231) (2001,34.26036) (2002,33.74938) (2003,34.61909) (2004,34.24027) (2005,34.11385) (2006,35.00261) (2007,33.88799) (2008,34.17249) (2009,34.06293) (2010,33.93907) (2011,33.81535) (2012,34.51807) (2013,34.83316) (2014,34.73571) (2015,35.00000) (2016,34.46437) (2017,34.44396) (2018,34.71783) (2019,34.53713) (2020,34.32904) (2021,33.94204) (2022,34.48723) (2023,35.03030) (2024,34.26987) (2025,34.26871)};
\node[anchor=north,font=\scriptsize\bfseries,fill=white,fill opacity=.85,text opacity=1,inner sep=1pt] at (rel axis cs:.5,.98) {Net Income Edges};
\end{axis}\end{tikzpicture}
\par\vspace{1pt}
\begin{tikzpicture}\begin{axis}[width=0.99\columnwidth,height=3.4cm,xmin=2000,xmax=2025,ymin=0,ymax=40,xtick={2000,2001,2002,2003,2004,2005,2006,2007,2008,2009,2010,2011,2012,2013,2014,2015,2016,2017,2018,2019,2020,2021,2022,2023,2024,2025},xticklabels={2000,,,,,2005,,,,,2010,,,,,2015,,,,,2020,,,,,2025},scaled x ticks=false,ytick={0,10,20,30,40},ylabel={Extractions (\%)},xlabel={Filing year},tick label style={font=\scriptsize},label style={font=\scriptsize},title style={font=\small\bfseries},axis lines=left,ymajorgrids,grid style={gray!15},clip=false]
\addplot[thick,teal] coordinates {(2000,18.32761) (2001,21.05263) (2002,17.88194) (2003,15.89992) (2004,15.02630) (2005,12.68882) (2006,16.49954) (2007,15.33742) (2008,16.90617) (2009,15.70319) (2010,16.81682) (2011,17.22973) (2012,18.29826) (2013,15.22248) (2014,19.06780) (2015,17.77939) (2016,16.39344) (2017,17.43697) (2018,17.46377) (2019,19.72050) (2020,18.33910) (2021,18.94081) (2022,17.27786) (2023,17.66645) (2024,18.38488) (2025,17.55776)};
\addplot[thick,orange] coordinates {(2000,27.85571) (2001,29.85481) (2002,27.81250) (2003,27.92793) (2004,25.62620) (2005,25.34745) (2006,27.53623) (2007,27.19130) (2008,26.28940) (2009,22.92432) (2010,26.91638) (2011,28.31276) (2012,27.79751) (2013,25.12973) (2014,31.42427) (2015,33.66646) (2016,31.75632) (2017,32.76770) (2018,30.70326) (2019,31.82406) (2020,33.49633) (2021,33.06624) (2022,31.62393) (2023,32.89195) (2024,35.06429) (2025,35.28139)};
\addplot[thick,violet] coordinates {(2000,18.02935) (2001,19.67063) (2002,16.93122) (2003,12.88256) (2004,12.62393) (2005,12.09781) (2006,15.71772) (2007,13.30846) (2008,14.22535) (2009,13.95673) (2010,13.92515) (2011,18.32907) (2012,17.61062) (2013,13.19603) (2014,18.29470) (2015,18.39247) (2016,15.92727) (2017,17.93292) (2018,17.37089) (2019,16.30838) (2020,17.59834) (2021,19.79665) (2022,16.07263) (2023,18.05913) (2024,19.50801) (2025,19.01042)};
\addplot[thick,blue] coordinates {(2000,23.78200) (2001,25.92731) (2002,23.06996) (2003,23.27482) (2004,23.74441) (2005,23.99194) (2006,23.57312) (2007,24.67631) (2008,23.37811) (2009,22.73524) (2010,23.30567) (2011,24.54861) (2012,24.70717) (2013,22.74025) (2014,23.92990) (2015,22.81259) (2016,24.77351) (2017,23.58892) (2018,23.04666) (2019,24.71627) (2020,25.04323) (2021,25.96592) (2022,24.36074) (2023,23.86106) (2024,25.29984) (2025,24.70807)};
\addplot[thick,plotRed] coordinates {(2000,35.70082) (2001,36.39053) (2002,36.39541) (2003,35.58341) (2004,35.66346) (2005,36.21316) (2006,34.95044) (2007,35.63312) (2008,35.75758) (2009,35.79571) (2010,35.59594) (2011,35.98443) (2012,35.48193) (2013,35.38928) (2014,35.70658) (2015,34.23077) (2016,35.48231) (2017,35.82865) (2018,35.30474) (2019,35.19837) (2020,35.61581) (2021,36.28424) (2022,35.48248) (2023,35.39394) (2024,36.37708) (2025,35.45918)};
\node[anchor=north,font=\scriptsize\bfseries,fill=white,fill opacity=.85,text opacity=1,inner sep=1pt] at (rel axis cs:.5,1.08) {Equity Edges};
\end{axis}\end{tikzpicture}
\par\vspace{1pt}
\caption{Annual percentage share of each model's extractions across the three edge categories, using 50 filings per year. Panel y-axis ranges differ.}
\label{fig:results_annual_allocation}
\end{figure}
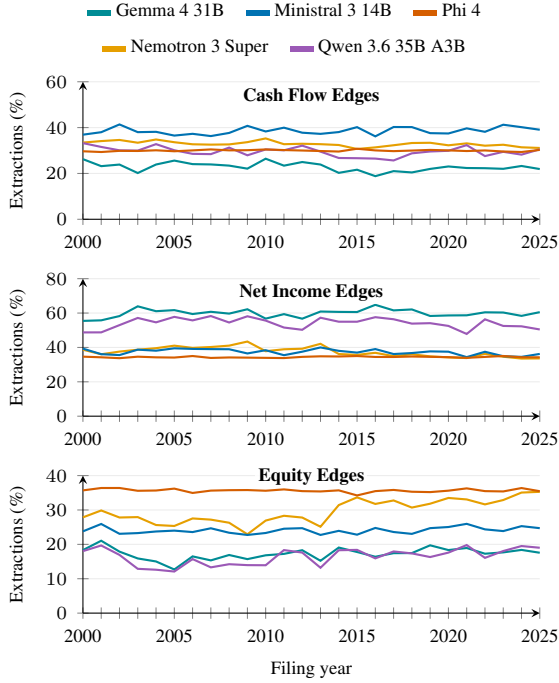

\subsection{Human-Annotated Classification Benchmark}
\label{sec:classification_benchmark}

Using the retrieved passages from Section~\ref{sec:extraction_study}, we construct a pool of 500 candidate passages from the 6 common comparison years (300 firm-year filings) that were retrieved by all models. Shared retrieval focuses annotation on graph-relevant passages. Three accounting reviewers (1 professor, 2 PhD candidates) assign one or more of the 11 edges to each passage. Inter-reviewer agreement across all 500 passages is Krippendorff's $\alpha=0.764$ using MASI distance for set-valued labels \citep{passonneau-2006-measuring}, details appear in Appendix~\ref{sec:reviewer_agreement}. Reviews define a nonempty accepted answer set $Y_i$ per passage. Review and aggregation details appear in Appendix~\ref{sec:accepted_answer_sets}. After excluding irrelevant passages, 483 passages from 259 firm-year filings remain. Models receive the passage, surrounding Item 7 context (\(\approx8192\) tokens), and schema, then predict one edge. Correctness requires membership in $Y_i$ (Section~\ref{sec:classification_task}).

\setcounter{topnumber}{3}
\setcounter{bottomnumber}{2}
\setcounter{totalnumber}{5}
\setcounter{dbltopnumber}{2}
\renewcommand{\topfraction}{0.95}
\renewcommand{\bottomfraction}{0.90}
\renewcommand{\dbltopfraction}{0.95}
\renewcommand{\textfraction}{0.05}
\renewcommand{\floatpagefraction}{0.80}
\renewcommand{\dblfloatpagefraction}{0.85}
\setlength{\textfloatsep}{8pt plus 2pt minus 2pt}
\setlength{\dbltextfloatsep}{8pt plus 2pt minus 2pt}
\setlength{\floatsep}{6pt plus 2pt minus 2pt}
\setlength{\dblfloatsep}{6pt plus 2pt minus 2pt}
\setlength{\intextsep}{7pt plus 2pt minus 2pt}


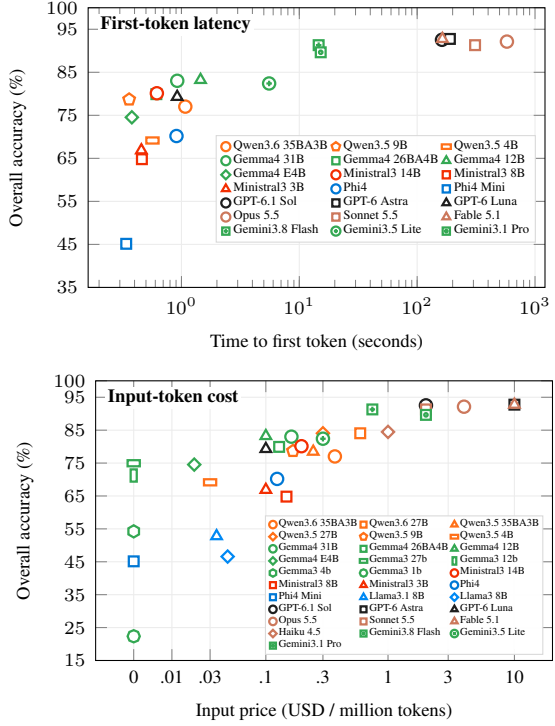
\begin{figure}[!t]\centering
\begin{tikzpicture}\begin{axis}[width=.99\columnwidth,height=5.28cm,
xlabel={Time to first token (seconds)},ylabel={Overall accuracy (\%)},
ymin=35,ymax=100,ytick={35,45,55,65,75,85,95,100},
grid=major,grid style={gray!15},tick label style={font=\scriptsize},
label style={font=\scriptsize},scaled x ticks=false,xmode=log,log basis x=10,
xticklabel style={/pgf/number format/1000 sep={}},clip=false,
legend columns=3,legend cell align=left,
legend style={at={(.98,.16)},anchor=south east,draw=gray!30,fill=white,fill opacity=.97,text opacity=1,font=\fontsize{4.8}{5.6}\selectfont,inner sep=1.5pt,nodes={inner xsep=.8pt,inner ysep=.2pt}},
legend image post style={mark size=1.7pt}]
\addplot[only marks,mark=effM1,mark size=2.3pt,line width=.85pt,mark options={fill=white},color=effColor1] coordinates {(1.078353,77.01863)};
\addlegendentry{Qwen3.6 35BA3B}
\addplot[only marks,mark=effM5,mark size=2.3pt,line width=.85pt,mark options={fill=white},color=effColor5] coordinates {(0.359954,78.67495)};
\addlegendentry{Qwen3.5 9B}
\addplot[only marks,mark=effM6,mark size=2.3pt,line width=.85pt,mark options={fill=white},color=effColor6] coordinates {(0.567587,69.15114)};
\addlegendentry{Qwen3.5 4B}
\addplot[only marks,mark=effM9,mark size=2.3pt,line width=.85pt,mark options={fill=white},color=effColor9] coordinates {(0.921271,83.02277)};
\addlegendentry{Gemma4 31B}
\addplot[only marks,mark=effM10,mark size=2.3pt,line width=.85pt,mark options={fill=white},color=effColor10] coordinates {(0.612753,79.91718)};
\addlegendentry{Gemma4 26BA4B}
\addplot[only marks,mark=effM11,mark size=2.3pt,line width=.85pt,mark options={fill=white},color=effColor11] coordinates {(1.452432,83.22981)};
\addlegendentry{Gemma4 12B}
\addplot[only marks,mark=effM12,mark size=2.3pt,line width=.85pt,mark options={fill=white},color=effColor12] coordinates {(0.379529,74.53416)};
\addlegendentry{Gemma4 E4B}
\addplot[only marks,mark=effM19,mark size=2.3pt,line width=.85pt,mark options={fill=white},color=effColor19] coordinates {(0.618655,80.12422)};
\addlegendentry{Ministral3 14B}
\addplot[only marks,mark=effM20,mark size=2.3pt,line width=.85pt,mark options={fill=white},color=effColor20] coordinates {(0.461746,64.80331)};
\addlegendentry{Ministral3 8B}
\addplot[only marks,mark=effM21,mark size=2.3pt,line width=.85pt,mark options={fill=white},color=effColor21] coordinates {(0.456937,66.87371)};
\addlegendentry{Ministral3 3B}
\addplot[only marks,mark=effM22,mark size=2.3pt,line width=.85pt,mark options={fill=white},color=effColor22] coordinates {(0.907524,70.18634)};
\addlegendentry{Phi4}
\addplot[only marks,mark=effM23,mark size=2.3pt,line width=.85pt,mark options={fill=white},color=effColor23] coordinates {(0.339904,45.13458)};
\addlegendentry{Phi4 Mini}
\addplot[only marks,mark=effM35,mark size=2.3pt,line width=.85pt,mark options={fill=white},color=effColor35] coordinates {(162.479526,92.54658)};
\addlegendentry{GPT-6.1 Sol}
\addplot[only marks,mark=effM36,mark size=2.3pt,line width=.85pt,mark options={fill=white},color=effColor36] coordinates {(189.574324,92.75362)};
\addlegendentry{GPT-6 Astra}
\addplot[only marks,mark=effM37,mark size=2.3pt,line width=.85pt,mark options={fill=white},color=effColor37] coordinates {(0.920029,79.29607)};
\addlegendentry{GPT-6 Luna}
\addplot[only marks,mark=effM38,mark size=2.3pt,line width=.85pt,mark options={fill=white},color=effColor38] coordinates {(577.578819,92.13251)};
\addlegendentry{Opus 5.5}
\addplot[only marks,mark=effM39,mark size=2.3pt,line width=.85pt,mark options={fill=white},color=effColor39] coordinates {(309.999847,91.30435)};
\addlegendentry{Sonnet 5.5}
\addplot[only marks,mark=effM40,mark size=2.3pt,line width=.85pt,mark options={fill=white},color=effColor40] coordinates {(163.814037,92.75362)};
\addlegendentry{Fable 5.1}
\addplot[only marks,mark=effM42,mark size=2.3pt,line width=.85pt,mark options={fill=white},color=effColor42] coordinates {(14.596179,91.30435)};
\addlegendentry{Gemini3.8 Flash}
\addplot[only marks,mark=effM43,mark size=2.3pt,line width=.85pt,mark options={fill=white},color=effColor43] coordinates {(5.545224,82.40166)};
\addlegendentry{Gemini3.5 Lite}
\addplot[only marks,mark=effM44,mark size=2.3pt,line width=.85pt,mark options={fill=white},color=effColor44] coordinates {(15.197533,89.64803)};
\addlegendentry{Gemini3.1 Pro}
\node[anchor=north west,font=\scriptsize\bfseries,fill=white,inner sep=1pt] at (rel axis cs:.02,.98) {First-token latency};
\end{axis}\end{tikzpicture}
\par\vspace{1pt}
\begin{tikzpicture}\begin{axis}[width=.99\columnwidth,height=5.28cm,
xlabel={Input price (USD / million tokens)},ylabel={Overall accuracy (\%)},
ymin=15,ymax=100,ytick={15,25,35,45,55,65,75,85,95,100},
grid=major,grid style={gray!15},tick label style={font=\scriptsize},
label style={font=\scriptsize},scaled x ticks=false,xtick={0,.30103,.60206,1.04139,1.49136,2.00432,2.47857,3.00043},xticklabels={0,.01,.03,.1,.3,1,3,10},
xticklabel style={/pgf/number format/1000 sep={}},clip=false,
legend columns=3,legend cell align=left,
legend style={at={(.98,.02)},anchor=south east,draw=gray!30,fill=white,fill opacity=.97,text opacity=1,font=\fontsize{4}{4.7}\selectfont,inner sep=1pt,nodes={inner xsep=.4pt,inner ysep=.1pt}},
legend image post style={mark size=1.4pt}]
\addplot[only marks,mark=effM1,mark size=2.3pt,line width=.85pt,mark options={fill=white},color=effColor1] coordinates {(1.585461,77.01863)};
\addlegendentry{Qwen3.6 35BA3B}
\addplot[only marks,mark=effM2,mark size=2.3pt,line width=.85pt,mark options={fill=white},color=effColor2] coordinates {(1.785330,84.05797)};
\addlegendentry{Qwen3.6 27B}
\addplot[only marks,mark=effM3,mark size=2.3pt,line width=.85pt,mark options={fill=white},color=effColor3] coordinates {(1.414973,78.46791)};
\addlegendentry{Qwen3.5 35BA3B}
\addplot[only marks,mark=effM4,mark size=2.3pt,line width=.85pt,mark options={fill=white},color=effColor4] coordinates {(1.491362,84.05797)};
\addlegendentry{Qwen3.5 27B}
\addplot[only marks,mark=effM5,mark size=2.3pt,line width=.85pt,mark options={fill=white},color=effColor5] coordinates {(1.255273,78.67495)};
\addlegendentry{Qwen3.5 9B}
\addplot[only marks,mark=effM6,mark size=2.3pt,line width=.85pt,mark options={fill=white},color=effColor6] coordinates {(0.602060,69.15114)};
\addlegendentry{Qwen3.5 4B}
\addplot[only marks,mark=effM9,mark size=2.3pt,line width=.85pt,mark options={fill=white},color=effColor9] coordinates {(1.243038,83.02277)};
\addlegendentry{Gemma4 31B}
\addplot[only marks,mark=effM10,mark size=2.3pt,line width=.85pt,mark options={fill=white},color=effColor10] coordinates {(1.146128,79.91718)};
\addlegendentry{Gemma4 26BA4B}
\addplot[only marks,mark=effM11,mark size=2.3pt,line width=.85pt,mark options={fill=white},color=effColor11] coordinates {(1.041393,83.22981)};
\addlegendentry{Gemma4 12B}
\addplot[only marks,mark=effM12,mark size=2.3pt,line width=.85pt,mark options={fill=white},color=effColor12] coordinates {(0.477121,74.53416)};
\addlegendentry{Gemma4 E4B}
\addplot[only marks,mark=effM14,mark size=2.3pt,line width=.85pt,mark options={fill=white},color=effColor14] coordinates {(0.000000,74.94824)};
\addlegendentry{Gemma3 27b}
\addplot[only marks,mark=effM15,mark size=2.3pt,line width=.85pt,mark options={fill=white},color=effColor15] coordinates {(0.000000,71.22153)};
\addlegendentry{Gemma3 12b}
\addplot[only marks,mark=effM16,mark size=2.3pt,line width=.85pt,mark options={fill=white},color=effColor16] coordinates {(0.000000,54.24431)};
\addlegendentry{Gemma3 4b}
\addplot[only marks,mark=effM17,mark size=2.3pt,line width=.85pt,mark options={fill=white},color=effColor17] coordinates {(0.000000,22.36025)};
\addlegendentry{Gemma3 1b}
\addplot[only marks,mark=effM19,mark size=2.3pt,line width=.85pt,mark options={fill=white},color=effColor19] coordinates {(1.322219,80.12422)};
\addlegendentry{Ministral3 14B}
\addplot[only marks,mark=effM20,mark size=2.3pt,line width=.85pt,mark options={fill=white},color=effColor20] coordinates {(1.204120,64.80331)};
\addlegendentry{Ministral3 8B}
\addplot[only marks,mark=effM21,mark size=2.3pt,line width=.85pt,mark options={fill=white},color=effColor21] coordinates {(1.041393,66.87371)};
\addlegendentry{Ministral3 3B}
\addplot[only marks,mark=effM22,mark size=2.3pt,line width=.85pt,mark options={fill=white},color=effColor22] coordinates {(1.130334,70.18634)};
\addlegendentry{Phi4}
\addplot[only marks,mark=effM23,mark size=2.3pt,line width=.85pt,mark options={fill=white},color=effColor23] coordinates {(0.000000,45.13458)};
\addlegendentry{Phi4 Mini}
\addplot[only marks,mark=effM30,mark size=2.3pt,line width=.85pt,mark options={fill=white},color=effColor30] coordinates {(0.653213,52.79503)};
\addlegendentry{Llama3.1 8B}
\addplot[only marks,mark=effM31,mark size=2.3pt,line width=.85pt,mark options={fill=white},color=effColor31] coordinates {(0.740363,46.58385)};
\addlegendentry{Llama3 8B}
\addplot[only marks,mark=effM35,mark size=2.3pt,line width=.85pt,mark options={fill=white},color=effColor35] coordinates {(2.303196,92.54658)};
\addlegendentry{GPT-6.1 Sol}
\addplot[only marks,mark=effM36,mark size=2.3pt,line width=.85pt,mark options={fill=white},color=effColor36] coordinates {(3.000434,92.75362)};
\addlegendentry{GPT-6 Astra}
\addplot[only marks,mark=effM37,mark size=2.3pt,line width=.85pt,mark options={fill=white},color=effColor37] coordinates {(1.041393,79.29607)};
\addlegendentry{GPT-6 Luna}
\addplot[only marks,mark=effM38,mark size=2.3pt,line width=.85pt,mark options={fill=white},color=effColor38] coordinates {(2.603144,92.13251)};
\addlegendentry{Opus 5.5}
\addplot[only marks,mark=effM39,mark size=2.3pt,line width=.85pt,mark options={fill=white},color=effColor39] coordinates {(2.303196,91.30435)};
\addlegendentry{Sonnet 5.5}
\addplot[only marks,mark=effM40,mark size=2.3pt,line width=.85pt,mark options={fill=white},color=effColor40] coordinates {(3.000434,92.75362)};
\addlegendentry{Fable 5.1}
\addplot[only marks,mark=effM41,mark size=2.3pt,line width=.85pt,mark options={fill=white},color=effColor41] coordinates {(2.004321,84.47205)};
\addlegendentry{Haiku 4.5}
\addplot[only marks,mark=effM42,mark size=2.3pt,line width=.85pt,mark options={fill=white},color=effColor42] coordinates {(1.880814,91.30435)};
\addlegendentry{Gemini3.8 Flash}
\addplot[only marks,mark=effM43,mark size=2.3pt,line width=.85pt,mark options={fill=white},color=effColor43] coordinates {(1.491362,82.40166)};
\addlegendentry{Gemini3.5 Lite}
\addplot[only marks,mark=effM44,mark size=2.3pt,line width=.85pt,mark options={fill=white},color=effColor44] coordinates {(2.303196,89.64803)};
\addlegendentry{Gemini3.1 Pro}
\node[anchor=north west,font=\scriptsize\bfseries,fill=white,inner sep=1pt] at (rel axis cs:.02,.98) {Input-token cost};
\end{axis}\end{tikzpicture}
\par\vspace{2pt}

\caption{Overall accuracy versus first-token latency (top) and input price (bottom), for models with available Artificial Analysis data as of 10 October 2026. Colors identify providers. Horizontal positions use $\log_{10}t$ for latency and $\log_{10}(1+c/0.01)$ for cost, with $t$ in seconds and $c$ in USD per million input tokens.}
\label{fig:results_overall_efficiency}
\end{figure}

\subsection{Model Coverage and Evaluation Setup}
\label{sec:model_coverage}

Task I compares representative models of Gemma, Nemotron, Qwen, Ministral, and Phi open-weight families against Gemini, Claude, and GPT proprietary references. Task II broadens the comparison to 34 open-weight models from six families - Gemma, Qwen, Ministral, Phi, Llama, and DeepSeek, and ten proprietary models from the Gemini (Google), Claude (Anthropic), and GPT (OpenAI) families. Model sizes and generations vary within the open-weight families. For Task II, model selection is motivated by locally deployable processing of large financial-text corpora: all 34 open-weight classifiers have fewer than 36B total parameters. All models are used in instruct-mode with reasoning/thinking disabled where possible or set to the lowest supported level for proprietary models for uniform analysis. Each open-weight model was evaluated in BF16 precision on a single NVIDIA H100 GPU with 80\,GB of memory. This coverage permits both family-level and within-family comparisons. Models within each task use a common task formulation and requested output format for uniformity. Task-specific inference configurations and context settings are detailed in Appendices~\ref{app:retrieval_protocol} and~\ref{app:classification_protocol}.

\section{Results}
\label{sec:results}
\raggedbottom
\setcounter{topnumber}{1}
\setcounter{bottomnumber}{1}
\setcounter{totalnumber}{2}
\setcounter{dbltopnumber}{1}


\subsection{Evidence Allocation and Cross-Model Agreement}

Figure~\ref{fig:results_annual_allocation} shows each model's percentage share of extractions across the three edge categories. Distinct allocation profiles persist across all 26 years: greater relative emphasis on Equity edges (Phi and Nemotron), Net Income edges (Gemma and Qwen), and Cash Flow edges (Ministral). A common graph therefore guides retrieval without imposing a common emphasis.

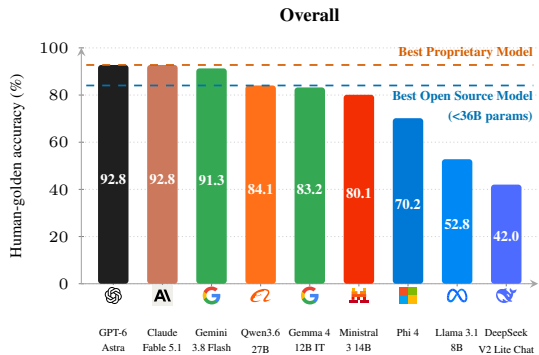
\begin{figure}[b]\centering
\begin{tikzpicture}\begin{axis}[width=.99\columnwidth,height=4.7cm,
title={Overall},title style={font=\scriptsize\bfseries},
xmin=-.65,xmax=8.65,ymin=0,ymax=100,
ytick={0,20,40,60,80,100},ylabel={Human-golden accuracy (\%)},
xtick={0,1,2,3,4,5,6,7,8},xticklabels={{\shortstack{GPT-6\\Astra}},{\shortstack{Claude\\Fable 5.1}},{\shortstack{Gemini\\3.8 Flash}},{\shortstack{Qwen3.6\\27B}},{\shortstack{Gemma 4\\12B IT}},{\shortstack{Ministral\\3 14B}},{\shortstack{Phi 4}},{\shortstack{Llama 3.1\\8B}},{\shortstack{DeepSeek\\V2 Lite Chat}}},
xticklabel style={rotate=0,anchor=north,font=\fontsize{4}{4.8}\selectfont,yshift=-12pt},
yticklabel style={font=\tiny},label style={font=\tiny},ylabel style={yshift=-2pt},major tick length=2pt,
axis lines=left,axis line style={gray!70},ymajorgrids=true,
grid style={gray!35,densely dotted},clip=false]
\path[fill=familyBaropenai,rounded corners=1.2pt] (axis cs:-0.31,0) rectangle (axis cs:0.31,92.75362);
\node[text=white,font=\fontsize{5.5}{6}\selectfont\bfseries] at (axis cs:0,44.52174) {92.8};
\node[inner sep=0pt] at (axis cs:0,-5) {\scalebox{.7}{\providerLogoopenai}};
\path[fill=familyBaranthropic,rounded corners=1.2pt] (axis cs:0.69,0) rectangle (axis cs:1.31,92.75362);
\node[text=white,font=\fontsize{5.5}{6}\selectfont\bfseries] at (axis cs:1,44.52174) {92.8};
\node[inner sep=0pt] at (axis cs:1,-5) {\scalebox{.7}{\providerLogoanthropic}};
\path[fill=familyBargoogle,rounded corners=1.2pt] (axis cs:1.69,0) rectangle (axis cs:2.31,91.30435);
\node[text=white,font=\fontsize{5.5}{6}\selectfont\bfseries] at (axis cs:2,43.82609) {91.3};
\node[inner sep=0pt] at (axis cs:2,-5) {\scalebox{.7}{\providerLogogoogle}};
\path[fill=familyBaralibaba,rounded corners=1.2pt] (axis cs:2.69,0) rectangle (axis cs:3.31,84.05797);
\node[text=white,font=\fontsize{5.5}{6}\selectfont\bfseries] at (axis cs:3,40.34783) {84.1};
\node[inner sep=0pt] at (axis cs:3,-5) {\scalebox{.7}{\providerLogoalibaba}};
\path[fill=familyBargoogle,rounded corners=1.2pt] (axis cs:3.69,0) rectangle (axis cs:4.31,83.22981);
\node[text=white,font=\fontsize{5.5}{6}\selectfont\bfseries] at (axis cs:4,39.95031) {83.2};
\node[inner sep=0pt] at (axis cs:4,-5) {\scalebox{.7}{\providerLogogoogle}};
\path[fill=familyBarmistral,rounded corners=1.2pt] (axis cs:4.69,0) rectangle (axis cs:5.31,80.12422);
\node[text=white,font=\fontsize{5.5}{6}\selectfont\bfseries] at (axis cs:5,38.45963) {80.1};
\node[inner sep=0pt] at (axis cs:5,-5) {\scalebox{.7}{\providerLogomistral}};
\path[fill=familyBarazure,rounded corners=1.2pt] (axis cs:5.69,0) rectangle (axis cs:6.31,70.18634);
\node[text=white,font=\fontsize{5.5}{6}\selectfont\bfseries] at (axis cs:6,33.68944) {70.2};
\node[inner sep=0pt] at (axis cs:6,-5) {\scalebox{.7}{\providerLogoazure}};
\path[fill=familyBarmeta,rounded corners=1.2pt] (axis cs:6.69,0) rectangle (axis cs:7.31,52.79503);
\node[text=white,font=\fontsize{5.5}{6}\selectfont\bfseries] at (axis cs:7,25.34161) {52.8};
\node[inner sep=0pt] at (axis cs:7,-5) {\scalebox{.7}{\providerLogometa}};
\path[fill=familyBardeepseek,rounded corners=1.2pt] (axis cs:7.69,0) rectangle (axis cs:8.31,42.02899);
\node[text=white,font=\fontsize{5.5}{6}\selectfont\bfseries] at (axis cs:8,20.17391) {42.0};
\node[inner sep=0pt] at (axis cs:8,-5) {\scalebox{.7}{\providerLogodeepseek}};
\draw[dashed,line width=.65pt,color=familyBestPro] (axis cs:-.55,92.75362) -- (axis cs:8.55,92.75362);
\node[anchor=south east,yshift=2pt,text=familyBestPro,font=\fontsize{5}{5.5}\selectfont\bfseries,fill=none,inner sep=0pt] at (axis cs:8.55,92.75362) {Best Proprietary Model};
\draw[dashed,line width=.65pt,color=familyBestOpen] (axis cs:-.55,84.05797) -- (axis cs:8.55,84.05797);
\node[anchor=north east,yshift=-2pt,text=familyBestOpen,font=\fontsize{5}{5.5}\selectfont\bfseries,fill=none,inner sep=0pt] at (axis cs:8.55,84.05797) {\shortstack[r]{Best Open Source Model\\(\textless36B params)}};
\end{axis}\end{tikzpicture}
\caption{Best overall classification accuracy within each model family (values from Table \ref{tab:results_human_leaderboard}). Dashed lines show the strongest proprietary and open-weight models.}
\label{fig:results_family_best_overall}
\end{figure}

Extraction volume matters when comparing models with reference outputs (Table~\ref{tab:results_reference_agreement}). Ministral produces the most extractions and achieves the highest retrieval recall against every reference. Broader extraction creates more opportunities to recover reference passages, so recall alone can favor verbose outputs. Jaccard also accounts for unmatched extractions: Ministral trails Gemma and Qwen against Gemini and GPT, although it leads against Sonnet. Conditional classification precision asks whether shared evidence receives the same edge assignment. Gemma leads consistently here, showing that retrieval coverage and agreement on financial interpretation are distinct strengths.

\subsection{Tag--Edge Consistency}

High tag--edge purity means that a model repeatedly associates a tag with the same accounting relationship, making its vocabulary more predictable for analysts (Table~\ref{tab:results_tag_organization}). Qwen leads the open-weight models in purity and has the lowest conditional entropy, with Gemma close in homogeneity. Ministral and Phi generate expansive vocabularies with many one-off tags, suggesting passage-specific descriptions rather than reusable relation labels. In contrast, Qwen reuses a compact vocabulary with higher purity, showing that lexical variety need not imply consistent alignment with the graph schema.


\subsection{Human-Grounded Classification}

Category-level accuracy reveals where the open-weight gap is concentrated (Table~\ref{tab:results_human_leaderboard}). Cash Flow edges are the strongest challenge, with a 14.3-point gap between the best open-weight and proprietary models. Net Income edges are more competitive, with several open-weight models approaching proprietary performance and a best-model gap of 2.8 points. Equity edges lie between these cases at 6.4 points. This pattern identifies cash-flow interpretation as a priority for improving locally deployable models. Domain-expert feedback suggests that varied expressions for the same transaction may contribute to this difficulty, such as a dividend in one filing being described as a distribution in another. This lexical variation may complicate alignment with the fixed accounting schema.

Overall, the strongest open-weight models reach 84.1\% accuracy, 8.7 points below the proprietary leaders. Qwen and Gemma lead the open-weight group across categories and overall (Fig.~\ref{fig:results_family_best_overall}) under our instruction-following settings, while larger models and more extensive reasoning remain outside this small-to-mid-sized model comparison. The bounded 11-edge task offers a concrete target for relation-specific training to narrow this gap.

\paragraph{Shared and model-specific errors.}
Equal accuracy does not imply identical failures. Qwen3.5-27B and Qwen3.6-27B both score 84.1\%, but share only 60 of their 77 errors, leaving 17 unique errors each. Qwen3.6-27B and Gemma4-12B have 58 shared errors (failure-set Jaccard: 58.0\%). Errors therefore combine recurring challenges with model-specific weaknesses, as illustrated in Appendix~\ref{sec:failure_overlap}.

Figure~\ref{fig:results_overall_efficiency} complements these accuracy comparisons with externally reported latency and input prices, helping analysts weigh relation-classification performance against hosted deployment requirements.


\section{Conclusion and Practical Recommendations}

Open-weight models show competitive reference-relative retrieval (RQ1), while both model groups retain room for semantic consistency and schema alignment (RQ2). Qwen3.5/3.6 and Gemma4, notably Gemma4-12B, lead open-weight classification (RQ3), approaching proprietary models on Net Income edges but lagging on Cash Flow edges and overall. The external API comparison in Fig.~\ref{fig:results_overall_efficiency} suggests that modest accuracy tradeoffs can offer lower input costs and faster first-token responses, providing practical alternatives for financial workflows. Model selection should balance overall accuracy, category-specific task performance, model size, cost, and first-token latency.

\section{Limitations}

Our evaluation focuses on narrative evidence in annual US SEC Form 10-K filings under a fixed accounting graph. It measures evidence discovery and relation classification, rather than numerical reasoning, table interpretation, or general financial decision-making. Other reporting regimes and document types remain directions for extending the benchmark.

Task I measures agreement with proprietary reference-model extractions, not exhaustive human-labeled retrieval coverage because of the large corpus. Task II provides human-grounded classification evaluation on passages selected through shared retrieval, focusing annotation effort on graph-relevant evidence rather than a random sample of filing text.

Model comparisons reflect the evaluated instruction-following configurations, with limited reasoning and locally deployable open-weight classifiers. Additional reasoning may change model rankings while increasing latency and cost. High-volume workflows motivate our instruction-following setting, although practitioners may choose reasoning-enabled configurations for other use cases. Larger models, task-specific fine-tuning, and additional reasoning may also yield different tradeoffs. External API prices and first-token latencies provide deployment context from a verified source (Artificial Analysis), not controlled measurements of our task-specific runtimes.

\section{Ethical Considerations}

This study uses publicly available financial disclosures from Item 7 of SEC Form 10-K filings. The evaluation dataset was manually annotated by faculty members and doctoral researchers with relevant domain expertise (with industry and academic experience in accounting), as part of their regular research responsibilities and compensated academic appointments. No external crowdsourcing was employed.

Although the underlying documents are publicly accessible, automated evidence extraction and classification may introduce errors in identifying financial relationships. Accordingly, the resulting annotations and model outputs are intended for research and evaluation purposes rather than automated financial decision-making.


\bibliography{custom}

\appendix
\setcounter{bottomnumber}{0}
\renewcommand{\dblfloatpagefraction}{0.55}

\clearpage
\section{Accounting Schema and Filing Examples}
\label{app:accounting_schema}

The graph represents Cash Flow edges (E1--E3), Net Income edges (E4--E7), and Equity edges (E8--E11). Table~\ref{tab:appendix_edge_definitions} gives the accounting meaning of each relation. Classification identifies the relationship between its endpoints, irrespective of the direction of a particular transaction.

\begin{table*}[b]\centering\small
\begin{tabularx}{\textwidth}{@{}llX@{}}\toprule
Edge & Relation & Financial Interpretation \\ \midrule
E1 & Financing--Net Non-cash Assets & Financing flows associated with changes in non-cash assets and liabilities, such as borrowing and debt repayment.\\
E2 & Investment--Net Non-cash Assets & Investing flows associated with acquiring or disposing of non-cash assets.\\
E3 & Operations--Net Non-cash Assets & Operating cash flows and their connection to working capital and non-cash adjustments.\\
E4 & Revenue--Net Non-cash Assets & Revenue recognition and related changes in non-cash assets or liabilities.\\
E5 & Expenses--Net Non-cash Assets & Expense recognition and related changes in non-cash assets or liabilities.\\
E6 & Interest Expenses--Net Non-cash Assets & Interest expense and associated accruals or other non-cash balance changes.\\
E7 & Tax--Net Non-cash Assets & Tax expense and associated tax assets and liabilities.\\
E8 & OCI--Net Non-cash Assets & Other comprehensive income and associated non-cash balance changes.\\
E9 & Financing--Dividends & Cash distributions to shareholders.\\
E10 & Non-cash Equity & Contributed equity associated with non-cash assets or liabilities, including non-cash equity transactions.\\
E11 & Cash Equity & Financing cash flows associated with contributed equity, including share issuance and repurchases.\\
\bottomrule\end{tabularx}
\caption{Definitions of the 11 accounting edges. OCI denotes other comprehensive income.}
\label{tab:appendix_edge_definitions}
\end{table*}

\subsection{From a Filing to Item 7}
The supplied Apple 2025 Form 10-K illustrates the document structure. Item 7 contains MD\&A, while Item 8 contains the financial statements and accompanying notes. The screenshots are illustrative and do not establish that this particular filing was sampled for the benchmark.

\begin{figure*}[t]\centering
\IfFileExists{appendix_assets/apple_10k_contents.png}{%
{\setlength{\fboxsep}{0pt}\setlength{\fboxrule}{0.8pt}%
\fbox{\includegraphics[width=.96\textwidth]{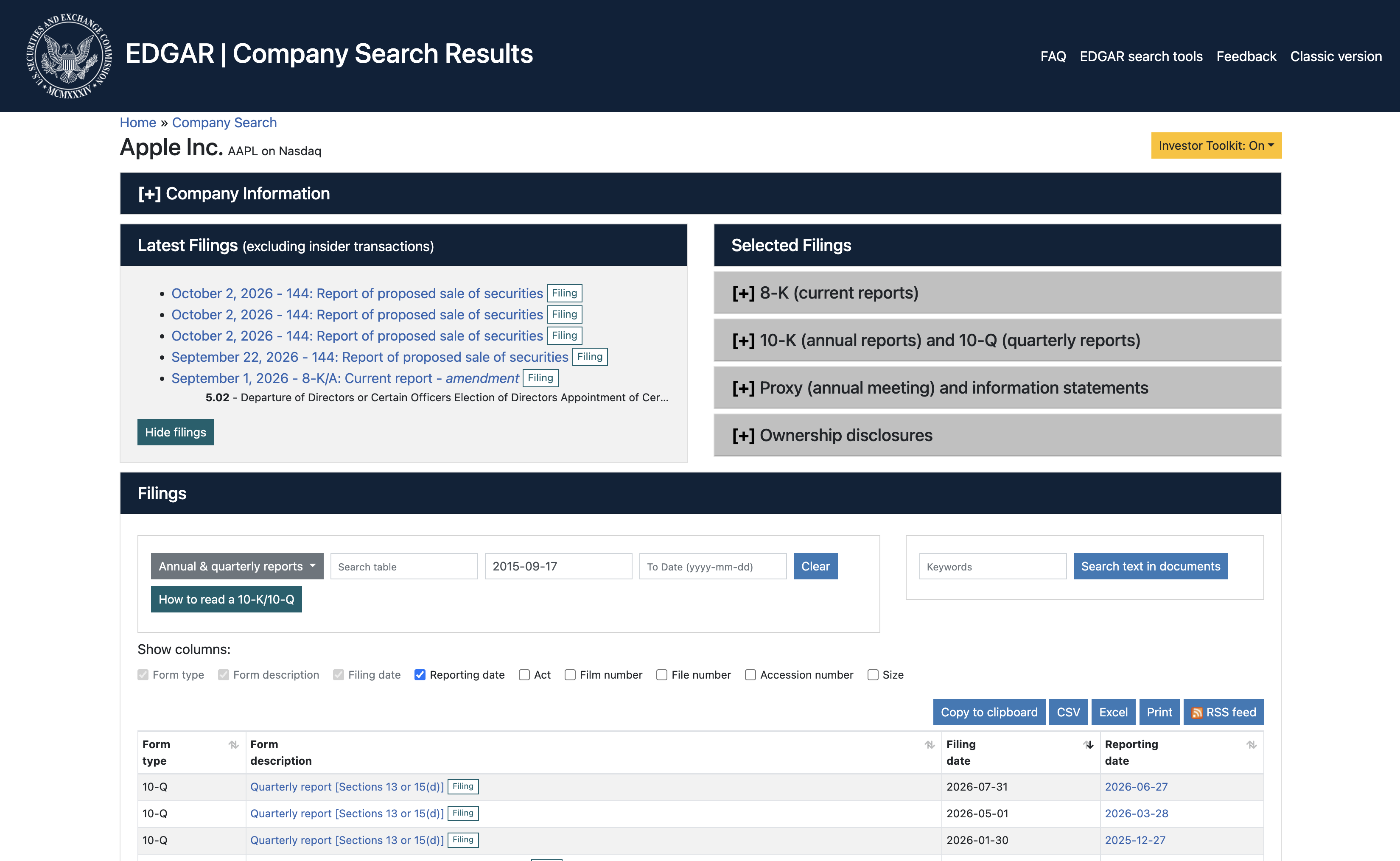}}\par\medskip
\fbox{\includegraphics[width=.96\textwidth]{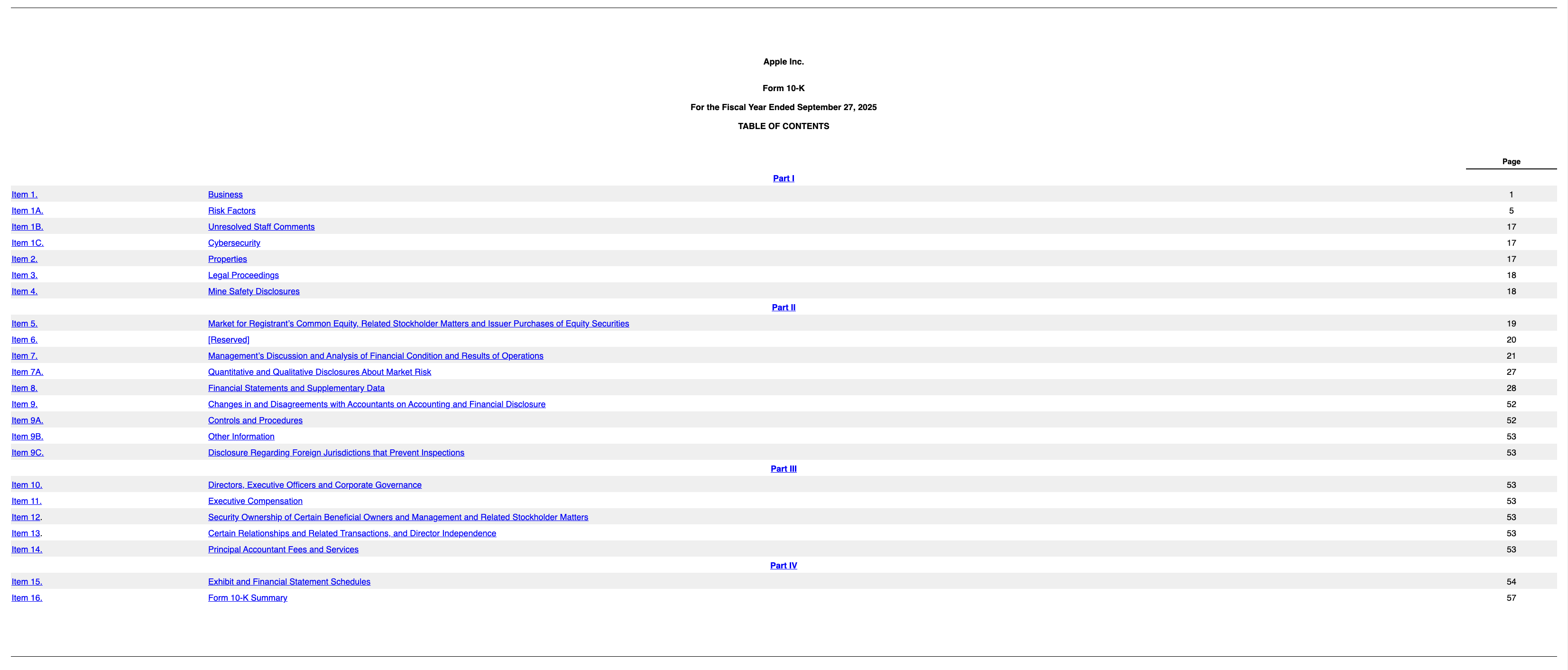}}\par\medskip
\fbox{\includegraphics[width=.96\textwidth]{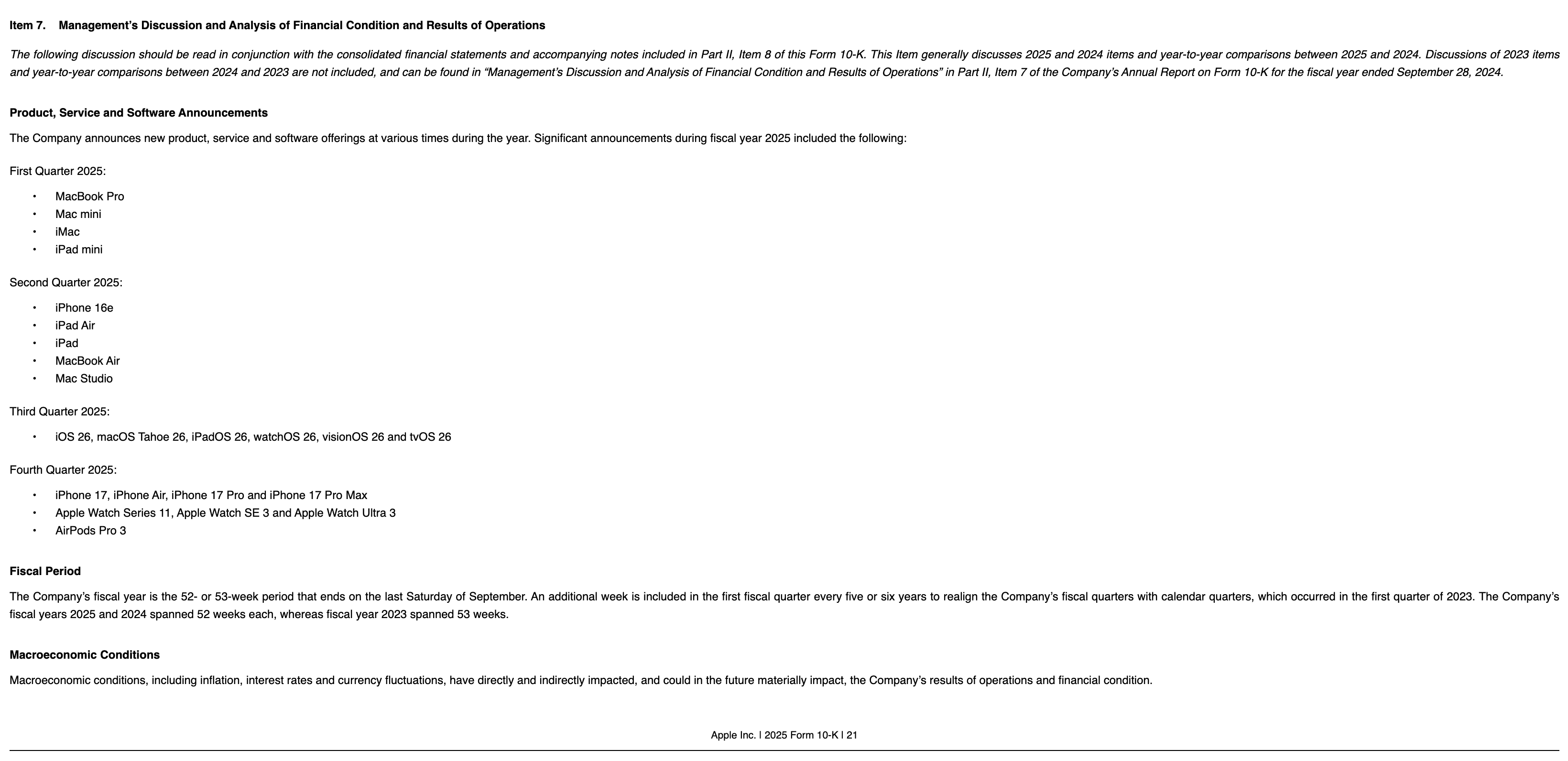}}}%
}{%
\begin{appendixquote}{Illustrative filing structure}
Apple Inc., 2025 Form 10-K. The contents identify Item 7 as Management's Discussion and Analysis of Financial Condition and Results of Operations, beginning on page 21, and Item 8 as Financial Statements and Supplementary Data, beginning on page 28. Item 7 combines narrative headings, discussion, and financial context.
\end{appendixquote}}
\caption{Apple's Form 10-K contents and opening Item 7 page.}
\label{fig:appendix_filing_example}
\end{figure*}

\section{Corpus Characteristics and SEC-Based Firm Size}
\label{sec:appendix}

\subsection{Coverage and Industry Composition}
\label{app:industry_composition}

The corpus contains 1,300 distinct company-year filings from 2000--2025,
with 50 filings per year and 1,237 distinct SEC Central Index Keys (CIKs).
Industry metadata were retrieved from the SEC submissions API and grouped
into broad Standard Industrial Classification (SIC) divisions.\footnote{%
\href{https://www.sec.gov/search-filings/edgar-application-programming-interfaces}{SEC EDGAR API documentation};
\href{https://www.sec.gov/search-filings/standard-industrial-classification-sic-code-list}{SEC SIC code list}.}
All registrants were looked up: 1,222 have a classified industry and 15
remain unknown or unclassified, corresponding to 17 filings. No records
are removed from the distribution. These are current SEC classifications,
not verified historical filing-year classifications.

Nine broad divisions are represented. Manufacturing, finance, insurance
and real estate, and services account for 76.0\% of filings; the corpus
has industry breadth, but is not industry-balanced.
Table~\ref{tab:appendix_industry} distinguishes registrant counts from
filing counts so repeated appearances do not inflate the number of firms.

\begin{table}[t]\centering\footnotesize
\setlength{\tabcolsep}{3pt}
\begin{tabularx}{\columnwidth}{@{}>{\raggedright\arraybackslash}Xrr@{}}
\toprule
Industry division & Registrants & Filings\\
\midrule
Manufacturing & 463 & 486\\
Finance, insurance and real estate & 253 & 269\\
Services & 221 & 233\\
Transportation, communications and utilities & 116 & 123\\
Retail trade & 67 & 68\\
Mining & 58 & 60\\
Wholesale trade & 26 & 26\\
Unknown / unclassified & 15 & 17\\
Construction & 15 & 15\\
Agriculture, forestry and fishing & 3 & 3\\
Total & 1,237 & 1,300\\
\bottomrule\end{tabularx}
\caption{Full-corpus industry composition using current SEC SIC metadata. Registrants are distinct CIKs; filings are distinct CIK--accession pairs.}
\label{tab:appendix_industry}
\end{table}

\subsection{Reported Total Assets, 2012--2025}
\label{app:asset_size}

We purposefully restrict asset-size analysis to the 700 selected filings
from 2012--2025, where SEC structured-data coverage is substantially more
complete. Total reported assets are obtained through the SEC company-concept
API using \texttt{us-gaap:Assets}. This restriction concerns size
characterization only: all 1,300 filings remain in the extraction corpus
and industry analysis. The full-corpus API audit recovered 656 exact-filing
values and left 644 unknown; only the 647 recovered values in the chosen
2012--2025 window are analyzed below.

Values are matched to the selected 10-K's exact accession and its latest
balance-sheet date, without substituting a comparative prior-period balance
or another filing. Values are available for 647 of 700 filings (92.4\%);
53 remain explicitly unknown. The window contains 675 distinct registrants,
of which 623 have matched values. Table~\ref{tab:appendix_asset_coverage}
reports annual coverage.

Total assets measure balance-sheet scale, not equity market capitalization.
We use descriptive asset-size bands rather than small-, mid-, or large-cap
classes. Values are nominal USD, not inflation-adjusted; differences in
asset intensity across industries limit interpretation as a universal
measure of company size.

Among matched filings, median assets are \$568.7 million (interquartile
range: \$61.4 million--\$3.00 billion). The 95th percentile is \$37.56 billion
and the maximum is \$3.17 trillion. Three filings report zero assets;
these are retained as reported values, not marked as missing.

\begin{table}[t]\centering\footnotesize
\setlength{\tabcolsep}{3pt}
\begin{tabularx}{\columnwidth}{@{}>{\raggedright\arraybackslash}Xrr@{}}
\toprule
Total assets (USD) & Filings & Share (\%)\\
\midrule
Below \$100 million & 189 & 29.2\\
\$100 million--below \$1 billion & 187 & 28.9\\
\$1 billion--below \$10 billion & 192 & 29.7\\
At least \$10 billion & 79 & 12.2\\
Matched total & 647 & 100.0\\
\bottomrule\end{tabularx}
\caption{Asset-size distribution for 2012--2025. Percentages use the 647 matched filings; the 53 unknown values remain in coverage counts. These are not market-capitalization classes.}
\label{tab:appendix_asset_bands}
\end{table}
\begin{table}[t]\centering\footnotesize
\setlength{\tabcolsep}{3pt}
\begin{tabular}{@{}lrrr@{}}
\toprule
Year & Selected & Matched & Unknown\\
\midrule
2012 & 50 & 42 & 8\\
2013 & 50 & 45 & 5\\
2014 & 50 & 47 & 3\\
2015 & 50 & 49 & 1\\
2016 & 50 & 47 & 3\\
2017 & 50 & 44 & 6\\
2018 & 50 & 44 & 6\\
2019 & 50 & 45 & 5\\
2020 & 50 & 48 & 2\\
2021 & 50 & 45 & 5\\
2022 & 50 & 48 & 2\\
2023 & 50 & 48 & 2\\
2024 & 50 & 48 & 2\\
2025 & 50 & 47 & 3\\
Total & 700 & 647 & 53\\
\bottomrule\end{tabular}
\caption{SEC API asset coverage by dataset year. The asset date is the balance-sheet date in the selected filing, not necessarily calendar year-end.}
\label{tab:appendix_asset_coverage}
\end{table}

\subsection{Item 7 Document Length}
\label{app:document_length}

Recorded processed-input lengths are available for all 1,300 documents.
The median is 57,395.5 characters (interquartile range:
36,616.5--78,265 characters); the range is 2,543--501,163 characters.
These are text-input character counts, not HTML byte sizes, page counts,
or tokenizer-dependent token counts.

\begin{table}[t]\centering\footnotesize
\setlength{\tabcolsep}{3pt}
\begin{tabular}{@{}lr@{}}
\toprule
Statistic & Input characters\\
\midrule
Minimum & 2,543\\
25th percentile & 36,616.5\\
Median & 57,395.5\\
75th percentile & 78,265\\
95th percentile & 136,180.1\\
Maximum & 501,163\\
\bottomrule\end{tabular}
\caption{Processed Item 7 input lengths across all 1,300 filings. Percentiles are computed over filings, not over model outputs.}
\label{tab:appendix_document_lengths}
\end{table}

\section{Task I: Retrieval Protocol}
\label{app:retrieval_protocol}

Each document is divided into approximately 4,096-token chunks while retaining natural text boundaries. A shared instruction prompt requests quoted evidence, graph endpoints, descriptive tags, and supported attributes. The following box reproduces an archived system prompt, including the original schema terminology. In quoted prompts, \texttt{contributed} or ``Contributed Capital'' denotes the concept called Contributed Equity in the paper.
\begin{figure*}[p]\centering
\begin{appendixquote}[fontupper={\fontsize{7.4}{8.1}\selectfont\raggedright\sloppy},before skip=0pt,after skip=0pt]{Task I: complete archived prompt}
\setlength{\columnsep}{14pt}\setlength{\parskip}{1pt}
\begin{multicols}{2}
You are a financial expert specializing in document analysis and information extraction.
\par
Your task is to extract structured information from unstructured financial documents with high accuracy.
\par
Guidelines:
\par
1. Be precise and extract only information explicitly stated in the document
\par
2. When extracting related information, maintain logical relationships
\par
Output Format:
\par
- Return valid JSON only (no markdown, no code blocks)
\par
- Follow the provided schema structure exactly
\par
Task:
\par
Find evidence in this 10-K filing that bears on the candidate financial
\par
relationships below. The evidence may be direct or suggestive. The goal
\par
is to surface useful evidence for providing context for  an edge in the graph.
\par
Prefer faithful evidence over confident conclusions.
\par
Consider these candidate bi-directional edges:
\par
- revenue \texttt{<-->} net\_\allowbreak{}noncash\_\allowbreak{}assets
\par
- net\_\allowbreak{}noncash\_\allowbreak{}assets \texttt{<-->} cash\_\allowbreak{}flow\_\allowbreak{}from\_\allowbreak{}operations
\par
- net\_\allowbreak{}noncash\_\allowbreak{}assets \texttt{<-->} cash\_\allowbreak{}flow\_\allowbreak{}from\_\allowbreak{}finances
\par
- cash\_\allowbreak{}flow\_\allowbreak{}from\_\allowbreak{}finances \texttt{<-->} dividends
\par
- net\_\allowbreak{}noncash\_\allowbreak{}assets \texttt{<-->} tax
\par
- net\_\allowbreak{}noncash\_\allowbreak{}assets \texttt{<-->} int\_\allowbreak{}expense
\par
- net\_\allowbreak{}noncash\_\allowbreak{}assets \texttt{<-->} expense
\par
- cash\_\allowbreak{}flow\_\allowbreak{}from\_\allowbreak{}investments \texttt{<-->} net\_\allowbreak{}noncash\_\allowbreak{}assets
\par
- other\_\allowbreak{}comprehensive\_\allowbreak{}income \texttt{<-->} net\_\allowbreak{}noncash\_\allowbreak{}assets
\par
- contributed \texttt{<-->} net\_\allowbreak{}noncash\_\allowbreak{}assets
\par
- cash\_\allowbreak{}flow\_\allowbreak{}from\_\allowbreak{}finances \texttt{<-->} contributed
\par
Evidence rules:
\par
- extraction\_\allowbreak{}text MUST be an exact, contiguous quote from the document.
\par
- Do not use "...." to quote text. Try to quote the full text verbatim.
\par
- Direct evidence explicitly describes the relationship.
\par
- Suggestive evidence may connect relevant concepts through a table,
\par
reconciliation, comparison, disclosure, or nearby explanation even when
\par
it does not establish causality. Label this with
\par
evidence\_\allowbreak{}strength: suggestive.
\par
- Ontology mapping is allowed when it is a reasonable interpretation of
\par
the quoted financial disclosure, but do not invent facts absent from the
\par
quote.
\par
Attributes rules:
\par
- For the atributes, you may include any relevant structured information
\par
that is explicitly supported by the quote. You can also use one of these:
\par
- impact\_\allowbreak{}direction --- increase, decrease, favorable, unfavorable, unclear
\par
- relationship\_\allowbreak{}type --- drives, includes, offsets, associated\_\allowbreak{}with, supports
\par
- evidence\_\allowbreak{}strength --- direct, suggestive
\par
- amount --- absolute monetary quantity
\par
- percentage --- a stated percentage, e.g. 14
\par
- rate --- tax rate, interest rate, discount rate, etc.
\par
- change\_\allowbreak{}amount --- absolute change, e.g. \$10.7 billion decrease
\par
- change\_\allowbreak{}percentage --- relative change, e.g. 8\% decrease
\par
- period --- fiscal\_\allowbreak{}year, quarter, next\_\allowbreak{}12\_\allowbreak{}months, etc.
\par
- comparison\_\allowbreak{}period --- 2025\_\allowbreak{}vs\_\allowbreak{}2024
\par
- driver --- the reason for the financial change
\par
- impact\_\allowbreak{}area --- list of affected areas, e.g. ["revenue", "gross\_\allowbreak{}margin"]
\par
- instrument --- notes, commercial paper, lease, etc.
\par
- expense\_\allowbreak{}type --- R\&D, SG\&A, interest, tax, etc.
\par
- asset\_\allowbreak{}type --- inventory, PP\&E, receivables, intangible assets, etc.
\par
- cash\_\allowbreak{}flow\_\allowbreak{}type --- operating, investing, financing
\par
- maturity --- maturity date/period when explicitly stated
\par
- authorization\_\allowbreak{}amount --- useful for repurchase programs and credit facilities
\par
- outlook --- forward-looking expectation or management outlook
\par
- Try to use multiple attributes to create granular structured information.
\par
For example, convert "\$10.7 billion" to amount: 10700000000 and currency: USD.
\par
- Try to include description as an attribute which should be a very short (1-2
\par
sentence) explaination of why the quote is relevant to the edge.
\par
Extraction class rules:
\par
- Use extraction\_\allowbreak{}class to describe the type of annotation. It may be one
\par
label or a list of applicable labels. You can devise
\par
your own extraction class tags or use one of the following:
\par
revenue\_\allowbreak{}driver, revenue\_\allowbreak{}growth, revenue\_\allowbreak{}mix, volume\_\allowbreak{}change,
\par
price\_\allowbreak{}change, foreign\_\allowbreak{}exchange\_\allowbreak{}impact,  cost\_\allowbreak{}driver,
\par
operating\_\allowbreak{}expense\_\allowbreak{}driver, gross\_\allowbreak{}margin\_\allowbreak{}driver, margin\_\allowbreak{}impact,
\par
working\_\allowbreak{}capital\_\allowbreak{}change, operating\_\allowbreak{}cash\_\allowbreak{}flow\_\allowbreak{}driver,
\par
investing\_\allowbreak{}cash\_\allowbreak{}flow\_\allowbreak{}driver, financing\_\allowbreak{}cash\_\allowbreak{}flow\_\allowbreak{}driver,
\par
capital\_\allowbreak{}expenditure, asset\_\allowbreak{}investment, debt\_\allowbreak{}financing,
\par
equity\_\allowbreak{}financing, dividend\_\allowbreak{}payment, share\_\allowbreak{}repurchase,
\par
tax\_\allowbreak{}rate\_\allowbreak{}driver, interest\_\allowbreak{}rate\_\allowbreak{}impact,  liquidity\_\allowbreak{}source,
\par
liquidity\_\allowbreak{}use, contractual\_\allowbreak{}obligation,  macroeconomic\_\allowbreak{}impact
\par
- You may classify one extraction into multiple classes in a list. However,
\par
try to use the most specific class possible.
\par
Additional Notes:
\par
- Always include extraction\_\allowbreak{}class, extraction\_\allowbreak{}text, source\_\allowbreak{}node, and target\_\allowbreak{}node.
Include attributes when the quote supports additional structured details;
\par
otherwise omit the optional attributes field.
\par
- extraction\_\allowbreak{}class must describe the evidence or relationship identified.
\par
- source\_\allowbreak{}node and target\_\allowbreak{}node MUST match one of the candidate edges above.
\par
- Attributes are not fixed - you can extract as many as long as they are
\par
supported by the evidence.
\par
- Try to find as many extractions as possible, as long as they are supported by
\par
the document.
\par
- We are interested in specific evidence as opposed to generic statements.
\par
- You can extract multiple extractions for the same edge.
\par
- Some extractions may be qualitative, while others may be quantitative. Include
\par
all relevant extractions.
\par
- Return ONLY the valid JSON, nothing else.
\par
MUST return valid JSON with this structure:
\par
\{
\par
"extractions": [
\par
\{
\par
"extraction\_\allowbreak{}class": "Revenue\_\allowbreak{}impact\textbar{}\allowbreak{}tax\_\allowbreak{}expense\textbar{}\allowbreak{}operating\_\allowbreak{}expense\textbar{}\allowbreak{}dividend\_\allowbreak{}impact\textbar{}\allowbreak{}etc",
\par
"extraction\_\allowbreak{}text": "exact quote from document",
\par
"source\_\allowbreak{}node": "source item from list",
\par
"target\_\allowbreak{}node": "target item from list",
\par
"attributes": \{
\par
....
\par
\}
\par
\}
\par
]
\par
\}
\par
Return ONLY valid JSON, no additional text.
\par\medskip\textbf{User-message template}\par
Document:
\par
[Item 7 document chunk]

\end{multicols}
\end{appendixquote}
\end{figure*}

The bracketed field denotes variable source text, not a literal instruction. Outputs are validated as structured extractions with admissible graph endpoints. Prompt examples were disabled in the illustrated request.

\subsection{Saved Extraction Example}
The following saved Qwen3.6-35B-A3B extraction corresponds to the viewer screenshot for Bain Capital Specialty Finance. Its financial period is quoted as generated, rather than inferred from the filing year.
\begin{appendixquote}{Task I: saved model response}
\{\par
  "source\_\allowbreak{}node": "cash\_\allowbreak{}flow\_\allowbreak{}from\_\allowbreak{}investments",\par
  "target\_\allowbreak{}node": "net\_\allowbreak{}noncash\_\allowbreak{}assets",\par
  "extraction\_\allowbreak{}text": "The decrease in cash used for operating activities was primarily related to purchases of investments of \$1,626.0 million, which was offset by proceeds from principal payments and sales of investments of \$1,486.3 million and a net increase in assets resulting from operations of \$119.4 million.",\par
  "extraction\_\allowbreak{}class": "asset\_\allowbreak{}investment",\par
  "attributes": \{\par
    "amount": 1626000000,\par
    "currency": "USD",\par
    "period": "fiscal\_\allowbreak{}year\_\allowbreak{}2024",\par
    "asset\_\allowbreak{}type": "investments",\par
    "impact\_\allowbreak{}direction": "decrease",\par
    "relationship\_\allowbreak{}type": "drives",\par
    "evidence\_\allowbreak{}strength": "direct",\par
    "description": "Large purchases of investments (noncash assets) are the primary driver of cash usage in this context."\par
  \}\par
\}\par
\end{appendixquote}

\begin{figure*}[t]\centering
\IfFileExists{appendix_assets/extraction_viewer.png}{%
{\setlength{\fboxsep}{0pt}\setlength{\fboxrule}{0.8pt}%
\fbox{\includegraphics[width=.96\textwidth]{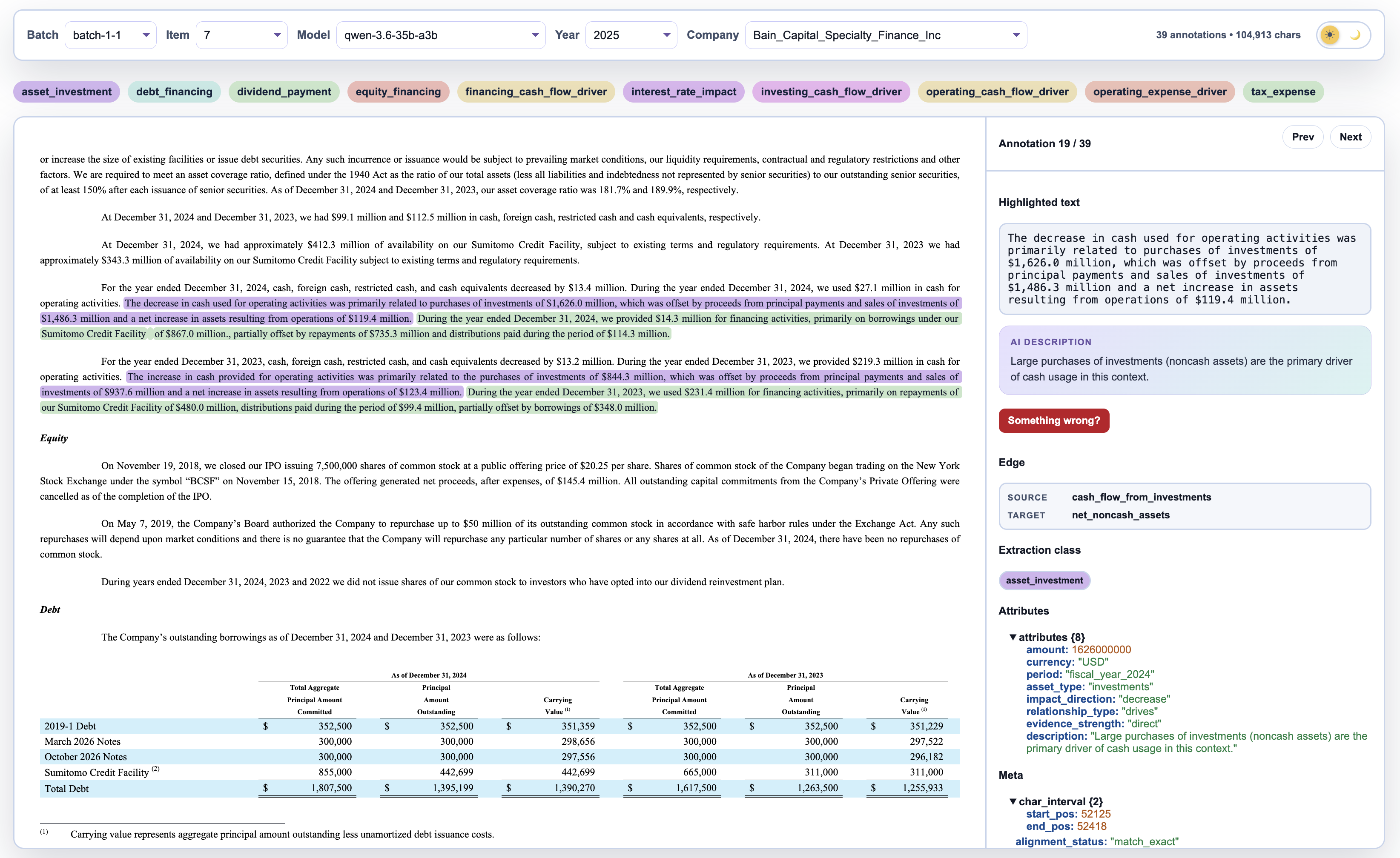}}}
}{\begin{appendixquote}{Extraction viewer}
The source passage is highlighted beside its model-assigned edge, descriptive tags, and supported attributes. The viewer permits inspection of each extraction in its original filing context.
\end{appendixquote}}
\caption{Task I extraction viewer: source evidence, assigned relation, tags, and attributes.}
\label{fig:appendix_extraction_viewer}
\end{figure*}
\begin{table}[t]\centering\scriptsize
\setlength{\tabcolsep}{3pt}
\begin{tabularx}{\columnwidth}{@{}Xrrr@{}}\toprule
Model & Temperature & Top-p & Output cap \\ \midrule
Gemma 4 31B & 0.7 & 0.9 & 16384 \\
Nemotron 3 Super & 1.0 & 0.95 & 16384 \\
Qwen 3.6 35B A3B & 0.7 & 0.9 & 9216 \\
Ministral 3 14B & 0.15 & 0.95 & 8192 \\
Phi 4 & 0.15 & 0.95 & 5120 \\
\midrule
Gemini 3.1 Flash Lite & 0.7 & 0.9 & 16384 \\
Claude Sonnet 5 & -- & -- & 32768 \\
GPT-6 Sol & -- & -- & 32768 \\
\bottomrule\end{tabularx}
\caption{Representative archived Task I configurations. Caps are generation tokens. ``--'' denotes a sampling field omitted by the provider request. Configuration values are not asserted to be uniform across every run.}
\label{tab:appendix_task1_settings}
\end{table}
\section{Human Annotation and Reference Construction}

\subsection{Annotation Interface}
Reviewers inspect highlighted evidence within its filing context and select all supported relationships, or None alone. Practice submissions are separate from the evaluation records. The screenshots illustrate the interface, not an adjudicated answer.
\begin{figure*}[t]\centering
\IfFileExists{appendix_assets/human_review.png}{%
{\setlength{\fboxsep}{0pt}\setlength{\fboxrule}{0.8pt}%
\fbox{\includegraphics[width=.9\textwidth]{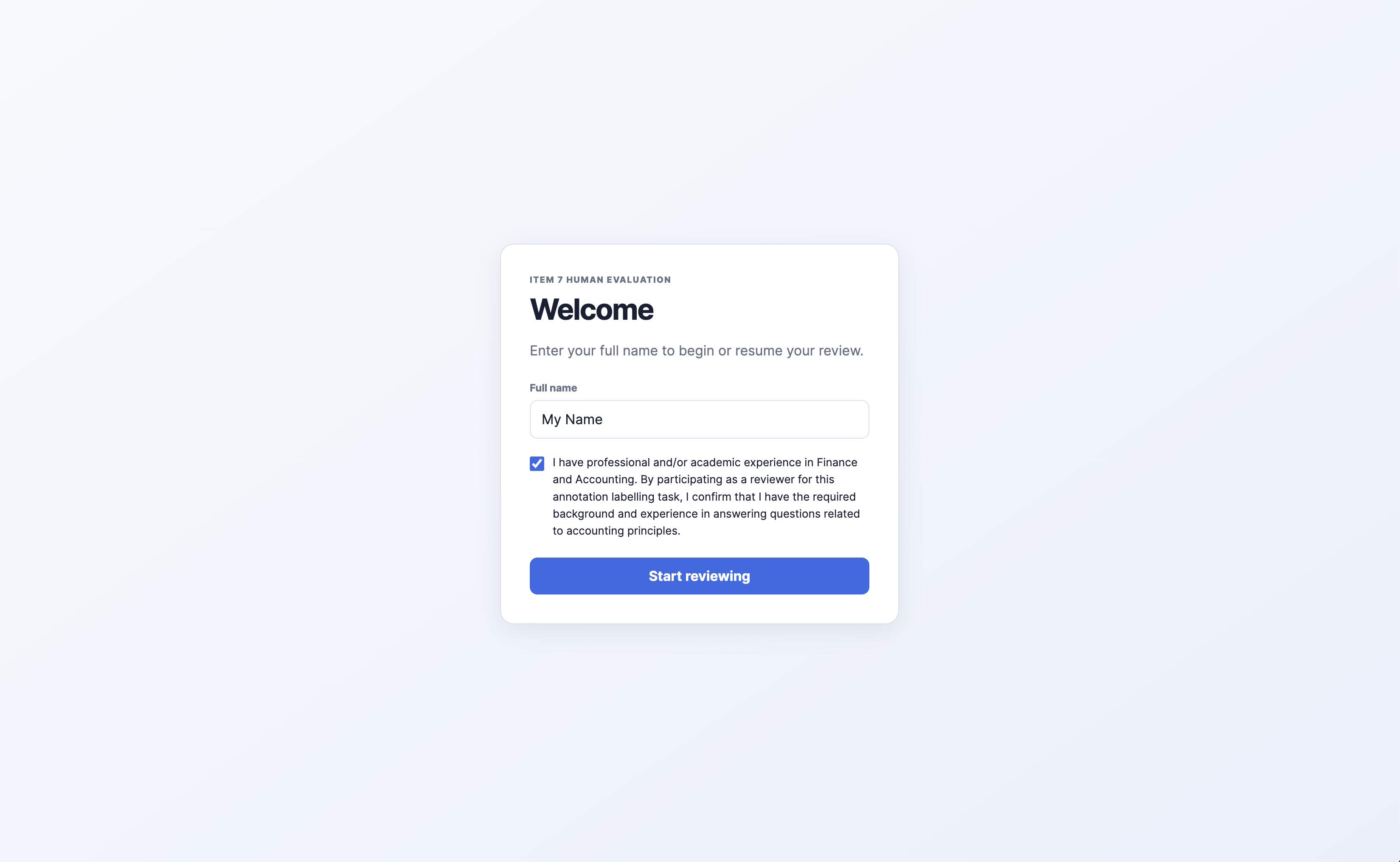}}\\[0.75em]
\fbox{\includegraphics[width=.9\textwidth]{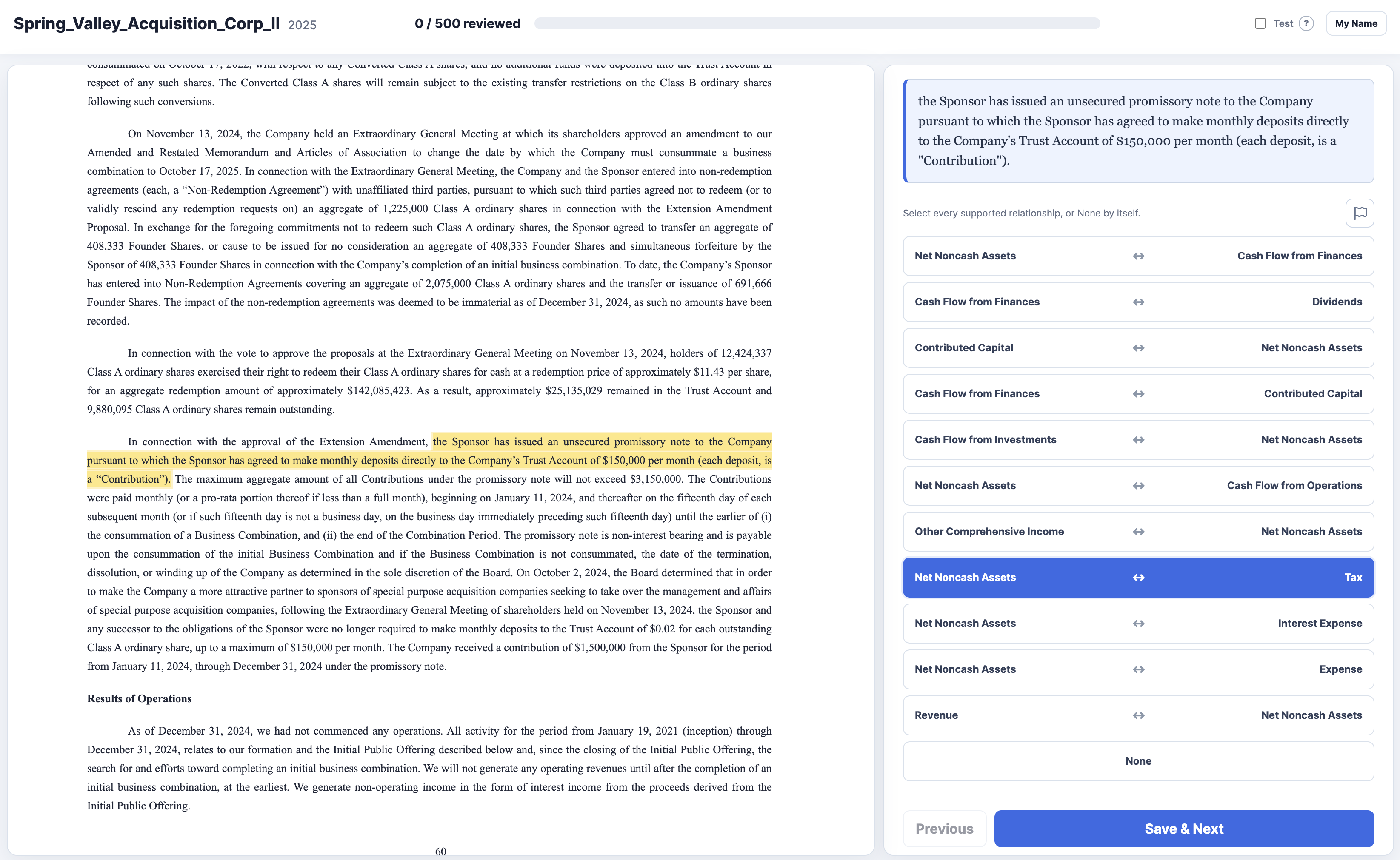}}}%
}{\begin{appendixquote}{Human annotation interface}
The review screen presents source context and highlighted evidence alongside the 11 selectable accounting relationships. Reviewers can select multiple relationships, choose None alone, flag a passage, and revise their saved selections.
\end{appendixquote}}
\caption{Human review interface: entry screen and passage-level relationship selection.}
\label{fig:appendix_human_interface}
\end{figure*}

\subsection{Inter-Reviewer Agreement}
\label{sec:reviewer_agreement}

Agreement is calculated from the three reviewers' latest non-test submissions
for all 500 candidate passages (1,500 decisions), before constructing golden
answer sets or excluding the 17 majority-None passages. Each review is a set
of selected edges; None is an empty set. This preserves multi-edge selections
rather than reducing them to a single label.

We report Krippendorff's $\alpha$ with MASI distance
\citep{passonneau-2006-measuring}, which distinguishes identical, nested,
partially overlapping, and disjoint sets. For sets $A$ and $B$,
$d(A,B)=1-m(A,B)|A\cap B|/|A\cup B|$, where $m$ is 1 for identical
sets, 0.67 for proper containment, 0.33 for non-nested overlap, and 0
for disjoint sets. These weights follow the conventional NLTK implementation.
The distance between two empty sets is defined as 0; between an empty
and a nonempty set it is 1.

\begin{equation}
\alpha=1-\frac{D_o}{D_e}.
\end{equation}
$D_o$ is the mean distance over the three reviewer pairs for each passage
(1,500 pairs). $D_e$ is the mean distance over all distinct pairs of decisions
in the pooled 1,500 ratings, without replacement. We obtain
$D_o=0.213385$, $D_e=0.903824$, and $\alpha=0.764$.

As a descriptive complement, Table~\ref{tab:reviewer_agreement} reports
exact-set agreement and mean Jaccard similarity $|A\cap B|/|A\cup B|$,
with similarity 1 for two empty sets. Mean pairwise Jaccard is 80.6\%;
all three reviewers select identical sets on 331/500 passages (66.2\%).
These statistics describe agreement, not annotation correctness.

\begin{table}[t]\centering\footnotesize
\setlength{\tabcolsep}{4pt}
\begin{tabular}{@{}lrr@{}}\toprule
Reviewer pair & Exact sets (\%) & Jaccard (\%)\\
\midrule
R1--R2 & 70.0 & 75.9\\
R1--R3 & 72.2 & 78.8\\
R2--R3 & 83.6 & 87.1\\
\bottomrule\end{tabular}
\caption{Inter-reviewer agreement on all 500 candidate passages.}
\label{tab:reviewer_agreement}
\end{table}

\subsection{Accepted Answer Sets and Passage Context}
\label{sec:accepted_answer_sets}

Each reviewer may select multiple edges or None. An edge supported by at least two of the three reviewers enters the primary answer set. If that set is empty, the answer set is the union of the reviewers' selected edges. Passages with majority None are excluded. This yields 472 primary-set passages, 11 fallback-set passages, and 17 exclusions from the 500 candidates. The final 483 passages come from 259 firm-year filings. Of these, 398 accept one edge, 64 accept two, 19 accept three, and 2 accept four. Thus, 85 passages have multiple accepted edges. Any edge in the accepted set earns classification credit.

Classification inputs preserve the evidence passage and its enclosing section, paragraph, or local source window within an 8,192-token total budget. The exact prompt is documented in Appendix~\ref{app:classification_protocol}.

\section{Task II: Classification Protocol}
\label{app:classification_protocol}

Models receive the evidence passage and source context under a total 8,192-token budget. Context is drawn from the enclosing section when it fits, otherwise from the paragraph or a bounded local window. Thus, 8,192 denotes a budget rather than a fixed input length. The system prompt below is reproduced from the saved evaluation dataset.
\begin{appendixquote}{Task II: system prompt}
Classify the evidence into exactly one of these undirected financial relationships. Return only the complete relationship name exactly as written below.
\par\smallskip
- Net Noncash Assets <--> Cash Flow from Finances
\par\smallskip
- Cash Flow from Finances <--> Dividends
\par\smallskip
- Contributed Capital <--> Net Noncash Assets
\par\smallskip
- Cash Flow from Finances <--> Contributed Capital
\par\smallskip
- Cash Flow from Investments <--> Net Noncash Assets
\par\smallskip
- Net Noncash Assets <--> Cash Flow from Operations
\par\smallskip
- Other Comprehensive Income <--> Net Noncash Assets
\par\smallskip
- Net Noncash Assets <--> Tax
\par\smallskip
- Net Noncash Assets <--> Interest Expense
\par\smallskip
- Net Noncash Assets <--> Expense
\par\smallskip
- Revenue <--> Net Noncash Assets
\end{appendixquote}\begin{appendixquote}{Task II: user-message template}
Evidence text:
\par\smallskip
[Quoted evidence passage]
\par\smallskip
Item 7 context:
\par\smallskip
[Source context]
\end{appendixquote}
The model returns one listed relationship. Saved predictions are checked against the corresponding prompt and model identity before scoring. All 44 classifiers contribute predictions on the same 483 eligible passages.
Open-weight instruction-following classifiers used temperature 0, top-p 0.95, a 128-token output cap, and reasoning off. Proprietary requests used low reasoning for Claude Fable 5.1, Opus 5.5, Sonnet 5.5, Gemini 3.1 Pro, Gemini 3.8 Flash, GPT-6 Astra, and GPT-6.1 Sol. Reasoning-enabled batches generally reserved 4,096 output tokens, with Sonnet 5.5 retaining a 128-token cap. Claude Haiku 4.5 disabled thinking, and GPT-6 Luna used reasoning \texttt{none}, both with 128-token caps. Gemini 3.5 Flash Lite used a 4,096-token cap without an explicit reasoning override. Sampling controls omitted by proprietary APIs remained provider-managed.

\section{Metric Definitions and Statistical Procedures}
\label{sec:results_metric_details}

Cross-model evidence matches are one-to-one. Retrieval recall, evidence Jaccard, and conditional classification precision measure agreement with reference-model outputs rather than human retrieval completeness. Tag--edge association scores use model-supplied strings without case or synonym normalization, with each tagged extraction contributing total weight one. The used-once column counts tag types appearing in a single extraction. Vocabulary Jaccard uses mechanically normalized distinct tag sets and measures lexical rather than semantic overlap.

Human classification compares each prediction with the full accepted answer set. Category cohorts can overlap, whereas overall accuracy counts each passage once. All models have 483 predictions. Family winners and medal highlights summarize observed scores rather than statistical significance.

The Artificial Analysis snapshot records external hosted-service measurements, not controlled Task II runtimes or local GPU inference costs. Latency uses 1K-token prompts and provider reasoning configurations may differ from ours. Listed zero prices denote free API offerings. All models with available measurements are shown, with 21 latency observations and 31 input-price observations. Missing measurements are omitted.

\subsection{Evidence Matching and Vocabulary Normalization}

Evidence is matched within the same filing, without using predicted edges. Quotes undergo Unicode and punctuation normalization, case folding, and whitespace consolidation. The overlap score $s(a,b)$ combines exact agreement, contiguous containment, shared sentence fragments, and conservative token-order overlap. Maximum-weight one-to-one assignment accepts pairs with $s(a,b)\geq0.80$, preventing repeated extractions from matching the same reference passage. This measures textual overlap rather than semantic equivalence or numerical fidelity.

For vocabulary overlap, let $\nu(t)$ be the mechanical normalization of a tag and $V_m=\{\nu(t):t\text{ is a valid tag from model }m\}$. Normalization standardizes Unicode, separates CamelCase, folds case, harmonizes separators, expands a fixed list of financial abbreviations, and applies curated spelling and plural corrections. It does not use embedding similarity, general stemming, or semantic-bucket merging. Malformed strings and placeholder labels are excluded from vocabulary sets.
\begin{equation}
J_{\mathrm{vocab}}(a,b)=\frac{|V_a\cap V_b|}{|V_a\cup V_b|}.
\end{equation}
Tag--edge purity, entropy, and homogeneity instead retain the model's supplied tag strings. An extraction with $k$ tags contributes weight $1/k$ to each tag--edge pair, so each tagged extraction contributes total weight one.

\subsection{Baselines and Confidence Intervals}

For accepted answer set $Y_i$ and prediction $\hat e_i$, correctness is $\mathbb{1}[\hat e_i\in Y_i]$. A uniform random selector chooses among all 11 edges, giving the exact expected accuracy
\begin{equation}
\operatorname{Acc}_{\mathrm{random}}(S)
=\frac{1}{|S|}\sum_{i\in S}\frac{|Y_i|}{11}.
\end{equation}
The same expression applies to category or edge cohorts $S$. The majority-selector baseline always predicts the most frequent accepted edge in the benchmark.

Accuracy intervals use a source-filing cluster bootstrap. We sample filings with replacement and retain every passage belonging to each sampled filing, preserving within-filing dependence. Each replicate recomputes the ratio of correct predictions to sampled passages, rather than averaging filing accuracies. The 2.5th and 97.5th percentiles of 2,000 replicates give the 95\% interval, using seed 20261010. Paired comparisons resample the same filings for both models. Intervals describe sampling variability conditional on the benchmark labels.

\section{Extended Quantitative Results}

\par\medskip\noindent\begin{minipage}{\columnwidth}\centering\footnotesize
\setlength{\tabcolsep}{3pt}
\begin{tabularx}{\columnwidth}{@{}Xrr@{}}\toprule
Model & Accuracy (\%) & 95\% CI \\ \midrule
Claude-fable-5-1 & 92.8 & 90.2--94.9 \\
Claude-haiku-4-5 & 84.5 & 81.1--87.8 \\
Claude-opus-5-5 & 92.1 & 89.6--94.5 \\
Claude-sonnet-5-5 & 91.3 & 88.7--93.8 \\
Gemini-3.1-pro-preview & 89.6 & 86.7--92.2 \\
Gemini-3.5-flash-lite & 82.4 & 78.9--85.7 \\
Gemini-3.8-flash & 91.3 & 88.5--93.8 \\
GPT-6-astra & 92.8 & 90.3--95.1 \\
GPT-6-luna & 79.3 & 75.6--82.9 \\
GPT-6.1-sol & 92.5 & 90.0--94.9 \\
DeepSeek-V2-Lite-Chat & 42.0 & 37.8--46.7 \\
Llama-3.1-8B-Instruct & 52.8 & 48.1--57.4 \\
Llama-3.2-1B-Instruct & 23.0 & 19.1--27.1 \\
Llama-3.2-3B-Instruct & 48.0 & 43.4--52.8 \\
Meta-Llama-3-8B-Instruct & 46.6 & 41.9--51.4 \\
Ministral-3-14B-Instruct & 80.1 & 76.8--83.5 \\
Ministral-3-3B-Instruct & 66.9 & 62.6--71.0 \\
Ministral-3-8B-Instruct & 64.8 & 60.6--69.2 \\
Phi-3-medium-128k-instruct & 19.5 & 16.0--23.0 \\
Phi-3-mini-128k-instruct & 49.5 & 45.2--54.1 \\
Phi-3-small-128k-instruct & 58.8 & 54.4--63.2 \\
Phi-3.5-mini-instruct & 54.7 & 50.3--58.8 \\
Phi-4 & 70.2 & 66.0--74.2 \\
Phi-4-mini-instruct & 45.1 & 40.9--49.3 \\
Qwen3.5-0.8B & 36.4 & 32.2--41.2 \\
Qwen3.5-27B & 84.1 & 80.5--87.4 \\
Qwen3.5-2B & 41.8 & 37.1--46.7 \\
Qwen3.5-35B-A3B & 78.5 & 74.9--82.2 \\
Qwen3.5-4B & 69.2 & 64.8--73.7 \\
Qwen3.5-9B & 78.7 & 75.0--82.4 \\
Qwen3.6-27B & 84.1 & 80.7--87.3 \\
Qwen3.6-35B-A3B & 77.0 & 73.4--80.9 \\
DeepSeek-llm-7b-chat & 20.9 & 16.7--25.1 \\
DeepSeek-moe-16b-chat & 15.9 & 13.0--19.3 \\
Gemma-3-12b-it & 71.2 & 66.8--75.2 \\
Gemma-3-1b-it & 22.4 & 18.6--26.4 \\
Gemma-3-270m-it & 8.9 & 6.5--11.5 \\
Gemma-3-27b-it & 74.9 & 70.7--78.8 \\
Gemma-3-4b-it & 54.2 & 49.9--58.7 \\
Gemma-4-12B-it & 83.2 & 79.9--86.4 \\
Gemma-4-26B-A4B-it & 79.9 & 76.2--83.5 \\
Gemma-4-31B-it & 83.0 & 79.5--86.6 \\
Gemma-4-E2B-it & 65.2 & 60.7--69.6 \\
Gemma-4-E4B-it & 74.5 & 70.7--78.1 \\
\bottomrule\end{tabularx}
\appendixcaption{table}{Overall classification accuracy and source-filing cluster-bootstrap intervals on the common human-golden benchmark.}
\label{tab:appendix_accuracy_intervals}
\end{minipage}\par\medskip

\subsection{Edge Supports and Baselines}
Category cohorts contain passages with at least one accepted edge in that category. A passage may therefore belong to more than one category, while overall accuracy counts it once. Edge-level scores also test predictions against the complete accepted set, not only the edge defining the cohort.

\par\medskip\noindent\begin{minipage}{\columnwidth}\centering\small
\setlength{\tabcolsep}{3pt}
\begin{tabularx}{\columnwidth}{@{}Xrrr@{}}\toprule
Edge & Passages & Majority (\%) & Random (\%) \\ \midrule
E1 & 82 & 39.0 & 15.6 \\
E2 & 70 & 2.9 & 12.7 \\
E3 & 66 & 9.1 & 13.6 \\
E4 & 11 & 0.0 & 14.9 \\
E5 & 72 & 1.4 & 10.6 \\
E6 & 37 & 0.0 & 10.6 \\
E7 & 60 & 0.0 & 10.0 \\
E8 & 17 & 17.6 & 13.4 \\
E9 & 44 & 36.4 & 15.7 \\
E10 & 29 & 6.9 & 13.2 \\
E11 & 103 & 100.0 & 14.6 \\
\bottomrule\end{tabularx}
\appendixcaption{table}{Edge cohorts and exact baseline accuracies. Cohorts may overlap.}
\label{tab:appendix_edge_baselines}
\end{minipage}\par\medskip

\subsection{Category-Level Retrieval Agreement}
Retrieval recall and conditional classification precision are also reported within reference-edge categories. The matches are computed once over the filing and then partitioned by the reference assignment, avoiding separate rematching within each category.

\par\medskip\noindent\begin{minipage}{\columnwidth}\centering\footnotesize
\setlength{\tabcolsep}{2pt}
\begin{tabularx}{\columnwidth}{@{}Xrrrrrr@{}}\toprule
Model & \shortstack{CF\\Recall} & \shortstack{CF\\CCP} & \shortstack{NI\\Recall} & \shortstack{NI\\CCP} & \shortstack{EQ\\Recall} & \shortstack{EQ\\CCP} \\ \midrule
\multicolumn{7}{@{}l}{\textbf{Gemini 3.1 Flash Lite}} \\
Gemma 4 31B & 66.5 & 75.2 & 66.2 & 85.7 & 77.7 & 69.4 \\
Nemotron 3 Super & 58.4 & 72.3 & 42.9 & 58.0 & 70.4 & 47.7 \\
Qwen 3.6 35B A3B & 62.5 & 74.1 & 62.8 & 76.5 & 78.2 & 58.4 \\
Ministral 3 14B & 80.5 & 67.1 & 74.3 & 51.2 & 85.9 & 40.3 \\
Phi 4 & 44.7 & 62.2 & 38.2 & 33.4 & 61.4 & 42.1 \\
\midrule
\multicolumn{7}{@{}l}{\textbf{Claude Sonnet 5}} \\
Gemma 4 31B & 34.8 & 72.7 & 36.9 & 88.4 & 46.4 & 76.4 \\
Nemotron 3 Super & 37.3 & 64.4 & 30.0 & 63.0 & 46.5 & 52.2 \\
Qwen 3.6 35B A3B & 35.1 & 70.6 & 36.8 & 78.8 & 49.1 & 63.6 \\
Ministral 3 14B & 59.7 & 60.1 & 57.1 & 55.2 & 68.4 & 42.0 \\
Phi 4 & 33.7 & 54.1 & 28.6 & 43.2 & 41.9 & 42.2 \\
\midrule
\multicolumn{7}{@{}l}{\textbf{GPT-6 Sol}} \\
Gemma 4 31B & 35.0 & 72.5 & 36.2 & 83.8 & 50.4 & 73.4 \\
Nemotron 3 Super & 34.4 & 64.6 & 23.0 & 53.7 & 48.6 & 50.2 \\
Qwen 3.6 35B A3B & 34.6 & 69.3 & 35.2 & 74.2 & 52.3 & 59.9 \\
Ministral 3 14B & 57.2 & 56.4 & 52.4 & 47.2 & 71.8 & 39.4 \\
Phi 4 & 33.3 & 51.2 & 24.9 & 31.5 & 45.3 & 42.1 \\
\bottomrule\end{tabularx}
\appendixcaption{table}{Reference-relative retrieval recall and conditional classification precision (CCP), in percent. CF, NI, and EQ denote Cash Flow, Net Income, and Equity edges.}
\label{tab:appendix_category_agreement}
\end{minipage}\par\medskip

\subsection{Majority Voting on Shared Evidence}
The ensemble uses passages matched across all five open-weight extractors. Each model casts one edge vote, and at least three matching votes yield a prediction. Split votes abstain. The following comparison evaluates individual models and the ensemble on the same majority-decidable passages, against each proprietary reference.

\par\medskip\noindent\begin{minipage}{\columnwidth}\centering\small\setlength{\tabcolsep}{3pt}
\begin{tabularx}{\columnwidth}{@{}Xrrr@{}}\toprule
Model & Gemini & Claude & GPT \\ \midrule
Gemma 4 31B & 85.7 & 90.8 & 88.4 \\
Nemotron 3 Super & 72.5 & 78.6 & 75.7 \\
Qwen 3.6 35B A3B & 77.5 & 81.6 & 79.9 \\
Ministral 3 14B & 65.7 & 72.4 & 73.4 \\
Phi 4 & 55.2 & 64.5 & 63.6 \\
Majority ensemble & 79.9 & 87.3 & 84.8 \\
\bottomrule\end{tabularx}\appendixcaption{table}{Edge agreement (\%) on the same majority-decidable evidence: 364, 543, and 448 passages for the three references, respectively.}\label{tab:appendix_ensemble}\end{minipage}\par\medskip

\subsection{Vocabulary Reuse and Cross-Task Stability}
Tag recurrence distinguishes a reusable vocabulary from one-off labels. Cross-task stability asks a different question: whether a model preserves its edge assignment when the same retrieved evidence is supplied for classification.

\par\medskip\noindent\begin{minipage}{\columnwidth}\centering\footnotesize\setlength{\tabcolsep}{3pt}
\begin{tabularx}{\columnwidth}{@{}Xrrrrrr@{}}\toprule
Model & Once & Twice & 3--5 & 6--10 & 11--50 & $>50$ \\ \midrule
Gemma 4 31B & 91 & 15 & 11 & 11 & 12 & 25 \\
Nemotron 3 Super & 357 & 71 & 72 & 26 & 31 & 31 \\
Qwen 3.6 35B A3B & 38 & 12 & 7 & 2 & 9 & 24 \\
Ministral 3 14B & 1031 & 367 & 356 & 152 & 209 & 98 \\
Phi 4 & 2116 & 707 & 572 & 219 & 206 & 77 \\
Gemini 3.1 Flash Lite & 3 & 0 & 2 & 3 & 6 & 14 \\
Claude Sonnet 5 & 53 & 16 & 15 & 5 & 3 & 26 \\
GPT-6 Sol & 39 & 14 & 15 & 8 & 7 & 24 \\
\bottomrule\end{tabularx}\appendixcaption{table}{Distinct model-supplied tags by their number of extraction occurrences. Open-weight models cover 1,300 filings, proprietary references 300.}\label{tab:appendix_tag_reuse}\end{minipage}\par\medskip

Reclassification can change a model's edge assignment even when the underlying evidence is unchanged. Table~\ref{tab:appendix_label_stability} compares the four extractors with corresponding Task II classifiers on the same human-golden passages.

\par\medskip\noindent\begin{minipage}{\columnwidth}\centering
\footnotesize\setlength{\tabcolsep}{2pt}
\begin{tabularx}{\columnwidth}{@{}Xrrrrr@{}}\toprule
Model & $N$ & Changed & Change (\%) & Task I & Task II \\
\midrule
Gemma 4 31B & 483 & 105 & 21.7 & 77.8 & 83.0 \\
Qwen 3.6 35B A3B & 483 & 173 & 35.8 & 69.6 & 77.0 \\
Ministral-3-14B-Instruct & 483 & 286 & 59.2 & 53.4 & 80.1 \\
Phi 4 & 483 & 286 & 59.2 & 44.9 & 70.2 \\
\bottomrule\end{tabularx}
\appendixcaption{table}{Edge-label changes and accepted-set accuracy (\%) on the same 483 passages. $N$ is the number compared. Task I and Task II accuracy columns are percentages.}
\label{tab:appendix_label_stability}
\end{minipage}\par\medskip

\subsection{Annual Edge-Level Allocation}
Each panel reports the share of a model's annual extractions assigned to one edge. These distributions describe retrieval emphasis, not classification accuracy.

\begin{figure*}[p]
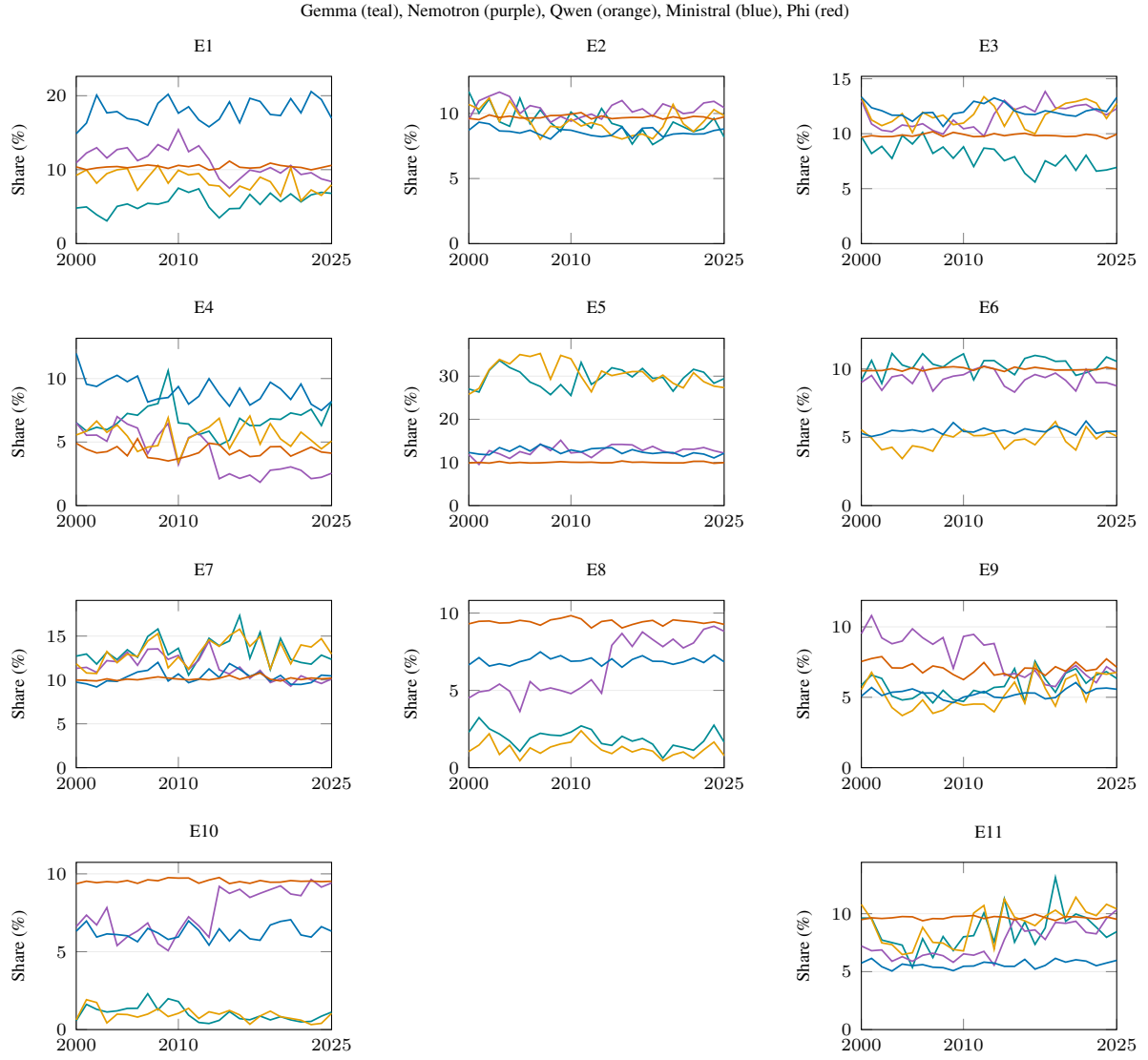
\centering
{\scriptsize Gemma (teal), Nemotron (purple), Qwen (orange), Ministral (blue), Phi (red)}\par\smallskip
\begin{minipage}[t]{.32\textwidth}\centering
\end{minipage}
\par\medskip
\caption{Annual allocation across all 11 edges. Each share uses the corresponding model and year's total extractions.}
\label{fig:appendix_annual_edges}
\end{figure*}

\subsection{Cross-Model Vocabulary Overlap}
\label{app:vocabulary_overlap}

\par\medskip\noindent\begin{minipage}{\columnwidth}\centering

\resizebox{\columnwidth}{!}{%

}
\par\smallskip
{\scriptsize\setlength{\tabcolsep}{3pt}\begin{tabular}{@{}ll@{}}
M1: Gemma 4 31B & M5: Phi 4 \\
M2: Nemotron 3 Super & M6: Gemini 3.1 Flash Lite \\
M3: Qwen 3.6 35B A3B & M7: Claude Sonnet 5 \\
M4: Ministral 3 14B & M8: GPT-6 Sol \\
\end{tabular}}\par
\appendixcaption{figure}{Vocabulary Jaccard (\%) on 300 common filings. Darker teal indicates greater overlap between mechanically normalized tag sets. Vocabulary overlap provides a complementary cross-model view: Qwen--Gemini and Gemma--Gemini are the strongest open/proprietary pairs. The latter also shares Google's model lineage, but shared training data cannot be inferred from lexical overlap alone.}
\label{fig:results_vocabulary_overlap}
\end{minipage}\par\medskip

\subsection{Overlap of Classification Failures}
\label{sec:failure_overlap}

For model $m$, let $F_m=\{i:\hat e_{m,i}\notin Y_i\}$ be its failed passages on the same 483 eligible examples. We compare all 44 models (946 pairs) using $J_{\mathrm{error}}(a,b)=|F_a\cap F_b|/|F_a\cup F_b|$. The figure shows the strongest model per family and the tied Qwen overall leaders. Failure overlap depends on error rates, so matched-accuracy pairs are particularly informative.

The two Qwen leaders share 60 errors out of a union of 94 (Jaccard: 63.8\%). Qwen3.6-27B and Gemma4-12B share 58 out of 100 (58.0\%). The equally accurate proprietary leaders, Claude Fable 5.1 and GPT-6-Astra, share 27 of a union of 43 errors (62.8\%). Only two passages are wrong for all 44 models. These patterns reveal shared difficulty alongside complementary strengths, without identifying whether the causes lie in training data, model behavior, or passage ambiguity.

\par\medskip\noindent\begin{minipage}{\columnwidth}\centering
\resizebox{.95\columnwidth}{!}{%
}
\par\vspace{2pt}
{\tiny\setlength{\tabcolsep}{3pt}\begin{tabular}{@{}ll@{}}
M1: Qwen3.6-27B ($|F|=77$) & M6: Llama3.1-8B ($|F|=228$) \\
M2: Qwen3.5-27B ($|F|=77$) & M7: DeepSeek-V2-Lite ($|F|=280$) \\
M3: Gemma4-12B ($|F|=81$) & M8: GPT-6-Astra ($|F|=35$) \\
M4: Ministral3-14B ($|F|=96$) & M9: Claude-Fable-5.1 ($|F|=35$) \\
M5: Phi4 ($|F|=144$) & M10: Gemini3.8-Flash ($|F|=42$) \\
\end{tabular}}
\appendixcaption{figure}{Failure-set Jaccard (\%) for family-leading classifiers on the common human-golden cohort. Darker teal indicates more shared errors. Legend entries give each model's error count.}
\label{fig:failure_overlap}
\end{minipage}\par\medskip

\makeatletter
\setlength{\@dblfptop}{0pt}\setlength{\@dblfpsep}{12pt}
\makeatother
\begin{figure*}[p]
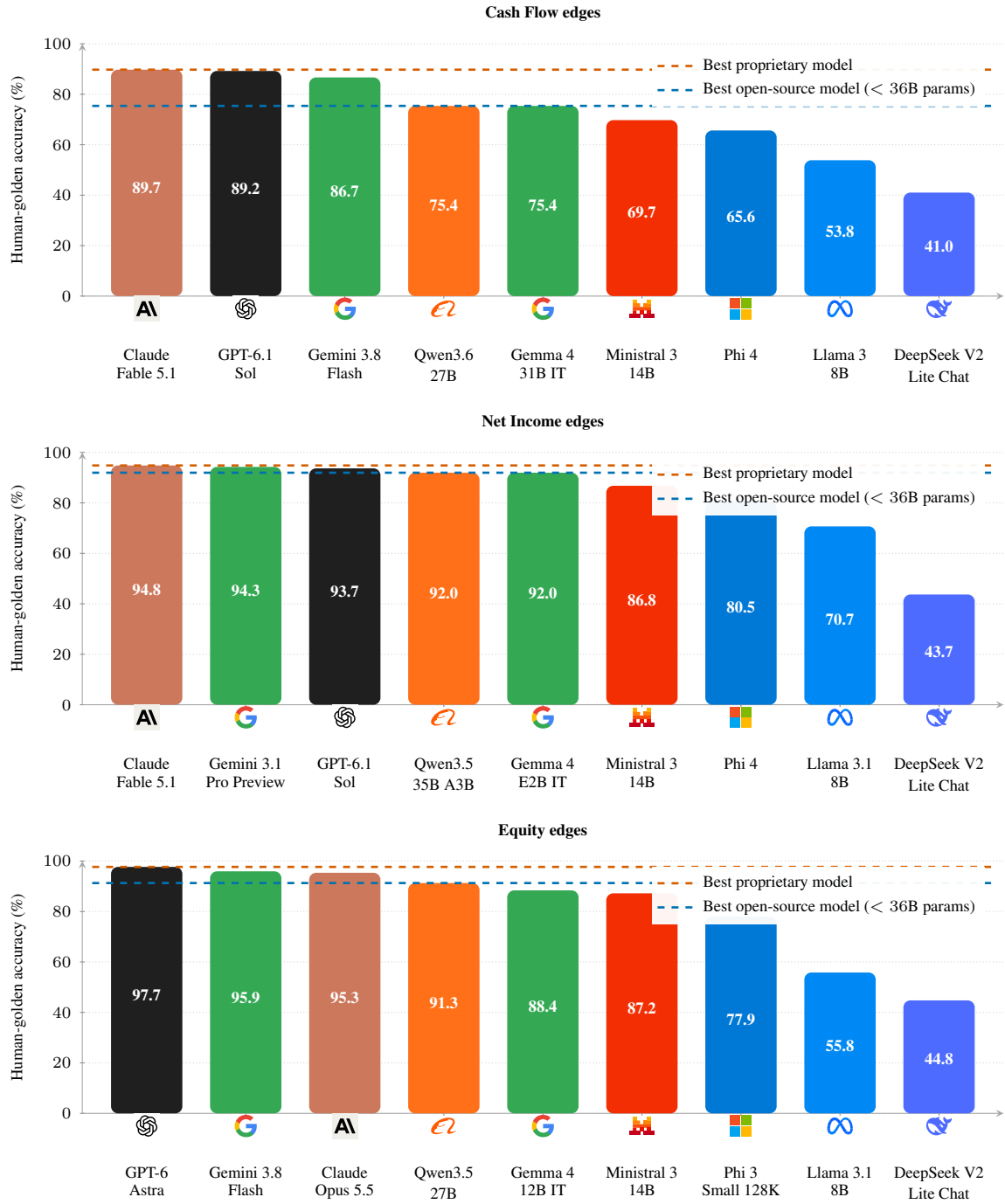
\centering
\textbf{Category-Level Family Leaders}\par\smallskip
%
\caption{Best observed model in each family for Cash Flow, Net Income, and Equity edges. Bars show accepted-set accuracy. Dashed lines identify the best proprietary and open-weight models ($<36$B parameters).}
\label{fig:appendix_category_family_best}
\end{figure*}

\begin{figure*}[p]
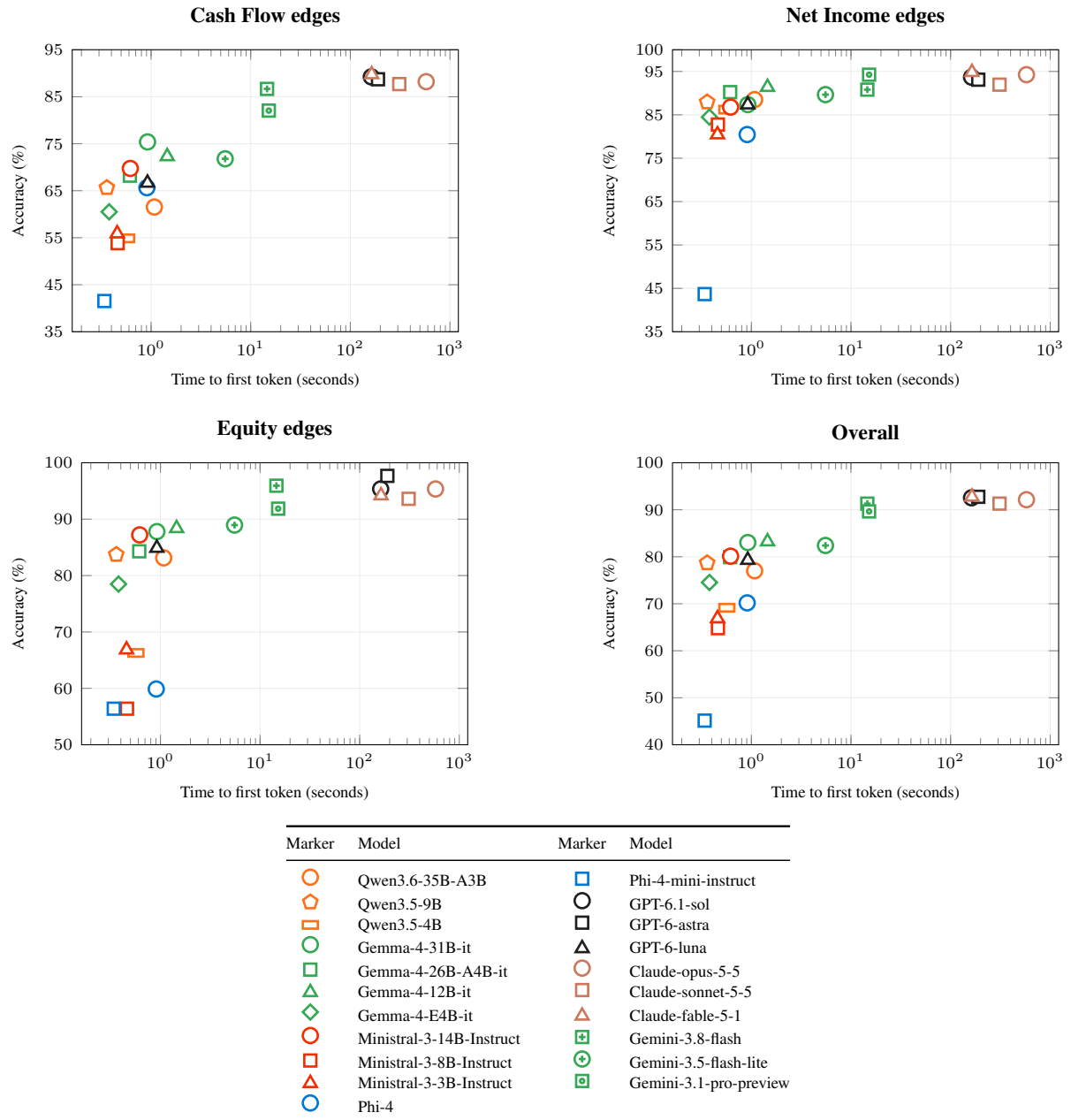
\centering

\par\medskip
{\scriptsize\setlength{\tabcolsep}{5pt}\renewcommand{\arraystretch}{1.08}
\begin{tabular}{@{}cl@{\hspace{16pt}}cl@{}}\toprule
Marker & Model & Marker & Model \\ \midrule
\tikz{\draw[color=effColor1,mark=effM1,mark size=3.2pt,line width=1pt,mark options={fill=white}] plot coordinates {(0,0)};} & Qwen3.6-35B-A3B & \tikz{\draw[color=effColor23,mark=effM23,mark size=3.2pt,line width=1pt,mark options={fill=white}] plot coordinates {(0,0)};} & Phi-4-mini-instruct \\
\tikz{\draw[color=effColor5,mark=effM5,mark size=3.2pt,line width=1pt,mark options={fill=white}] plot coordinates {(0,0)};} & Qwen3.5-9B & \tikz{\draw[color=effColor35,mark=effM35,mark size=3.2pt,line width=1pt,mark options={fill=white}] plot coordinates {(0,0)};} & GPT-6.1-sol \\
\tikz{\draw[color=effColor6,mark=effM6,mark size=3.2pt,line width=1pt,mark options={fill=white}] plot coordinates {(0,0)};} & Qwen3.5-4B & \tikz{\draw[color=effColor36,mark=effM36,mark size=3.2pt,line width=1pt,mark options={fill=white}] plot coordinates {(0,0)};} & GPT-6-astra \\
\tikz{\draw[color=effColor9,mark=effM9,mark size=3.2pt,line width=1pt,mark options={fill=white}] plot coordinates {(0,0)};} & Gemma-4-31B-it & \tikz{\draw[color=effColor37,mark=effM37,mark size=3.2pt,line width=1pt,mark options={fill=white}] plot coordinates {(0,0)};} & GPT-6-luna \\
\tikz{\draw[color=effColor10,mark=effM10,mark size=3.2pt,line width=1pt,mark options={fill=white}] plot coordinates {(0,0)};} & Gemma-4-26B-A4B-it & \tikz{\draw[color=effColor38,mark=effM38,mark size=3.2pt,line width=1pt,mark options={fill=white}] plot coordinates {(0,0)};} & Claude-opus-5-5 \\
\tikz{\draw[color=effColor11,mark=effM11,mark size=3.2pt,line width=1pt,mark options={fill=white}] plot coordinates {(0,0)};} & Gemma-4-12B-it & \tikz{\draw[color=effColor39,mark=effM39,mark size=3.2pt,line width=1pt,mark options={fill=white}] plot coordinates {(0,0)};} & Claude-sonnet-5-5 \\
\bottomrule\end{tabular}}
\caption{Accuracy versus first-token latency for every model with available Artificial Analysis data on 12 October 2026. Colors identify providers, markers identify models. The horizontal axis uses $\log_{10}t$, with $t$ in seconds.}
\label{fig:appendix_efficiency_ttft}
\end{figure*}

\begin{figure*}[p]
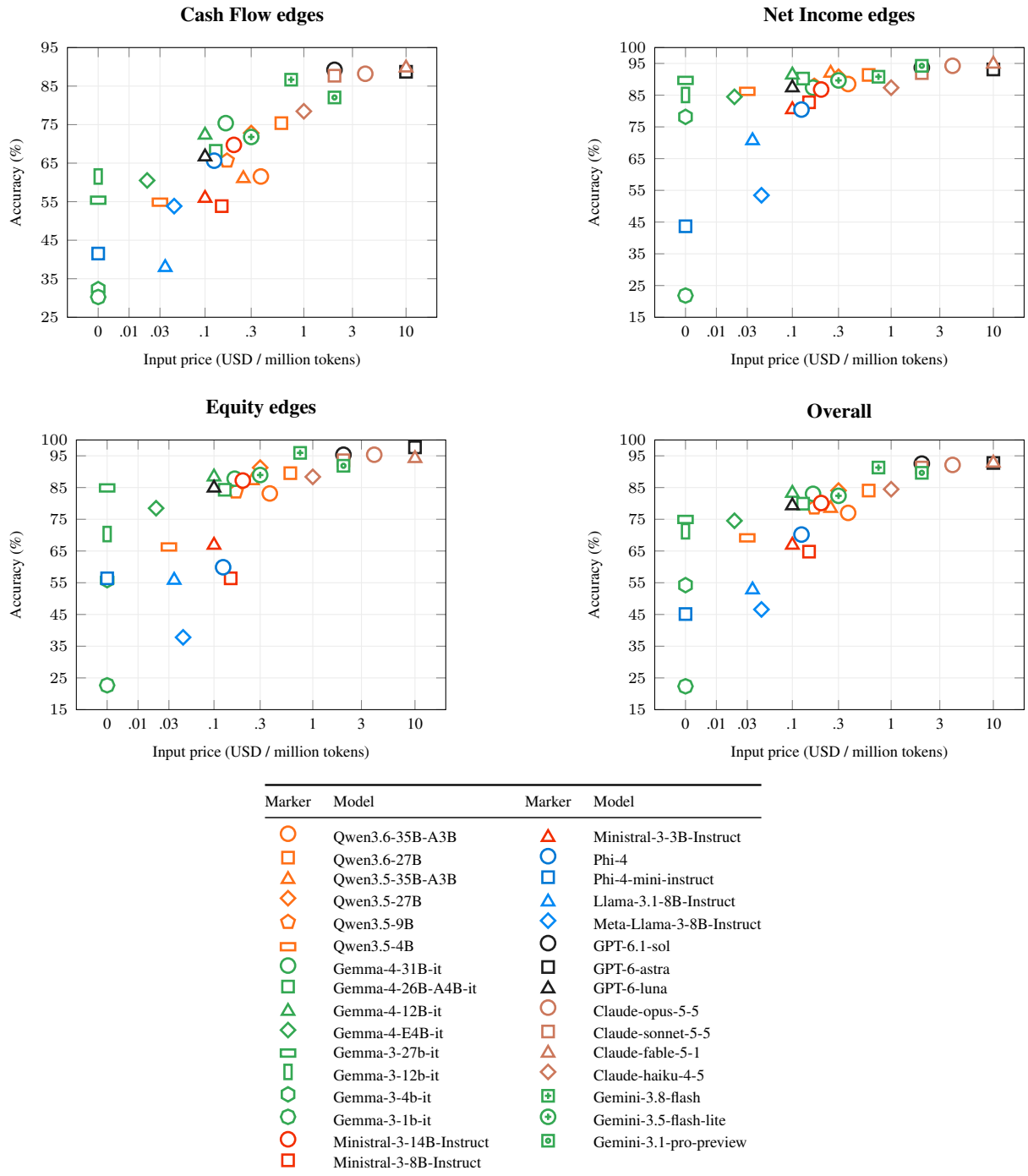
\centering

\par\medskip
{\scriptsize\setlength{\tabcolsep}{5pt}\renewcommand{\arraystretch}{1.08}
\begin{tabular}{@{}cl@{\hspace{16pt}}cl@{}}\toprule
Marker & Model & Marker & Model \\ \midrule
\tikz{\draw[color=effColor1,mark=effM1,mark size=3.2pt,line width=1pt,mark options={fill=white}] plot coordinates {(0,0)};} & Qwen3.6-35B-A3B & \tikz{\draw[color=effColor21,mark=effM21,mark size=3.2pt,line width=1pt,mark options={fill=white}] plot coordinates {(0,0)};} & Ministral-3-3B-Instruct \\
\tikz{\draw[color=effColor2,mark=effM2,mark size=3.2pt,line width=1pt,mark options={fill=white}] plot coordinates {(0,0)};} & Qwen3.6-27B & \tikz{\draw[color=effColor22,mark=effM22,mark size=3.2pt,line width=1pt,mark options={fill=white}] plot coordinates {(0,0)};} & Phi-4 \\
\tikz{\draw[color=effColor3,mark=effM3,mark size=3.2pt,line width=1pt,mark options={fill=white}] plot coordinates {(0,0)};} & Qwen3.5-35B-A3B & \tikz{\draw[color=effColor23,mark=effM23,mark size=3.2pt,line width=1pt,mark options={fill=white}] plot coordinates {(0,0)};} & Phi-4-mini-instruct \\
\tikz{\draw[color=effColor4,mark=effM4,mark size=3.2pt,line width=1pt,mark options={fill=white}] plot coordinates {(0,0)};} & Qwen3.5-27B & \tikz{\draw[color=effColor30,mark=effM30,mark size=3.2pt,line width=1pt,mark options={fill=white}] plot coordinates {(0,0)};} & Llama-3.1-8B-Instruct \\
\tikz{\draw[color=effColor5,mark=effM5,mark size=3.2pt,line width=1pt,mark options={fill=white}] plot coordinates {(0,0)};} & Qwen3.5-9B & \tikz{\draw[color=effColor31,mark=effM31,mark size=3.2pt,line width=1pt,mark options={fill=white}] plot coordinates {(0,0)};} & Meta-Llama-3-8B-Instruct \\
\tikz{\draw[color=effColor6,mark=effM6,mark size=3.2pt,line width=1pt,mark options={fill=white}] plot coordinates {(0,0)};} & Qwen3.5-4B & \tikz{\draw[color=effColor35,mark=effM35,mark size=3.2pt,line width=1pt,mark options={fill=white}] plot coordinates {(0,0)};} & GPT-6.1-sol \\
\bottomrule\end{tabular}}
\caption{Accuracy versus input-token price for every model with available Artificial Analysis data on 12 October 2026. Colors identify providers, markers identify models. Horizontal positions use $\log_{10}(1+c/0.01)$, with $c$ in USD per million input tokens.}
\label{fig:appendix_efficiency_cost}
\end{figure*}
\begin{figure*}[p]\raggedright
\section{Worked Annotation and Model-Response Examples}
\label{app:worked_examples}
The first two cases illustrate overlapping reviewer selections and divergent model predictions. The third shows unanimous reviewer agreement. Each panel includes all 44 classifiers, grouped by their saved answer. Cases illustrate these patterns rather than their prevalence.
\begin{appendixquote}[fontupper={\fontsize{8}{9}\selectfont\raggedright\sloppy}]{Human annotation instruction}
Select every supported relationship, or None by itself.
\end{appendixquote}
\textbf{Case 1: Converted Organics Inc (2010)}\par\smallskip
\begin{minipage}[t]{.47\textwidth}\vspace{0pt}
\begin{appendixquote}{Source evidence}
The capital outlay of \$14.6 million came from the \$25.4 million raised by our initial public offering of stock and the issuance of New Jersey Economic Development Bonds
\end{appendixquote}
\begin{appendixquote}{Reviewer selections and majority answer}
R1: E1, E2, E11\par R2: E1, E11\par R3: E1, E11\par Accepted answer set: E1, E11
\end{appendixquote}
\end{minipage}\hfill\begin{minipage}[t]{.51\textwidth}\vspace{0pt}
\begin{appendixquote}[fontupper={\fontsize{8}{9}\selectfont\raggedright\sloppy}]{Saved model answers}
\textbf{E1 (accepted):} Net Noncash Assets <--> Cash Flow from Finances\par Claude-fable-5-1, GPT-6-astra, GPT-6.1-sol, DeepSeek-V2-Lite-Chat, Llama-3.2-1B, Llama-3.2-3B, Llama-3-8B, Phi-3-mini-128k, Qwen3.5-0.8B, Qwen3.5-2B, Gemma-3-1b
\par\smallskip
\textbf{E2 (not accepted):} Cash Flow from Investments <--> Net Noncash Assets\par GPT-6-luna
\par\smallskip
\textbf{E3 (not accepted):} Net Noncash Assets <--> Cash Flow from Operations\par DeepSeek-llm-7b-chat
\par\smallskip
\textbf{E9 (not accepted):} Cash Flow from Finances <--> Dividends\par Gemma-3-270m
\par\smallskip
\textbf{E10 (not accepted):} Contributed Capital <--> Net Noncash Assets\par Gemini-3.1-pro-preview, Llama-3.1-8B, Ministral-3-8B, Phi-3-medium-128k, Qwen3.5-4B, Gemma-4-E2B
\par\smallskip
\textbf{E11 (accepted):} Cash Flow from Finances <--> Contributed Capital\par Claude-haiku-4-5, Claude-opus-5-5, Claude-sonnet-5-5, Gemini-3.5-flash-lite, Gemini-3.8-flash, Ministral-3-14B, Ministral-3-3B, Phi-3-small-128k, Phi-3.5-mini, Phi-4, Phi-4-mini, Qwen3.5-27B, Qwen3.5-35B-A3B, Qwen3.5-9B, Qwen3.6-27B, Qwen3.6-35B-A3B, DeepSeek-moe-16b-chat, Gemma-3-12b, Gemma-3-27b, Gemma-3-4b, Gemma-4-12B, Gemma-4-26B-A4B, Gemma-4-31B, Gemma-4-E4B
\end{appendixquote}
\end{minipage}\end{figure*}
\begin{figure*}[p]\raggedright
\textbf{Case 2: MICROMUSE INC (2005)}\par\smallskip
\begin{minipage}[t]{.47\textwidth}\vspace{0pt}
\begin{appendixquote}{Source evidence}
our future acquisitions, if any, may result in issuances of our equity securities which dilute our stockholders, the incurrence of debt, large one-time write-offs and creation of goodwill or other intangible assets that could result in amortization and impairment expense.
\end{appendixquote}
\begin{appendixquote}{Reviewer selections and majority answer}
R1: E1, E5, E10\par R2: E5, E10\par R3: E5, E10\par Accepted answer set: E5, E10
\end{appendixquote}
\end{minipage}\hfill\begin{minipage}[t]{.51\textwidth}\vspace{0pt}
\begin{appendixquote}[fontupper={\fontsize{8}{9}\selectfont\raggedright\sloppy}]{Saved model answers}
\textbf{E1 (not accepted):} Net Noncash Assets <--> Cash Flow from Finances\par DeepSeek-V2-Lite-Chat, Llama-3.2-1B, Llama-3-8B, Phi-3.5-mini, Phi-4, Phi-4-mini, Gemma-3-1b
\par\smallskip
\textbf{E3 (not accepted):} Net Noncash Assets <--> Cash Flow from Operations\par Qwen3.5-0.8B, Qwen3.5-2B
\par\smallskip
\textbf{E4 (not accepted):} Revenue <--> Net Noncash Assets\par Phi-3-medium-128k
\par\smallskip
\textbf{E5 (accepted):} Net Noncash Assets <--> Expense\par Claude-fable-5-1, Claude-haiku-4-5, Claude-sonnet-5-5, Gemini-3.1-pro-preview, Gemini-3.8-flash, GPT-6-astra, GPT-6.1-sol, Llama-3.1-8B, Llama-3.2-3B, Ministral-3-3B, Qwen3.5-27B, Qwen3.5-35B-A3B, Qwen3.5-4B, Qwen3.5-9B, Qwen3.6-27B, Qwen3.6-35B-A3B, Gemma-3-12b, Gemma-3-27b, Gemma-3-4b, Gemma-4-12B, Gemma-4-26B-A4B, Gemma-4-31B, Gemma-4-E2B
\par\smallskip
\textbf{E6 (not accepted):} Net Noncash Assets <--> Interest Expense\par Phi-3-mini-128k
\par\smallskip
\textbf{E9 (not accepted):} Cash Flow from Finances <--> Dividends\par DeepSeek-llm-7b-chat, Gemma-3-270m
\par\smallskip
\textbf{E10 (accepted):} Contributed Capital <--> Net Noncash Assets\par Claude-opus-5-5, Gemini-3.5-flash-lite, Ministral-3-14B, Ministral-3-8B
\par\smallskip
\textbf{E11 (not accepted):} Cash Flow from Finances <--> Contributed Capital\par GPT-6-luna, Phi-3-small-128k, DeepSeek-moe-16b-chat, Gemma-4-E4B
\end{appendixquote}
\end{minipage}\end{figure*}
\begin{figure*}[p]\raggedright
\textbf{Case 3: RIVIERA HOLDINGS CORP (2010)}\par\smallskip
\begin{minipage}[t]{.47\textwidth}\vspace{0pt}
\begin{appendixquote}{Source evidence}
Depreciation and amortization expenses decreased due to lower depreciation expense due to a reduction in depreciable assets during 2009.
\end{appendixquote}
\begin{appendixquote}{Reviewer selections and majority answer}
R1: E5\par R2: E5\par R3: E5\par Accepted answer set: E5
\end{appendixquote}
\begin{appendixquote}[fontupper={\fontsize{8}{9}\selectfont\raggedright\sloppy}]{Task II: complete saved user message}
Evidence text:\par Depreciation and amortization expenses decreased due to lower depreciation expense due to a reduction in depreciable assets during 2009.\par \par Item 7 context:\par Depreciation\par and amortization expenses decreased due to lower depreciation expense due to a\par reduction in depreciable assets during 2009.
\end{appendixquote}
\end{minipage}\hfill\begin{minipage}[t]{.51\textwidth}\vspace{0pt}
\begin{appendixquote}[fontupper={\fontsize{8}{9}\selectfont\raggedright\sloppy}]{Saved model answers}
\textbf{E1 (not accepted):} Net Noncash Assets <--> Cash Flow from Finances\par Llama-3.2-1B, Gemma-3-1b
\par\smallskip
\textbf{E2 (not accepted):} Cash Flow from Investments <--> Net Noncash Assets\par Phi-3-medium-128k, DeepSeek-llm-7b-chat
\par\smallskip
\textbf{E3 (not accepted):} Net Noncash Assets <--> Cash Flow from Operations\par Llama-3.2-3B, Phi-3-mini-128k, Phi-4-mini, Qwen3.5-2B
\par\smallskip
\textbf{E4 (not accepted):} Revenue <--> Net Noncash Assets\par Gemma-3-270m
\par\smallskip
\textbf{E5 (accepted):} Net Noncash Assets <--> Expense\par Claude-fable-5-1, Claude-haiku-4-5, Claude-opus-5-5, Claude-sonnet-5-5, Gemini-3.1-pro-preview, Gemini-3.5-flash-lite, Gemini-3.8-flash, GPT-6-astra, GPT-6-luna, GPT-6.1-sol, DeepSeek-V2-Lite-Chat, Llama-3.1-8B, Llama-3-8B, Ministral-3-14B, Ministral-3-3B, Ministral-3-8B, Phi-3-small-128k, Phi-3.5-mini, Phi-4, Qwen3.5-0.8B, Qwen3.5-27B, Qwen3.5-35B-A3B, Qwen3.5-4B, Qwen3.5-9B, Qwen3.6-27B, Qwen3.6-35B-A3B, Gemma-3-12b, Gemma-3-27b, Gemma-3-4b, Gemma-4-12B, Gemma-4-26B-A4B, Gemma-4-31B, Gemma-4-E2B, Gemma-4-E4B
\par\smallskip
\textbf{E9 (not accepted):} Cash Flow from Finances <--> Dividends\par DeepSeek-moe-16b-chat
\end{appendixquote}
\end{minipage}\end{figure*}

\end{document}